\documentclass[hidelinks,nohyperref]{article} %

\usepackage[accepted]{icml2023}

\usepackage{amsmath,amsfonts,bm}

\def\eqref#1{equation~\ref{#1}}

\def\1{\bm{1}}

\DeclareMathAlphabet{\mathsfit}{\encodingdefault}{\sfdefault}{m}{sl}
\SetMathAlphabet{\mathsfit}{bold}{\encodingdefault}{\sfdefault}{bx}{n}

\DeclareMathOperator*{\argmin}{arg\,min}

\usepackage{hyperref}
\usepackage{url}

\usepackage[utf8]{inputenc} %
\usepackage[T1]{fontenc}    %
\usepackage{hyperref}       %
\usepackage{url}            %
\usepackage{booktabs}       %
\usepackage{amsfonts}       %
\usepackage{nicefrac}       %
\usepackage{microtype}      %
\usepackage{lipsum}
\usepackage{fancyhdr}       %
\usepackage{graphicx}       %
\graphicspath{{media/}}     %

\usepackage{amsmath, amsthm}
\usepackage{amssymb}
\usepackage{booktabs}
\usepackage{float}
\usepackage{array}
\usepackage{xcolor}
\usepackage{todonotes}
\usepackage{multirow}
\usepackage{diagbox}
\usepackage{pbox}
\usepackage{caption}
\usepackage{subcaption}
\usepackage{lscape}
\usepackage{pdfpages}
\usepackage{svg}
\usepackage{wrapfig}
\usepackage{multicol}
\usepackage{multirow}
\usepackage{graphicx}

\usepackage[acronym]{glossaries}
\usepackage{mathtools}

\usepackage[capitalize,noabbrev]{cleveref}

\theoremstyle{plain}

\theoremstyle{definition}

\theoremstyle{remark}

\icmltitlerunning{\OurMethod{}: Tabular Data Leakage in Federated Learning}

\newcommand\OurMethod{TabLeak}
\newcommand{\eg}{\textit{e.g., }}
\newcommand{\ie}{\textit{i.e., }}
\usepackage{tikz}

\usepackage[capitalize]{cleveref}
\crefname{equation}{Eq.}{Eq.}
\creflabelformat{equation}{#2\textup{#1}#3}
\crefname{section}{Sec.}{Secs.}
\Crefname{section}{Section}{Sections}
\Crefname{table}{Table}{Tables}
\crefname{table}{Tab.}{Tabs.}
\Crefname{algorithm}{Algorithm}{Algorithms}
\crefname{algorithm}{Alg.}{Algs.}
\Crefname{appendix}{Appendix}{Appendices}
\crefname{appendix}{App.}{App.}
\crefname{figure}{Fig.}{Fig.}

\author{Antiquus S.~Hippocampus, Natalia Cerebro \& Amelie P. Amygdale \thanks{ Use footnote for providing further information
about author (webpage, alternative address)---\emph{not} for acknowledging
funding agencies.  Funding acknowledgements go at the end of the paper.} \\
Department of Computer Science\\
Cranberry-Lemon University\\
Pittsburgh, PA 15213, USA \\
\texttt{\{hippo,brain,jen\}@cs.cranberry-lemon.edu} \\
\And
Ji Q. Ren \& Yevgeny LeNet \\
Department of Computational Neuroscience \\
University of the Witwatersrand \\
Joburg, South Africa \\
\texttt{\{robot,net\}@wits.ac.za} \\
\AND
Coauthor \\
Affiliation \\
Address \\
\texttt{email}
}

\newcommand{\alglinelabel}{%
	\addtocounter{ALC@line}{-1}%
	\refstepcounter{ALC@line}%
	\label %
}

\begin{document}

\twocolumn[
\icmltitle{\OurMethod{}: Tabular Data Leakage in Federated Learning}

\icmlsetsymbol{equal}{*}

\begin{icmlauthorlist}
\icmlauthor{Mark Vero}{itet}
\icmlauthor{Mislav Balunović}{infk}
\icmlauthor{Dimitar I. Dimitrov}{infk}
\icmlauthor{Martin Vechev}{infk}
\end{icmlauthorlist}

\icmlaffiliation{itet}{Department of Information Technology and Electrical Engineering, ETH Zurich, Zurich, Switzerland}
\icmlaffiliation{infk}{Department of Computer Science, ETH Zurich, Zurich, Switzerland}

\icmlcorrespondingauthor{Mark Vero}{mveroe@ethz.ch}

\icmlkeywords{Machine Learning, ICML}

\vskip 0.3in
]

\printAffiliationsAndNotice{}  %

\begin{abstract}
    While federated learning (FL) promises to preserve privacy, recent works in the image and text domains have shown that training updates leak private client data. However, most high-stakes applications of FL (\eg in healthcare and finance) use tabular data, where the risk of data leakage has not yet been explored. A successful attack for tabular data must address two key challenges unique to the domain: (i) obtaining a solution to a high-variance mixed discrete-continuous optimization problem, and (ii) enabling human assessment of the reconstruction as unlike for image and text data, direct human inspection is not possible.
    In this work we address these challenges and propose \OurMethod{}, the first comprehensive reconstruction attack on tabular data. \OurMethod{} is based on two key contributions: (i) a method which leverages a softmax relaxation and pooled ensembling to solve the optimization problem, and (ii) an entropy-based uncertainty quantification scheme to enable human assessment. 
    We evaluate \OurMethod{} on four tabular datasets for both FedSGD and FedAvg training protocols, and show that it successfully breaks several settings previously deemed safe. For instance, we extract large subsets of private data at $>90\%$ accuracy even at the large batch size of $128$. Our findings demonstrate that current high-stakes tabular FL is excessively vulnerable to leakage attacks.
\end{abstract}

\section{Introduction}
\label{sec:introduction}
\begin{figure}[t]
	\centering
	\includegraphics[trim={0 3.9cm 2.5cm 0},clip, width=0.45\textwidth]{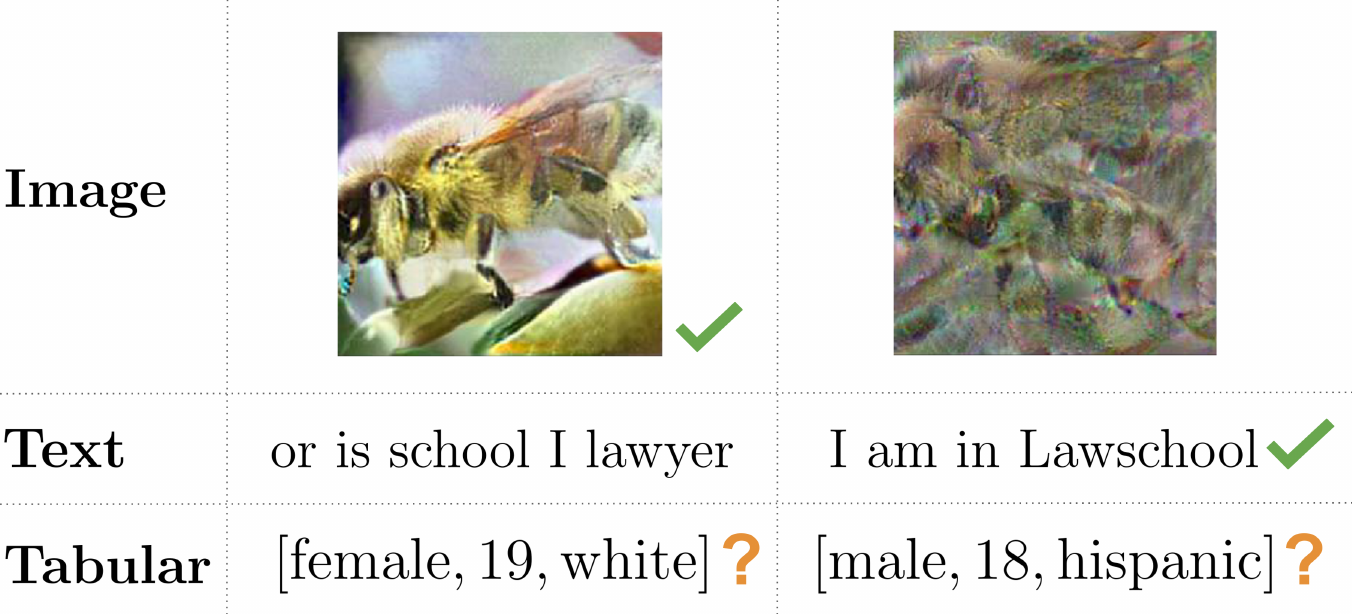}
	\vspace{-0.5em}
	\caption{Comparison of image, text, and tabular data reconstruction. While the attack success can be judged by human inspection in images and text, for tabular data it is not possible, as both reconstructions \textit{look} plausible. The image reconstruction example is taken from \citet{Yin2021}.}
	\vspace{-0.9em}
	\label{fig:comparison}
\end{figure}

\begin{figure*}[t]
	\centering
	\includegraphics[trim={0 4.2cm 2.7cm 0},clip, width=.99\textwidth]{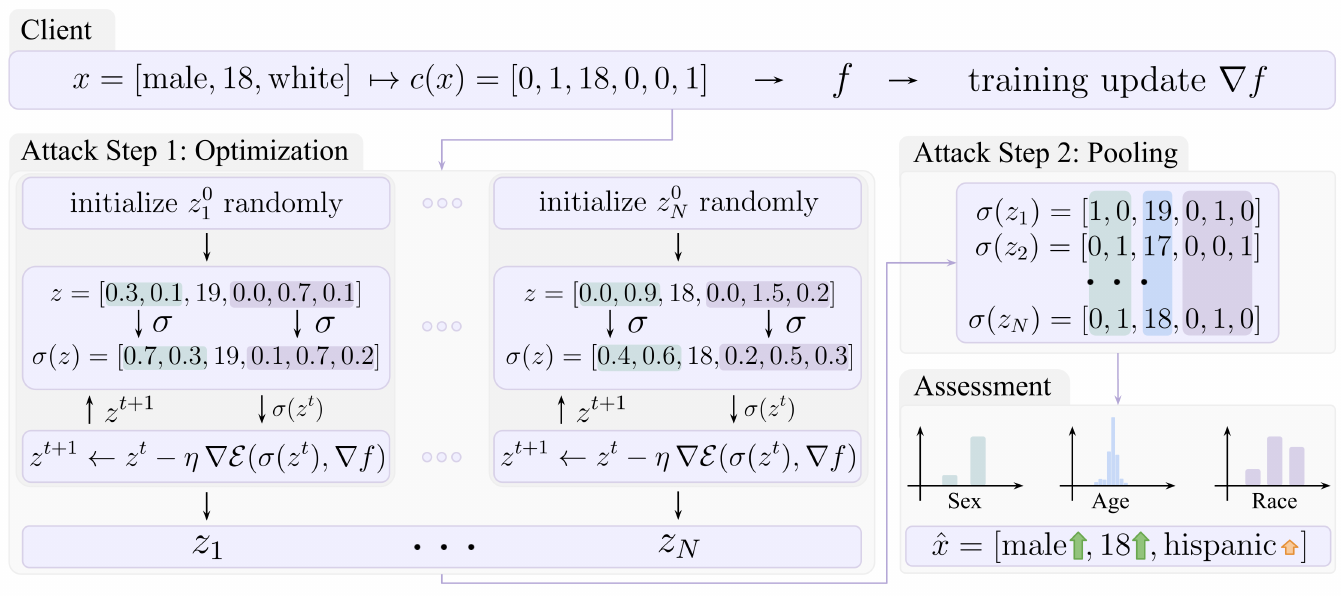}
	\vspace{-0.2cm}
	\caption[short]{Overview of \OurMethod{}. Our approach transforms the optimization problem into a fully continuous one by optimizing continuous versions of the discrete features, obtained by applying softmax (Attack Step 1, middle boxes), resulting in $N$ candidate solutions (Attack Step 1, bottom). Then, we pool together an ensemble of $N$ different solutions $z_1, z_2, ..., z_N$ obtained from the optimization to reduce the variance of the reconstruction (Attack Step 2). Finally, we assess the quality of the reconstruction by computing the entropy from the feature distributions in the ensemble (Assessment).}
	\label{fig:acceptance_figure}
	\vspace{-1.0em}
\end{figure*}

Federated Learning~\citep{McMahan2016} (FL) has emerged as the most prominent approach to training machine learning models collaboratively without requiring sensitive data of different parties to be collected in a central database.
While prior work has examined privacy leakage from exchanged updates in FL on images~\citep{Zhu2019,Geiping2020,Yin2021} and text~\citep{Deng2021, Dimitrov2022, Gupta2022},
many applications of FL involve tabular datasets incorporating highly sensitive personal data such as financial information and health status~\citep{Borisov2021,Long2021, Rieke2020}. However, as no prior work has studied the issue of privacy leakage in tabular data, we are unaware of the true extent of its risks. This is also a cause of concern for US and UK public institutions which have recently launched a \$1.6 mil. prize competition\footnote{\url{https://petsprizechallenges.com/}} to develop privacy-preserving FL solutions for financial fraud detection and infection risk prediction, both being tabular datasets.

\paragraph{Ingredients of a Data Leakage Attack}
A successful attack builds on two pillars: (i) ability to reconstruct private data from client updates with high accuracy, and (ii) a mechanism that allows a human to assess the obtained reconstructions \emph{without} knowledge of the true data. Advancing along the first pillar typically requires leveraging the unique aspects of the given domain, \eg image attacks employ image priors~\citep{Geiping2020,Yin2021}, while attacks on text make use of pre-trained language models~\citep{Dimitrov2022, Gupta2022}. However, in the image and text domains, the second pillar naturally comes for free, as the credibility of the obtained data can be assessed simply by human inspection, in contrast to tabular data, where this is not possible, as illustrated in \cref{fig:comparison}.

\paragraph{Key Challenges}
A strong attack for tabular data must address two unique challenges, one along each pillar: (i) due to the presence of both discrete and continuous features, the attack needs to solve a mixed discrete-continuous optimization problem of high variance, and (ii) unlike with image and text data, assessing the quality of the reconstruction is no longer possible via human inspection, requiring a mechanism to quantify the uncertainty of the reconstruction. %

\paragraph{This Work}
In this work we propose the first comprehensive attack on tabular data, \OurMethod{}, addressing the above challenges. Using our attack, we conduct the first comprehensive evaluation of the privacy risks posed by data leakage in tabular FL.
We provide an overview of our approach in~\cref{fig:acceptance_figure}, showing the reconstruction of a client's private data point $x = [\texttt{male, 18, white}]$, from the corresponding update $\nabla f$ received by the server. We tackle the first challenge in two steps. In Attack Step 1, we create $N$ separate optimization problems with different initializations. We transform the mixed discrete-continuous optimization problem into a fully continuous one using a softmax relaxation. Once optimization completes, in Attack Step 2, we reduce the variance of the final reconstruction by pooling over the different solutions. To address challenge (ii, Assessment), we rely on the observation that when the $N$ reconstructions agree on a certain feature, it tends to be reconstructed well. We measure the agreement using entropy. In our example, \textit{sex} and \textit{age} exhibit a low entropy reconstruction and are also correct. Meanwhile, the high disagreement over the \textit{race} feature is indicative of its incorrect reconstruction.

Comparing our domain-specific attack with prior works adapted from other domains on both FL protocols, FedSGD and FedAvg in various settings on four popular tabular datasets, we reveal the high vulnerability of such systems on tabular data, even in scenarios previously deemed as safe. We observe that on small batch sizes tabular FL systems are nearly transparent, where most attacks recover $>90\%$ of the private data. Further, our attack retrieves $70.8\%$ - $84.9\%$ of the client data at the practically relevant batch size of $32$ on the examined datasets, improving by $12.7\%$ - $14.5\%$ on prior art. Additionally, even on batch sizes as large as $128$, we show how an adversary can recover a quarter of the private data well above $90\%$ accuracy; leading to alarming conclusions about the privacy of FL on tabular data.

\paragraph{Main Contributions}

Our main contributions are:

\begin{itemize}
	\item First effective domain-specific data leakage attack on tabular data called \OurMethod{}, enabling novel insights into the unique aspect of tabular data leakage.
	\item An effective uncertainty quantification scheme, enabling the assessment of obtained samples and allowing an attacker to extract highly accurate subsets of features even from poor reconstructions.
	\item An extensive experimental evaluation, revealing the excessively high vulnerability of FL with tabular data by successfully conducting attacks even in setups previously deemed safe.
\end{itemize}

\section{Background and Related Work}
\label{sec:background_and_related_work}

\paragraph{Federated Learning}
FL is a framework developed to facilitate the distributed training of a parametric model while preserving the privacy of the data at source~\citep{McMahan2016}. Formally, we have a parametric function $f_{\theta}(x)=y$, where $\theta$ are the parameters. Given a dataset as the union of private datasets of clients $\mathcal{S} = \bigcup_{k=1}^K \mathcal{S}_k$, we now wish to find a $\theta^*$ such that $\frac{1}{N}\sum_{(x_i, y_i) \in \mathcal{S}} \,\mathcal{L}(f_{\theta^*}(x_i), y_i)$ is minimized, without first collecting the dataset $\mathcal{S}$ in a central database.
\citet{McMahan2016} propose two training algorithms: FedSGD (a similar algorithm was also proposed by \citet{Shokri2014}) and FedAvg, that allow for the distributed training of $f_{\theta}$, while keeping the data partitions $\mathcal{S}_k$ at client sources. The two protocols differ in how the clients compute their local updates in each step of training. In FedSGD, each client calculates the update gradient with respect to a randomly selected batch of their own data and shares it with the server. During FedAvg, the clients conduct a few epochs of local training on their own data before sharing their resulting parameters with the server. In each case, after the server has received the gradients/parameters from the clients, it aggregates them, updates the model, and broadcasts it to the clients; concluding an FL training step.

\paragraph{Data Leakage Attacks}
Although the design goal of FL was to preserve the privacy of clients' data, recent work has uncovered substantial vulnerabilities. \citet{Melis2019} first presented how one can infer certain properties of the clients' data. Later, \citet{Zhu2019} demonstrated that an \textit{honest-but-curious} server can use the current state of the model and the received updates to reconstruct the clients' data, breaking the privacy promise of FL. Under this threat model, there has been extensive research on designing tailored attacks for images~\citep{Geiping2020, Zhao2020, Geng2021, huang2021evaluating, Jin2021, Balunovic2021, Yin2021, Jeon2021,Dimitrov22} and natural language~\citep{Deng2021, Dimitrov2022, Gupta2022}. However, no prior work has comprehensively dealt with data leakage attacks on tabular data, despite its significance in real-world high-stakes applications~\citep{Borisov2021}. While, \citet{tabexploring2022} describe an attack on tabular data where a malicious client learns some distributional information from other clients, they do not reconstruct any private data points. Some works also consider a threat scenario where a \textit{malicious server} may change the model or the updates sent to the clients~\citep{Fowl2021,Wen2022}; but in this work we focus on the honest-but-curious setting.

In FedSGD, given the gradient $\nabla_{\theta}\, \mathcal{L}(f_{\theta}(x), y)$ of some client (shorthand: $g(x,y)$), we solve the following optimization problem to retrieve the client's private data $(x,y)$:
\vspace{-0.5em}
\begin{equation}
    \vspace{-0.5em}
    \label{eq:gradient_inversion}
    \hat{x}, \,\hat{y} = \argmin_{x',y'} \mathcal{E}(g(x,y), g(x', y')) + \lambda \mathcal{R}(x').
\end{equation}
In \cref{eq:gradient_inversion} we denote the \textit{gradient matching loss} as $\mathcal{E}$ and $\mathcal{R}$ is an optional regularizer for the reconstruction. The work of \citet{Zhu2019} used the mean squared error for $\mathcal{E}$, on which \citet{Geiping2020} improved using the cosine similarity loss. \citet{Zhao2020} first demonstrated that the private labels $y$ can be estimated before solving \cref{eq:gradient_inversion}, reducing the complexity of \cref{eq:gradient_inversion} and improving the attack results. Their method was later extended to batches by \citet{Yin2021} and refined by \citet{Geng2021}. \cref{eq:gradient_inversion} is typically solved using continuous optimization tools such as L-BFGS~\citep{Liu1989} and Adam~\citep{Kingma2014}.
Although analytical approaches exist, they do not generalize to batches with more than a single data point~\citep{Zhu2020}.

\paragraph{Domain-Specific Attacks}
Depending on the data domain, distinct tailored alterations to \cref{eq:gradient_inversion} have been proposed in the literature, \eg using the total variation regularizer for images~\citep{Geiping2020} and exploiting pre-trained language models in language tasks~\citep{Dimitrov2022, Gupta2022}. These mostly non-transferable domain-specific solutions are necessary as each domain poses unique challenges. Our work is first to identify and tackle the key challenges to data leakage in the tabular domain.

\paragraph{Privacy Threat of Tabular FL}
Regulations and personal interests prevent institutions from sharing privacy-sensitive tabular data, such as STI and drug test results, social security numbers, credit scores, and passwords. To this end, FL was proposed to enable inter-owner usage of such data. However, in a strict sense, if FL on tabular data leaks \emph{any} private information, it does not fulfill its original design purpose, severely undermining trust in institutions employing such solutions. In our work we show that tabular FL, in fact, leaks \emph{large amounts} of private information.

\paragraph{Mixed Type Tabular Data}
Mixed type tabular data is commonly used in healthcare, finance, and social sciences, which entail high-stakes privacy-critical applications~\citep{Borisov2021}. Here, data is collected in a table with mostly human-interpretable columns, \eg age, and race of an individual. Formally, let $x \in \mathcal{X}$ be one \textit{row} of data and let $\mathcal{X}$ contain $K$ discrete columns and $L$ continuous columns, \ie $\mathcal{X} = \mathcal{D}_1 \times \dots \times \mathcal{D}_K \times \mathcal{U}_1 \times \dots \times \mathcal{U}_L$, where $\mathcal{D}_i \subset \mathbb{N}$ and $\mathcal{U}_i \subset \mathbb{R}$. For processing with neural networks, discrete features are usually one-hot encoded, while continuous features are preserved.
The one-hot encoding of the $i$-th discrete feature $x^D_i$ is a binary vector $c^D_i(x)$ of length $|\mathcal{D}_i|$ that has a single non-zero entry at the position marking the encoded category. We retrieve the represented category by taking the argmax of $c^D_i(x)$ (projection to obtain $x$). Using the described encoding, one row of data $x \in \mathcal{X}$ is encoded as: $c(x) = \left[c^D_1(x), \, \dots, \,c^D_K(x),\, x^C_1,\, \dots,\, x^C_L \right]$, containing $d \coloneqq L + \sum_{i=1}^K |\mathcal{D}_i|$ entries.

\section{Tabular Leakage}
\label{sec:methods}
In this section, we briefly summarize the challenges in tabular leakage and present our solution to these, followed by our end-to-end reconstruction attack.

\paragraph{Key Challenges}
In the tabular domain, a strong attack has to address two unique challenges: (i) the presence of both categorical and continuous features requires the attacker to solve a significantly harder mixed discrete-continuous optimization problem of higher variance (addressed in \cref{subsec:softmax_trick} and \cref{subsec:pooled_ensembling}), and (ii) as exemplified previously in \cref{fig:comparison}, in contrast to images and text, it is hard for an unassisted adversary to assess the credibility of the reconstructed data in the tabular domain (addressed in \cref{subsec:entropy_uncertainty}).

\subsection{Building a Strong Base Attack}
We solve challenge (i) by introducing two components to our attack; a softmax relaxation to turn the mixed discrete-continuous problem into a fully continuous one (see \cref{subsec:softmax_trick}), and pooled ensembling to reduce the variance in the final reconstruction (see \cref{subsec:pooled_ensembling}).

\subsubsection{The Softmax Relaxation}
\label{subsec:softmax_trick}
In accordance with prior literature on data leakage attacks, we aim to conduct the optimization in continuous domain. For this we employ the softmax relaxation, which turns the hard mixed discrete-continuous optimization problem into a fully continuous one. This drastically reduces its complexity, while still facilitating the recovery of correct discrete structures.

The recovery of one-hot vectors requires the integer constraints of all entries taking values in $\{0, 1\}$ and summing to one. Relaxing the integer constraints by allowing the reconstructed entries to take real values in $[0, 1]$, we are still left with a constrained optimization problem not well suited for popular continuous optimization tools, such as Adam~\citep{Kingma2014}. Therefore, we aim to implicitly enforce the constraints introduced above.

For this, we extend the method of \citet{Zhu2019} used for inverting the discrete labels when jointly optimizing for both the labels and the data. Let $z \in \mathbb{R}^d$ be our approximate intermediate solution for the true one-hot encoded data $c(x)$ during optimization. Then we can implicitly enforce all constraints described above by applying a \textit{softmax} to $z_i^D$ for all $i$ between 1 and $K$, \ie define:
\begin{equation}
\label{eq:softmax_definition}
    \sigma(z_i^D)[j] \coloneqq \frac{\text{exp}(z^{D}_i[j])}{\sum_{k=1}^{|\mathcal{D}_i|}\text{exp}(z^D_i[k])}\qquad \forall j \in \mathcal{D}_i.
\end{equation}
Therefore, in each round of optimization we will have the following approximation of the true data point: $c(x) \approx \sigma(z) = \left [ \sigma(z^{D}_1), \, \dots,\, \sigma(z^{D}_K), \, z^C_1, \, \dots, \, z^C_L \right]$. In order to preserve notational simplicity, we write $\sigma(z)$ to mean the application of softmax to each group of entries representing a given categorical variable separately. Inverting a batch of data, the softmax is applied in parallel to the batch points.

\subsubsection{Pooled Ensembling}
\label{subsec:pooled_ensembling}
In general, the data leakage optimization problem possesses multiple local minima~\citep{Zhu2020} and is sensitive to initialization~\citep{Wei2020}. Additionally, we observed and confirmed in a targeted experiment in \cref{appendix:studying_pooling} that in tabular data the mix of discrete and continuous features introduces further variance, in contrast to image and text, where the problem is fully continuous or fully discrete, respectively. We alleviate this problem by running independent optimization processes with different initializations and ensembling their results through feature-wise pooling.

Exploiting the structural regularity of tabular data, we can combine independent reconstructions to obtain an improved and more robust final estimate of the true data by applying feature-wise pooling. Formally, we run $N$ independent rounds of optimization with $i.i.d.$ initializations recovering potentially different reconstructions $\left \{ \sigma(z_j) \right \}_{j=1}^N$. Then, we obtain a final estimate of the true encoded data, denoted as $\sigma(\hat{z})$, by pooling across these reconstructions in parallel for each batch-point and feature:
\begin{align}
\label{eq:pooling_non_projected}
    &\sigma^D_i(\hat{z}) = \text{pool}\left(\left \{ \sigma_{i}^{D}(z_{j}) \right \}_{j=1}^N \right) &\forall i \in [K]\\
    &\hat{z}_i^C = \text{pool}\left( \left\{ (z^C_{i})_{j} \right\}_{j=1}^N \right) &\forall i \in [L].
\end{align}
Where the $\text{pool}(\cdot)$ operation can be any permutation invariant mapping. In our attack we use median pooling.

\begin{figure}
	\centering
	\includegraphics[trim={0 6.2cm 0 0},clip, width=0.48\textwidth]{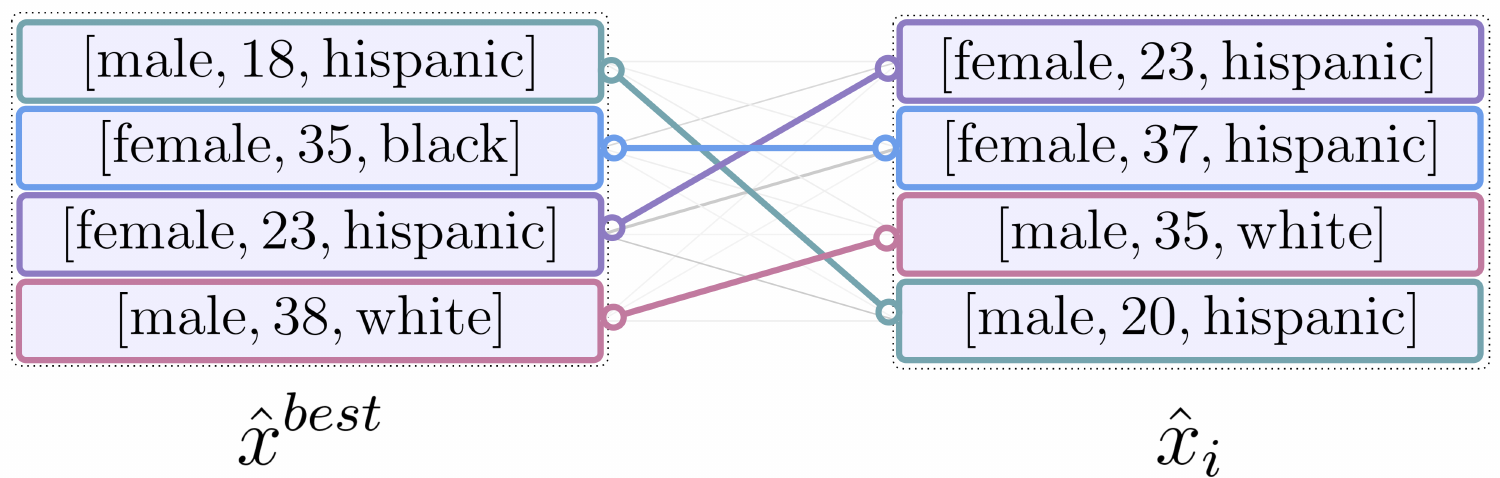}
	\vspace{-1.6em}
	\caption{Maximum similarity matching of a sample $\hat{x}_i$ of batch size $4$ from the collection of reconstructions to the best-loss sample $\hat{x}^{best}$.}
	\vspace{-1.5em}
	\label{fig:matching_figure}
\end{figure}

However, the above equations can not be applied in a straight-forward manner as soon as we aim to reconstruct batches containing more than just a single data point. As the batch-gradient is an average of the per-sample gradients, when running the leakage attack we may retrieve the batch-points in a different order at every optimization instance. Hence, it is not immediately clear how we can combine the obtained samples; \ie we need to reorder each batch such that their rows match to each other, and only then we can pool. We reorder by first selecting the sample that produced the best reconstruction loss at the end of optimization $\hat{z}^{best}$, with projection $\hat{x}^{best}$. Then, we match the rows of every other sample in the collection with respect to $\hat{x}^{best}$. Concretely, we calculate the similarity (shown in \cref{eq:accuracy} in \cref{sec:results}) between each pair of rows of $\hat{x}^{best}$ and another sample $\hat{x}_{i}$ in the collection and find the maximum similarity reordering of the rows with the help of bipartite matching solved by the Hungarian algorithm~\citep{Kuhn1955}. This process is depicted in \cref{fig:matching_figure}. Repeating this for each sample, we reorder the entire collection with respect to the best-loss sample, effectively reversing the permutation differences in the independent reconstructions. Finally, we can apply feature-wise pooling for each row over the collection.

\subsection{Assessment via Entropy}
\label{subsec:entropy_uncertainty}

We now address challenge (ii), assessing reconstructions. To recap, it is close-to-impossible for an uninformed adversary to assess the quality of the obtained private sample when it comes to tabular data, as almost any reconstruction may constitute a credible data point when projected back to mixed discrete-continuous space. This challenge does not arise as prominently in the image (or text) domain, because one can easily judge by looking at a picture, if it is just noise or an actual image, as exemplified in \cref{fig:comparison}. To address this issue, we propose to estimate the reconstruction uncertainty by looking at the level of agreement over a certain feature for different reconstructions. Concretely, given a collection of leaked samples as in \cref{subsec:pooled_ensembling}, we can observe the distribution of each feature over the samples. Intuitively, if this distribution is "peaky", \ie concentrates the mass heavily on a certain value, then we can assume that the feature has been reconstructed correctly, whereas if there is high disagreement between the reconstructed samples, we can assume that this feature's recovered final value should not be trusted. We can quantify this by measuring the entropy of the feature distributions induced by the recovered samples.

\paragraph{Discrete Features}
Let $p(\hat{x}_i^D)_m \coloneqq \frac{1}{N}\, \text{Count}_j(\hat{x}_{ij}^D=m)$ be the relative frequency of projected reconstructions of the $i$-th discrete feature of value $m$ in the ensemble. Then, we can calculate the normalized entropy of the feature as $\Bar{H}^D_i = \frac{-1}{\log\,|\mathcal{D}_i|}\,\sum_{m=1}^{D_i} p(\hat{x}_i^D)_m\,\log\,p(\hat{x}_i^D)_m$. Note that the normalization allows for comparing features with different domain sizes, \ie it ensures that $\Bar{H}^D_i \in [0, 1]$, as $H(k) \in [0, \log|\mathcal{K}|]$ for any finite discrete random variable $k \in \mathcal{K}$.

\paragraph{Continuous Features}
In case of continuous features, we calculate the entropy by first making the standard assumption that the errors of the reconstructed continuous features follow a Gaussian distribution. As such, we first estimate the sample variance $\hat{\sigma}^2_i$ for the $i$-th continuous feature and then plug it in to calculate the entropy of the corresponding Gaussian: $H^C_i = \frac{1}{2} + \frac{1}{2}\,\log\, 2 \pi \hat{\sigma}^2_i$. Cross-feature comparability can be achieved by scaling all features, \eg standardization.

        \begin{algorithm}[H]
            \caption{\OurMethod{} against training by FedSGD}
            \label{algorithm:comb_attack}
            \begin{algorithmic}[1]
                \FUNCTION{\textsc{\textbf{SingleInversion}}\,(Neural Network: $f_{\theta}$, Client Gradient: $g(c(x), y)$, Reconstructed Labels: $\hat{y}$, Initial Reconstruction: $z_j^0$, Iterations: $T$, \# Discrete Features: $K$)} \label{algline:SingleInversion}
                \FOR{$t$ \textbf{in} $0, 1, \dots, T-1$}
                    \FOR{$k$ \textbf{in} $1, 2, \dots, K$}
                        \STATE $\sigma(z_{kj}^D) \leftarrow$ \textbf{softmax}($z_{kj}^D$)\alglinelabel{algline:softmax_prior}
                    \ENDFOR
                    \STATE $z_j^{t+1} \leftarrow z_j^{t} - \eta \, \nabla_{z} \mathcal{E}_{CS}(g(c(x), y), g(\sigma(z_j^t), \hat{y}))$\label{algline:update}
                \ENDFOR
                \STATE \textbf{return} $z_i^T$
                \ENDFUNCTION
                \STATE
                \FUNCTION{\textsc{\textbf{TabLeak}}\,(Neural Network: $f_{\theta}$, Client Gradient: $g(c(x), y)$, Reconstructed Labels: $\hat{y}$, Ensemble Size: $N$, Iterations: $T$, \# Discrete Features: $K$)}
                \STATE $\left \{ z_j^0 \right \}_{j=1}^N \sim \mathcal{U}_{[0, 1]^d}$\alglinelabel{algline:initialization}
                \FOR{$j$ \textbf{in} $1, 2, \dots, N$}\alglinelabel{algline:samples}
                    \STATE $z_j^T \leftarrow$ \textsc{\textbf{SingleInversion}}($f_{\theta}$, $g(c(x), y)$, $\hat{y}$, $z_j^0$, $T$, $K$)
                \ENDFOR\alglinelabel{algline:end_optim}
                \STATE $\hat{z}^{best} \leftarrow \argmin_{z_j^T} \, \mathcal{E}_{CS}(g(c(x), y), g(\sigma(z_j^T), \hat{y}))$\alglinelabel{algline:select_lowest_loss}
                \STATE $\sigma(\hat{z}) \leftarrow$ \textsc{\textbf{MatchAndPool}}($\left \{ \sigma(z_j^T) \right \}_{j=1}^N, \hat{z}^{best}$)\alglinelabel{algline:match_and_pool}
                \STATE $\Bar{H}^D, H^C \leftarrow$ \textsc{\textbf{CalculateEntropy}}($\left \{ \sigma(z_j^T) \right \}_{j=1}^N$)\label{algline:entropy_calc}
                \STATE $\hat{x} \leftarrow$ \textsc{\textbf{Project}}($\sigma(\hat{z})$)\label{algline:project}
                \STATE \textbf{return} $\hat{x}$, $\bar{H}^D$, $H^C$
                \ENDFUNCTION
            \end{algorithmic}
        \end{algorithm}
\vspace{-2.1em}

\subsection{Combined Attack}
\label{subsec:combined_attack}
Following \citet{Geiping2020}, we use the cosine similarity loss as our reconstruction objective, defined as:
\vspace{-0.15em}
\begin{equation}
    \mathcal{E}_{CS}(z) \coloneqq 1-\frac{\langle g(c(x), y), \,g(\sigma(z), \hat{y}) \rangle}{\| g(c(x), y)\|_2\, \| g(\sigma(z), \hat{y}) \|_2},
\end{equation}
\vspace{-0.15em}where $(x, y)$ are the true data, $\hat{y}$ are the labels reconstructed beforehand, and we optimize for $z$.
Our end-to-end attack, \OurMethod{} is shown in~\cref{algorithm:comb_attack}.
First, we reconstruct the labels using the label reconstruction method of \citet{Geng2021} and input them into our attack. Then, we initialize $N$ independent dummy samples for an ensemble of size $N$ (Line~\ref{algline:initialization}). Starting from each initial sample we optimize independently (Lines~\ref{algline:samples}-\ref{algline:end_optim}) via the \textsc{SingleInversion} function. In each optimization step, we apply the softmax relaxation of \cref{subsec:softmax_trick}, and let the optimizer differentiate through it (Line~\ref{algline:softmax_prior}). After the optimization processes have reached the maximum number of allowed iterations $T$, we identify the sample $\hat{z}^{best}$ producing the best reconstruction loss (Line~\ref{algline:select_lowest_loss}). Using $\hat{z}^{best}$, we match and pool to obtain the final encoded reconstruction $\sigma(\hat{z})$ in Line~\ref{algline:match_and_pool} as described in \cref{subsec:pooled_ensembling}. Finally, we return the projected private data reconstruction $\hat{x}$ and the corresponding feature-entropies $\bar{H}^D$ and $H^C$, quantifying the uncertainty in the leaked sample.

\begin{table*}
	\caption{The mean inversion accuracy [\%] and standard deviation of different methods over varying batch sizes with given true labels (True $y$) and with reconstructed labels (Rec. $\hat{y}$) on the Adult dataset.}
	\label{table:rec_label_no_label}
	\centering
	\resizebox{1.85\columnwidth}{!}{
	\begin{tabular}{llcccccc}
		\toprule
		Label & Batch & \OurMethod{} & \OurMethod{} & \OurMethod{} & Inverting Gradients & Deep Gradient Leakage & Random\\
		& Size & & (no pooling) & (no softmax) & \citet{Geiping2020} & \citet{Zhu2019} &\\
		\midrule
		\multirow{5}{*}{True $y$}
		& $8$  & $\mathbf{95.2 \pm 8.8}$ &  $92.5 \pm 11.8$ &  $91.3 \pm 7.1$ & $91.1 \pm 7.3$ & $61.2 \pm 4.7$ & $53.9 \pm 4.4$\\
		& $16$  & $\mathbf{89.9 \pm 7.3}$ &  $85.3 \pm 9.7$ &  $79.0 \pm 4.0$ & $75.0 \pm 5.2$ & $60.2 \pm 3.3$ & $55.1 \pm 3.9$\\
		& $32$  & $\mathbf{79.3 \pm 4.5}$ &  $74.3 \pm 4.5$ &  $70.8 \pm 3.3$ & $66.6 \pm 3.5$ & $60.8 \pm 1.9$ & $58.0 \pm 2.9$\\
		& $64$  & $\mathbf{73.4 \pm 3.0}$  & $68.9 \pm 3.1$ & $67.3 \pm 3.2$ & $62.5 \pm 3.1$ & $61.3 \pm 1.4$ & $59.0 \pm 3.2$\\
		& $128$  & $\mathbf{71.4 \pm 1.2}$ & $67.4 \pm 1.4$ &  $65.2 \pm 2.1$ &   $59.5 \pm 2.1$ & $62.9 \pm 1.0$ & $61.2 \pm 3.1$\\
		\midrule
		\multirow{5}{*}{Rec. $\hat{y}$}
		& $8$  & $\mathbf{86.7 \pm 12.2}$ &  $83.8 \pm 13.6$ &  $82.7 \pm 10.5$ & $83.3 \pm 9.7$ & $56.1 \pm 5.4$ & $53.9 \pm 4.4$\\
		& $16$  & $\mathbf{83.0 \pm 7.7}$ &  $78.6 \pm 8.1$ &  $76.4 \pm 5.4$ & $73.0 \pm 3.5$ & $57.2 \pm 3.4$ & $55.1 \pm 3.9$\\
		& $32$  & $\mathbf{76.9 \pm 4.8}$  & $72.4 \pm 4.8$ & $68.9 \pm 4.2$ & $66.3 \pm 3.4$ & $58.4 \pm 2.5$ & $58.0 \pm 2.9$\\
		& $64$  & $\mathbf{72.8 \pm 3.3}$ & $68.5 \pm 3.5$ & $66.8 \pm 2.9$ & $63.1 \pm 3.2$ & $60.1 \pm 1.7$ & $59.0 \pm 3.2$\\
		& $128$ & $\mathbf{71.4 \pm 1.3}$ & $67.5 \pm 1.5$ & $65.0 \pm 2.2$ & $59.5 \pm 2.1$ & $62.3 \pm 1.0$ & $61.2 \pm 3.1$\\
		\bottomrule
	\end{tabular}
	}
    \vspace{-0.3em}
\end{table*}

\section{Experimental Evaluation}
\label{sec:results}

In this section, we first detail the evaluation metric we used to assess the obtained reconstructions and explain our experimental setup. Then, we evaluate our attack in various settings against prior methods, establishing a new state-of-the-art, while uncovering the significant vulnerability of tabular FL. Next, we demonstrate the effectiveness of our entropy-based uncertainty quantification method. Finally, we test our attack on varying architectures, over federated training, and against a defense mechanism. Our code is available at: \url{https://github.com/eth-sri/tableak}.

\paragraph{Evaluation Metric}
As no prior work on tabular data leakage exists, we propose a metric for measuring the reconstruction accuracy, inspired by the 0-1 loss, allowing the joint treatment of categorical and continuous features. For a reconstruction $\hat{x}$, we define the accuracy as:
\begin{equation}\label{eq:accuracy}
\begin{aligned}
    \text{accuracy}&(x, \,\hat{x}) \coloneqq \frac{1}{K+L} \Bigg( \sum_{i=1}^{K} \mathbb{I}\{x^D_i = \hat{x}^D_i\} \\&+ \sum_{i=1}^{L} \mathbb{I}\{\hat{x}^C_i \in [x^C_i-\epsilon_i,\, x^C_i+\epsilon_i]\} \Bigg),
\end{aligned}
\end{equation}
where $x$ is the ground truth and $\{\epsilon_i\}_{i=1}^L$ are constants determining how close the reconstructed continuous features have to be to the original value in order to be considered a privacy breach. We provide more details on our metric in \cref{appendix:accuracy_metric} and experiments with additional metrics in \cref{appendix:continuous_feature_recosntruction_measured_by_rmse}.

\paragraph{Baselines}
We consider two established prior attacks; the seminal work on gradient inversion of \citet{Zhu2019}, Deep Gradient Leakage, and the more recent strong attack of \citet{Geiping2020}, Inverting Gradients. For a fair comparison, we provide the labels to both attacks in the same manner as we do for \OurMethod{} and remove all non-tabular domain-specific elements from the attacks (\ie image priors).
Additionally, we also compare against random guessing. Here, ignoring the gradient updates, we randomly sample reconstructions from the per-feature marginals of the input dataset. To obtain the 1-way marginals, first we uniformly discretize the continuous features in $100$ bins, and then estimate the marginals of all features by counting. Although this baseline is usually not realizable in practice (as it assumes knowledge of the marginals), it is imperative to compare against it, as performing below this baseline signals that no private information is extracted from the client updates. Note that as both the selection of a batch and the random baseline represent sampling from the (approximate) data distribution, the random baseline monotonously increases in accuracy with growing batch size.

\paragraph{Experimental Setup}
For all attacks, we use the Adam optimizer~\citep{Kingma2014} with learning rate $0.06$ for $1\,500$ iterations and without a learning rate schedule to perform the optimization in \cref{algorithm:comb_attack}. Unless stated otherwise, we attack a fully connected neural network (NN) at initialization with two hidden layers of $100$ neurons each. All experiments were carried out on four popular mixed-type tabular binary classification datasets, the Adult census dataset~\citep{Dua:2019}, the German Credit dataset~\citep{Dua:2019}, the Lawschool Admission dataset~\citep{wightman1998lsac}, and the Health Heritage dataset from \citet{healthheritage}. Due to the space constraints, here we report only our results on the Adult dataset, and refer the reader to \cref{appendix:all_main_results} for full results on all four datasets. Finally, for all reported numbers below, we estimate the mean and standard deviation of each reported metric on $50$ randomly sampled batches. For further details on the experimental setup, we refer the reader to \cref{appendix:further_experimental_details}. For experiments with varying network sizes and attacks against defenses, please see \cref{appendix:further_experiments}. 

\begin{table*}
	\caption{Mean and standard deviation of the inversion accuracy [\%] on FedAvg with local dataset sizes of $32$ on the Adult dataset. The accuracy of the random baseline for $32$ data points is $57.7 \pm 3.6$.}
	\label{table:rec_fedavg}
	\centering
	\resizebox{1.65\columnwidth}{!}{
	\begin{tabular}{lcccccc}
		\toprule
		 & \multicolumn{3}{c}{\OurMethod{}} & \multicolumn{3}{c}{Inverting Gradients \citep{Geiping2020}}\\
		\cmidrule[1pt](l{5pt}r{5pt}){2-4}\cmidrule[1pt](l{5pt}r{5pt}){5-7}
		n. batches & 1 epoch & 5 epochs & 10 epochs & 1 epoch & 5 epochs & 10 epochs \\
		\midrule
		
		 1 & $\mathbf{80.7 \pm 3.8}$ & $\mathbf{75.8 \pm 3.3}$ & $\mathbf{72.8 \pm 3.2}$ & $65.2 \pm 2.7$ & $56.1 \pm 4.1$ & $53.1 \pm 4.2$ \\
		 2 & $\mathbf{79.2 \pm 4.2}$ & $\mathbf{75.6 \pm 2.7}$  & $\mathbf{73.1 \pm 5.0}$ & $64.8 \pm 3.3$ & $56.4 \pm 4.8$ & $56.2 \pm 4.8$ \\
		 4 & $\mathbf{79.7 \pm 3.6}$ & $\mathbf{76.2 \pm 3.0}$ & $\mathbf{73.7 \pm 3.6}$ & $64.8 \pm 3.4$ & $58.7 \pm 4.6$ & $56.6 \pm 5.0$ \\
		\bottomrule
	\end{tabular}
	}
	\vspace{-0.6em}
\end{table*}

\paragraph{General Results against FedSGD}
In \cref{table:rec_label_no_label} we present the results of \OurMethod{} against FedSGD training, together with two ablation experiments, each time removing either the pooling (no pooling) or the softmax component (no softmax). We compare our results to the baselines introduced above, on batch sizes $8, 16, 32, 64$, and $128$, once assuming knowledge of the true labels (top) and once using labels reconstructed by the method of \citet{Geng2021} (bottom). Notice that the noisy label reconstruction only influences the results for lower batch sizes, and manifests itself mostly in higher variance in the results. Further, we find that for batch size $8$ (and lower, see \cref{appendix:all_main_results}) most attacks reveal close to all the private data, exposing a trivial vulnerability of tabular FL of high concern. In case of larger batch sizes, even up to $128$, \OurMethod{} can uncover a significant portion of the client's private data, well above random guessing, while the attacks of \citet{Zhu2019} and \citet{Geiping2020} fail to do so, demonstrating the necessity of a domain tailored attack when evaluating the privacy threat.
Further, the results on the ablation attacks demonstrate the effectiveness of each attack component, both providing non-trivial improvements over the baseline attacks that are preserved when combined in \OurMethod{}. Demonstrating generalization beyond Adult, we include our results on the German Credit, Lawschool Admissions, and Health Heritage datasets in \cref{appendix:full_fedsgd_results_on_all_datasets}, where we also improve on the state-of-the-art by at least $12.7\% - 14.5\%$ on batch size $32$ on each dataset and up to $21.8\%$ on other batch sizes. As Inverting Gradients~\citep{Geiping2020} dominates Deep Gradient Leakage~\citep{Zhu2019} in almost all settings in \cref{table:rec_label_no_label}, we omit the attack of \citet{Zhu2019} from further comparisons.

\begin{figure}
	\centering
	\includegraphics[width=0.42\textwidth]{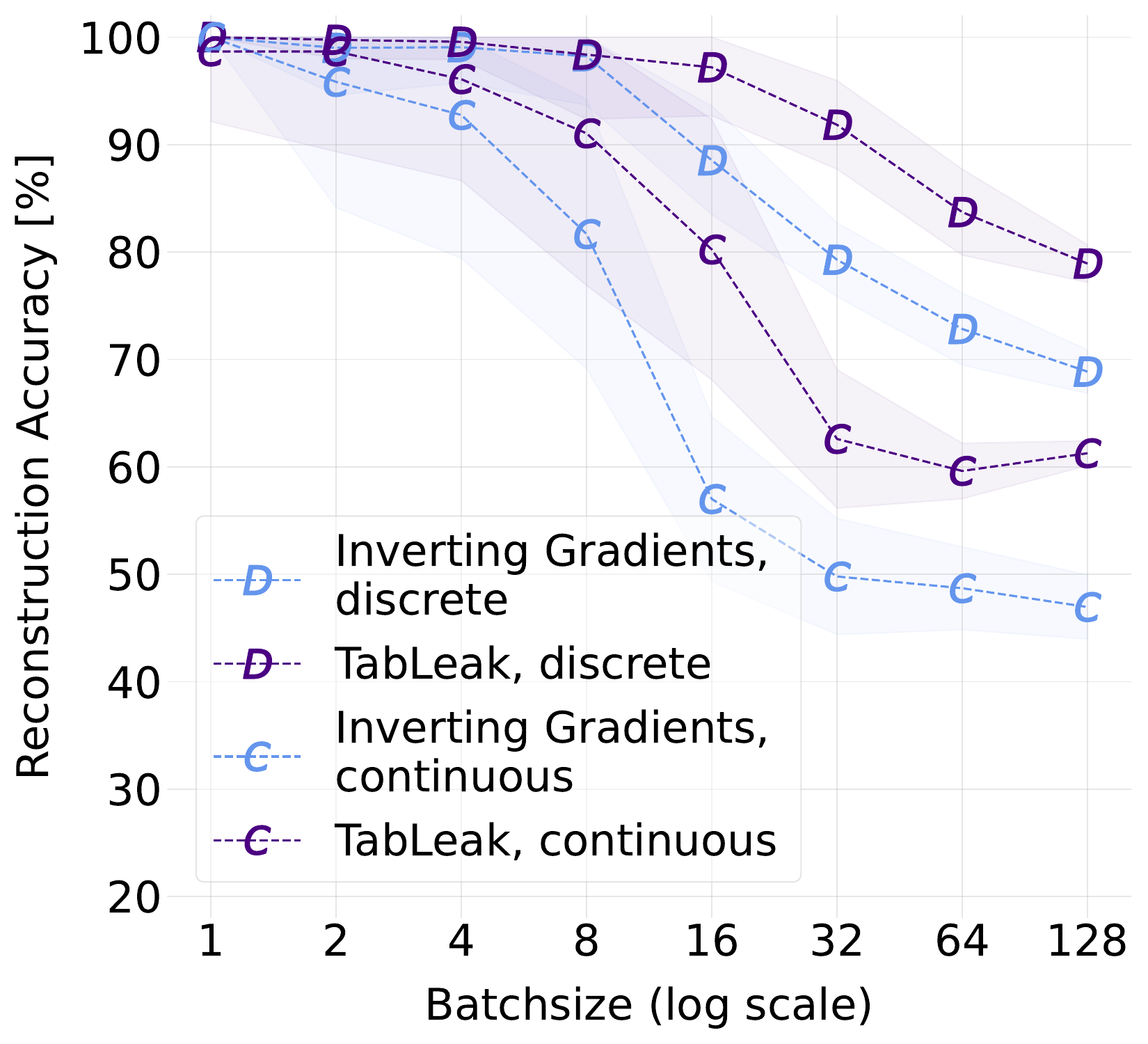}
	\caption[short]{The inversion accuracy on the Adult dataset over varying batch size separated for discrete (D) and continuous (C) features.}
	\vspace{-0.6em}
	\label{fig:attack_errors_adult_cat_cont}
\end{figure}

\paragraph{Categorical vs. Continuous Features}
An important consequence of having mixed type features is that the attack success clearly differs by feature type. As we can observe in \cref{fig:attack_errors_adult_cat_cont}, the continuous features exhibit an up to $30\%$ lower accuracy than the categorical features for the same batch size. We suggest that this is due to the discrete nature of categorical features and their encoding. While trying to match the gradients by optimizing the reconstruction, having the correct categorical features will have a greater effect on the gradient alignment, as when encoded, they take up the majority of the data vector. Also, when reconstructing a one-hot encoded categorical feature, we only have to be able to retrieve the location of the maximum in a vector of length $D_i$, whereas for the successful reconstruction of a continuous feature we have to retrieve its value correctly up to a small error. Therefore, especially when the optimization process is aware of the discrete structure (\eg by using the softmax relaxation), categorical features are much easier to attack. This finding of ours uncovers a critical privacy risk in tabular federated learning, as sensitive features are often categorical, \eg gender, race, or STI test results.

\begin{table*}
	\caption{The mean and the standard deviation of the attack accuracy [\%] over different architectures inverting batches of size $32$. The random baseline at this batch size is $58.0 \pm 2.9$.}
	\label{table:varying_architectures}
	\centering
	\resizebox{1.99\columnwidth}{!}{
	\begin{tabular}{lccccc}
		\toprule
		Attack & Linear & FC NN & FC NN large & CNN (BN) & ResNet (BN)\\
		\midrule
		 Inverting Gradients~\citep{Geiping2020} & $55.3 \pm 1.7$ & $66.6 \pm 3.5$ & $89.2 \pm 3.8$ & $43.4 \pm 2.0$ & $61.7 \pm 4.0$ \\
		 GradInversion~\citep{Yin2021} & $\mathbf{61.3 \pm 2.7}$ & $67.7 \pm 2.5$  & $88.0 \pm 3.0$ & $72.8 \pm 2.7$ & $67.6 \pm 2.6$ \\
		 \OurMethod{} & $44.5 \pm 1.9$ & $\mathbf{79.3 \pm 4.5}$ & $\mathbf{89.6 \pm 8.3}$ & $\mathbf{83.7 \pm 2.7}$ & $\mathbf{71.4 \pm 9.2}$ \\
		\bottomrule
	\end{tabular}
	}
	\vspace{-0.6em}
\end{table*}

\begin{table}
	\caption{The mean and standard deviation of the accuracy [\%] of each feature type in the top $25\%$ and the bottom $25\%$ ranked in the batch according to entropy (lowest on top).}
	\centering
	\resizebox{1.0\columnwidth}{!}{
	\begin{tabular}{lcccc}
		\toprule
		 Batch & \multicolumn{2}{c}{Categorical} & \multicolumn{2}{c}{Continuous}\\
		\cmidrule[1pt](l{5pt}r{5pt}){2-3}\cmidrule[1pt](l{5pt}r{5pt}){4-5}
		Size  & Top $25$\% & Bottom $25\%$ & Top $25\%$ & Bottom $25\%$ \\
		\midrule
		 8 & $99.7 \pm 2.3$ & $96.0 \pm 10.3$ & $97.2 \pm 9.4$ & $84.2 \pm 21.6$\\
		 16 & $99.9 \pm 0.6$ & $89.8 \pm 13.0$ & $98.2 \pm 3.5$ & $60.8 \pm 19.9$\\
		 32 & $99.1 \pm 2.6$ & $75.5 \pm 8.0$ & $94.2 \pm 4.7$ & $43.6 \pm 8.2$\\
		 64 & $97.8 \pm 2.6$ & $66.1 \pm 5.3$ & $92.9 \pm 3.5$ & $41.3 \pm 5.8$\\
		 128 & $94.3 \pm 1.9$ & $62.8 \pm 3.8$ & $93.5 \pm 2.3$ & $42.2 \pm 3.7$\\
		\bottomrule
	\end{tabular}}
	\vspace{-0.9em}
	\label{table:top_bottom_25}
\end{table}

\paragraph{Federated Averaging}
In training with FedAvg~\citep{McMahan2016} participating clients conduct local training of several updates before communicating their new parameters to the server. Note that the more local updates are conducted by the clients, the harder a leakage attack becomes, making FedAvg to be regarded as a more secure protocol. Although this training method is of significantly higher practical importance than FedSGD, most prior work does not evaluate against it. Transferring \OurMethod{} into the framework of~\citet{Dimitrov22} (for details please see \cref{appendix:further_experimental_details} and the work of \citet{Dimitrov22}), we evaluate our attack and the strong baseline attack of \citet{Geiping2020} in the setting of Federated Averaging. We present our results of retrieving a client dataset of size $32$ over varying number of local batches and epochs on the Adult dataset in \cref{table:rec_fedavg}, while assuming full knowledge of the true labels. We observe that our combined attack significantly outperforms the random baseline of $57.7\%$ accuracy even up to $40$ local updates, and in some cases beating the baseline attack of \citet{Geiping2020} by almost $20\%$. Meanwhile, the baseline attack fails to consistently outperform random guessing whenever the local training is longer than one epoch. This shows that non-domain-tailored attacks are not sufficient to uncover relevant vulnerabilities, risking an illusion of privacy. As FedAvg with tabular data is of high practical relevance, our results of successful attacks are concerning. Further details of the experimental setup and results on other datasets can be found in \cref{appendix:further_experimental_details} and \cref{appendix:all_main_results}, respectively.

\paragraph{Assessing Reconstructions via Entropy}
We demonstrate the effectiveness of our assessment mechanism, looking at reconstructions from \OurMethod{} and their corresponding feature entropies. For a reconstructed batch, we rank each per-sample feature (\ie all features over all rows in a single batch, distinguishing discrete and continuous features) according to decreasing entropy (lowest on top). Then, we take the top $25\%$ of the features of lowest entropy and the bottom $25\%$ from the resulting ranking, and report their accuracy for varying batch sizes in \cref{table:top_bottom_25}. The results in \cref{table:top_bottom_25} confirm our expectation that features with low uncertainty score tend to be reconstructed much better than those exhibiting high entropy, asserting that our assessment mechanism is effective in separating well reconstructed data from poorly reconstructed, exposing an accuracy gap of up to $50\%$ in the retrieved features. In fact, the top quarter of all features is reconstructed at an accuracy well above $90\%$ on all batch sizes. Note that this ranking is done \emph{without} any knowledge of the true data, hence it is accessible to a real adversary. This shows that even reconstructions of lower overall accuracy (\eg $71.4\%$ on batch size 128, see \cref{table:rec_label_no_label}) only provide a false sense of privacy, as an adversary can still extract correct data with high confidence, resulting in a strong breach of privacy even at large batch sizes, previously deemed as safe.
We include results on all four datasets in \cref{appendix:full_results_on_entropy_on_all_datasets}, leading to analogous conclusions.

\paragraph{Attack Performance and Model Architecture}
To assess how the attack success of our and competing methods depends on the chosen architecture, in this experiment we attack five different models, reconstructing batches of size $32$: a logistic regressor (Linear), the two-layer NN used in prior experiments (FC NN), a large-three layer NN with $400$ neurons in each layer (FC NN large), a convolutional NN (CNN) with a batch normalization (BN) layer, and a fully connected NN with residual connections and two BN layers (ResNet). For this experiment we also introduce an additional baseline, GradInversion~\citep{Yin2021}, which is an attack tailored to image data, and was designed to overcome the limitations of BN layers. Note that this attack operates under strictly stronger assumptions than \OurMethod{}, as it requires knowledge of the BN statistics, which \citet{huang2021evaluating} argue is unrealistic. To adapt GradInversion to tabular data, we change the original reconstruction loss of squared error to cosine similarity, as it leads to better results in this domain. Additionally, as this attack requires the selection of several additive prior parameters, we evaluate the attack on a grid, and report only the best results here, providing an unrealistic advantage to this baseline. Our results are shown in \cref{table:varying_architectures}. We make several important insights on the influence of the architecture on the attack success: (i) \OurMethod{} is the strongest overall attack, even when confronted with BN layers, (ii) larger networks are significantly more vulnerable to all attacks, (iii) linear models are seemingly unbreakable under realistic assumptions. Therefore, while BN layers might only provide a false sense of privacy, it is imperative to consider models of limited size to enhance the privacy of FL.

\paragraph{Impact of Network Training}
Before, we only attacked networks at initialization. To examine how the attack success depends on the training state of the network, we evaluate \OurMethod{} and the baseline attack of \citet{Geiping2020} attacking over the first $15$ epochs of federated training with batch size $32$. We report our results in \cref{table:attack_training_iters}. As expected, the attack performance gradually decreases as the network is fitted to the data, a phenomenon already known from prior works~\citep{Geiping2020,Dimitrov22}. However, \OurMethod{} maintains a strong performance further into the training, and preserves its advantage over the baseline attack consistently. Further, note that decreasing attack performance over training is of limited practical relevance, as nothing prevents the server from attacking in the early stages of training where the model is vulnerable, breaching the privacy of the participating clients.
\begin{table}
	\caption{The mean and standard deviation of the attack accuracy [\%] attacking a network at the $1^{\text{st}}$, $5^{\text{th}}$, $10^{\text{th}}$, and $15^{\text{th}}$ epoch of FedSGD training on batch size $32$.}
	\label{table:attack_training_iters}
	\centering
	\resizebox{0.85\columnwidth}{!}{
	\begin{tabular}{lcc}
		\toprule
		Training & \OurMethod{} & Inverting Gradients \\
		Epochs & & \citet{Geiping2020} \\
		\midrule
		$1$  & $\mathbf{79.1 \pm 4.2}$ &  $67.8 \pm 2.1$  \\
		$5$  & $\mathbf{76.4 \pm 5.7}$ &   $64.5 \pm 3.8$  \\
		$10$  & $\mathbf{74.5 \pm 5.7}$  & $60.9 \pm 3.7$ \\
		$15$  & $\mathbf{64.5 \pm 7.1}$  & $57.8 \pm 4.0$ \\
		\bottomrule
	\end{tabular}
	}
\end{table}

\paragraph{Defending with Noise}
As the undefended systems attacked above are critically vulnerable, we evaluate the effectiveness of a common defense method against gradient leakage attacks~\citep{Zhu2019,Dimitrov2022,Dimitrov22}. To defend against an honest-but-curious server the clients add Gaussian noise of zero mean and fixed scale to the gradients before communicating them to the server. Although this defense is inspired by differential privacy (DP)~\cite{Dwork2006}, due to no clipping, it does not provide theoretical guarantees. We test this defense method against \OurMethod{} and the baseline attack of \citet{Geiping2020} on varying noise scales at batch size $32$, and report our results in \cref{table:attack_noise}. Encouragingly, we can observe that using the right amount of noise poses a viable defense against leakage attacks on the Adult dataset at this batch size, reducing the attack accuracy, while having a $-0.5\%$ impact on the task accuracy of the NN trained on the noisy gradients. In \cref{appendix:attack_against_gaussian_dp}, we present our results against this defense on all four datasets and over varying batch sizes, where we observe that smaller batch sizes require more noise to reduce vulnerability.

\begin{table}
	\caption{The mean and standard deviation of the attack accuracy [\%] for Gaussian noise of varying scale added to the gradients. The right column reports the task accuracy [\%] of the NN trained with perturbed gradients.}
	\label{table:attack_noise}
	\centering
	\resizebox{0.99\columnwidth}{!}{
	\begin{tabular}{lccc}
		\toprule
		Noise & \OurMethod{} & Inverting Gradients & Network\\
		Scale & & \citet{Geiping2020} & Accuracy\\
		\midrule
		$0.0$      & $\mathbf{79.3 \pm 4.5}$ & $66.6 \pm 3.5$ & $84.6 \pm 0.1$\\
		$10^{-3}$  & $\mathbf{75.4 \pm 3.8}$ &  $64.1 \pm 3.2$  & $84.5 \pm 0.2$\\
		$10^{-2}$  & $\mathbf{58.0 \pm 2.3}$ &   $46.6 \pm 2.8$  & $84.4 \pm 0.2$\\
		$10^{-1}$  & $\mathbf{41.3 \pm 2.9}$  & $38.6 \pm 2.2$ & $84.1 \pm 0.2$\\
		\bottomrule
	\end{tabular}
	}
\end{table}

\section{Discussion}
\label{sec:discussion}
Using \OurMethod{}, we showed that an honest-but-curious FL server can reconstruct large amounts of private data with little effort, even in setups previously deemed as safe, such as large batch sizes and FedAvg. In particular, using our uncertainty quantification scheme, we can reconstruct a quarter of all features in a large batch of size $128$ at $>93\%$ accuracy. In a practical example, assuming all adults in the US have a bank account ($\approx 260$ million) at banks cooperating in FL, at least $65$ million people would be affected by a potential attack leaking their information with high confidence, deeming it the sixth-largest financial data breach in history~\citep{databreaches}.
Further, our discovery of the disproportionate vulnerability of discrete features argues for targeted mitigation, \eg by exploring alternative, safer representations.

Our results raise great concerns about current industrial FL systems potentially employed by financial, healthcare, or other institutions managing privacy-critical tabular data. Individuals' trust in such institutions when handing over sensitive information is fundamental in allowing for effective operation, to which end various legal institutions have been established, \eg \emph{barrister secrecy}, \emph{medical secrecy}, or latest, \emph{GDPR}. Apart from the damage inflicted on individuals whose private information may be abused by adversaries, the potential long-term loss of trust in institutions could lead to a wider impact on the services they are able to provide. %

\paragraph{Defenses}
In addition to our attacks on \emph{undefended} systems uncovering the excessive intrinsic risk in tabular FL, we showed in a promising experiment how a DP-inspired defense of adding noise to the communicated gradients can be leveraged to mitigate this risk. While this defense appears as effective in our experimental setup, in practice, the large associated cost in utility makes for limited applicability~\citep{Jayaraman2019EvaluatingDP} of DP methods. Therefore, it is necessary that further theoretically principled approaches are pursued in defending against data leakage attacks in FL.

\section{Conclusion}
\label{sec:conclusion}
In this work, we explored data leakage in tabular FL using \OurMethod{}, the first data leakage attack on tabular data, obtaining state-of-the-art results against both popular FL training protocols, and uncovering the excessive vulnerability of tabular FL, breaking several setups previously thought of as safe. As tabular data is ubiquitous in privacy critical applications, our results raise important concerns regarding practical systems currently using FL. Therefore, we advocate for further research on advancing provable defenses.

\paragraph{Acknowledgements}
This work has received funding from the Swiss State Secretariat for Education, Research and Innovation (SERI) (SERI-funded ERC Consolidator Grant).

\bibliography{references_.bib}
\bibliographystyle{icml2023}

\clearpage
\appendix
\onecolumn
\section{Accuracy Metric}
\label{appendix:accuracy_metric}
To ease the understanding, we start by repeating our accuracy metric here, where we measure the reconstruction accuracy between the retrieved sample $\hat{x}$ and the ground truth $x$ as:
\begin{equation}
    \text{accuracy}(x, \,\hat{x}) \coloneqq \frac{1}{K+L} \left (\sum_{i=1}^{K} \mathbb{I}\{x^D_i = \hat{x}^D_i\}+ \sum_{i=1}^{L} \mathbb{I}\{\hat{x}^C_i \in [x^C_i-\epsilon_i,\, x^C_i+\epsilon_i]\}\right).
\end{equation}
Note that the binary treatment of continuous features in our accuracy metric enables the combined measurement of the accuracy on both the discrete and the continuous features. From an intuitive point of view, this measure closely resembles how one would judge the correctness of numerical guesses. For example, guessing the age of a $25$ year old, one would deem the guess good if it is within $3$ to $4$ years of the true value, but the guesses $65$ and $87$ would be both qualitatively incorrect. In order to facilitate scalability of our experiments, we chose the $\left \{ \epsilon_i \right \}_{i=1}^L$ error-tolerance-bounds based on the global standard deviation if the given continuous feature $\sigma^C_i$ and multiplied it by a constant, concretely, we used $\epsilon_i = 0.319 \, \sigma^C_i$ for all our experiments. Note that $Pr[\mu - 0.319\,\sigma < x < \mu + 0.319\, \sigma] \approx 0.25$ for a Gaussian random variable $x$ with mean $\mu$ and variance $\sigma^2$. For our metric this means that assuming Gaussian zero-mean error in the reconstruction around the true value, we accept our reconstruction as privacy leakage as long as we fall into the $25\%$ error-probability range around the correct value. In \cref{table:tolerance_bounds} we list the tolerance bounds $\epsilon_i$ for the continuous features of the Adult dataset produced by this method. We would like to remark here, that we fixed our metric parameters before conducting any experiments, and did not adjust them based on any obtained results. Note also that in \cref{appendix:further_experiments} we provide results where the continuous feature reconstruction accuracy is measured using the commonly used regression metric of root mean squared error (RMSE), where \OurMethod{} also achieves the best results, signaling that the success of our method is independent of our chosen metric.

\begin{table}[H]
	\caption{Resulting tolerance bounds on the Adult dataset when using $\epsilon_i = 0.319\,\sigma_i^C$, as used by us for our experiments.}
	\label{table:tolerance_bounds}
	\centering
	\resizebox{0.97\columnwidth}{!}{
	\begin{tabular}{lcccccc}
		\toprule
		 feature & \texttt{age} & \texttt{fnlwgt} & \texttt{education-num} & \texttt{capital-gain} & \texttt{capital-loss} & \texttt{hours-per-week}\\
		\midrule
		tolerance & 4.2 & 33699 & 0.8 & 2395 & 129 & 3.8 \\
		\bottomrule
	\end{tabular}
	}
\end{table}

\section{Further Experimental Details}
\label{appendix:further_experimental_details}
Here we give an extended description to our experimental details provided in \cref{sec:results}, additionally we provide the specifications of each used dataset in \cref{table:datasets}. For all attacks, we use the Adam optimizer~\citep{Kingma2014} with learning rate $0.06$ for $1\,500$ iterations and without a learning rate schedule. We chose the learning rate based on our experiments on the baseline attack where it performed best. In line with \citet{Geiping2020}, we modify the update step of the optimizer by reducing the update gradient to its element-wise sign. We attack a fully connected neural network with two hidden layers of $100$ neurons each at initialization. However, we provide a network-size ablation in \cref{fig:network_size_all}, where we evaluate our attack against the baseline method for $5$ different network architectures. For each reported metric we conduct $50$ independent runs on 50 different batches to estimate their statistics. For all FedSGD experiments we clamp the continuous features to their valid ranges before measuring the reconstruction accuracy, both for our attacks and the baseline methods. Additionally, for \OurMethod{} and its ablation experiments, we encourage the continuous features to stay within bounds by wrapping them in a sigmoid function during optimization. Note that assuming some knowledge of the admissible ranges for the continuous columns is not unrealistic, as in most cases realistic ranges can be assumed based on the name of the feature column, especially if the feature standard deviations are known to the server (as in our case). For the network accuracy shown in the experiment "Defending with Noise", we train the two layer attacked network for 10 epochs at batch size 32 with learning rate 0.01 using mini-batch stochastic gradient descent. We ran each of our experiments on single cores of Intel(R) Xeon(R) CPU E5-2690 v4 @ 2.60GHz.

\begin{table}
	\caption{Dataset specifications.}
	\label{table:datasets}
	\centering
	\resizebox{0.99\columnwidth}{!}{
	\begin{tabular}{lccccc}
		\toprule
		 & Features & Discrete Features & Continuous Features & Encoded Features & Data Points\\
		\midrule
		Adult & $14$ & $8$ & $6$ & $105$ & $45\, 222$\\
		German & $20$ & $13$ & $7$ & $63$ & $1\, 000$\\
		Lawschool & $7$ & $5$ & $2$ & $39$ & $96\, 584$\\
		Health Heritage & $17$ & $6$ & $11$ & $110$ & $218\, 415$\\
		\bottomrule
	\end{tabular}
	}
\end{table}

\paragraph{Federated Averaging Experiments}
For experiments on attacking the FedAvg training algorithm, we fix the clients' local dataset size at $32$ and conduct an attack after local training with learning rate $0.01$ on the initialized network described above. We use the FedAvg attack-framework of \citet{Dimitrov22}, where for each local training epoch we initialize an independent mini-dataset matching the size of the client dataset, and simulate the local training of the client. At each reconstruction update, we use the mean squared error between the different epoch data means ($D_\text{inv} =\ell_2$ and $g=\text{mean}$ in \citet{Dimitrov22}) as the permutation invariant epoch prior required by the framework, ensuring the consistency of the reconstructed dataset. For the full technical details, please refer to the manuscript of \citet{Dimitrov22}. For choosing the prior parameter $\lambda_\text{inv}$, we conduct line-search on each setup and attack method pair individually on the parameters $[0.0, 0.5, 0.1, 0.05, 0.01, 0.005, 0.001]$, and pick the ones providing the best results. Further, to reduce computational overhead, we reduce the ensemble size of \OurMethod{} from $30$ to $15$ for these experiments on all datasets.

\section{Further Experiments}
\label{appendix:further_experiments}
In this subsection, we present several further experiments:
\begin{itemize}
	\item Results of attacking neural networks defended using differentially private noisy gradients in \cref{appendix:attack_against_gaussian_dp}.
	\item Ablation study on the impacts of the neural network's size on the reconstruction difficulty in \cref{appendix:varying_network_size}.
	\item Ablation study on the impact of the neural network's architecture on the reconstruction difficulty in \cref{appendix:varying_network_architecture}.
	\item Measuring the Root Mean Squared Error (RMSE) of the reconstruction of continuous features in \cref{appendix:continuous_feature_recosntruction_measured_by_rmse}.
	\item Testing \OurMethod{} and the baseline attack on a high-dimensional synthetic dataset in \cref{appendix:attacking_high_dimensional_datasets}.
	\item Ablation study on the impact of training on the attack difficulty in \cref{appendix:attacking_during_training}.
	\item Ablation study on the impact of number of attack iterations on the attack success in \cref{appendix:impact_of_attack_iterations}.
\end{itemize}

\subsection{Attack against Gaussian DP Defense}
\label{appendix:attack_against_gaussian_dp}
\begin{figure}
	\centering
	\begin{subfigure}{.33\textwidth}
		\centering
		\includegraphics[width=0.95\textwidth]{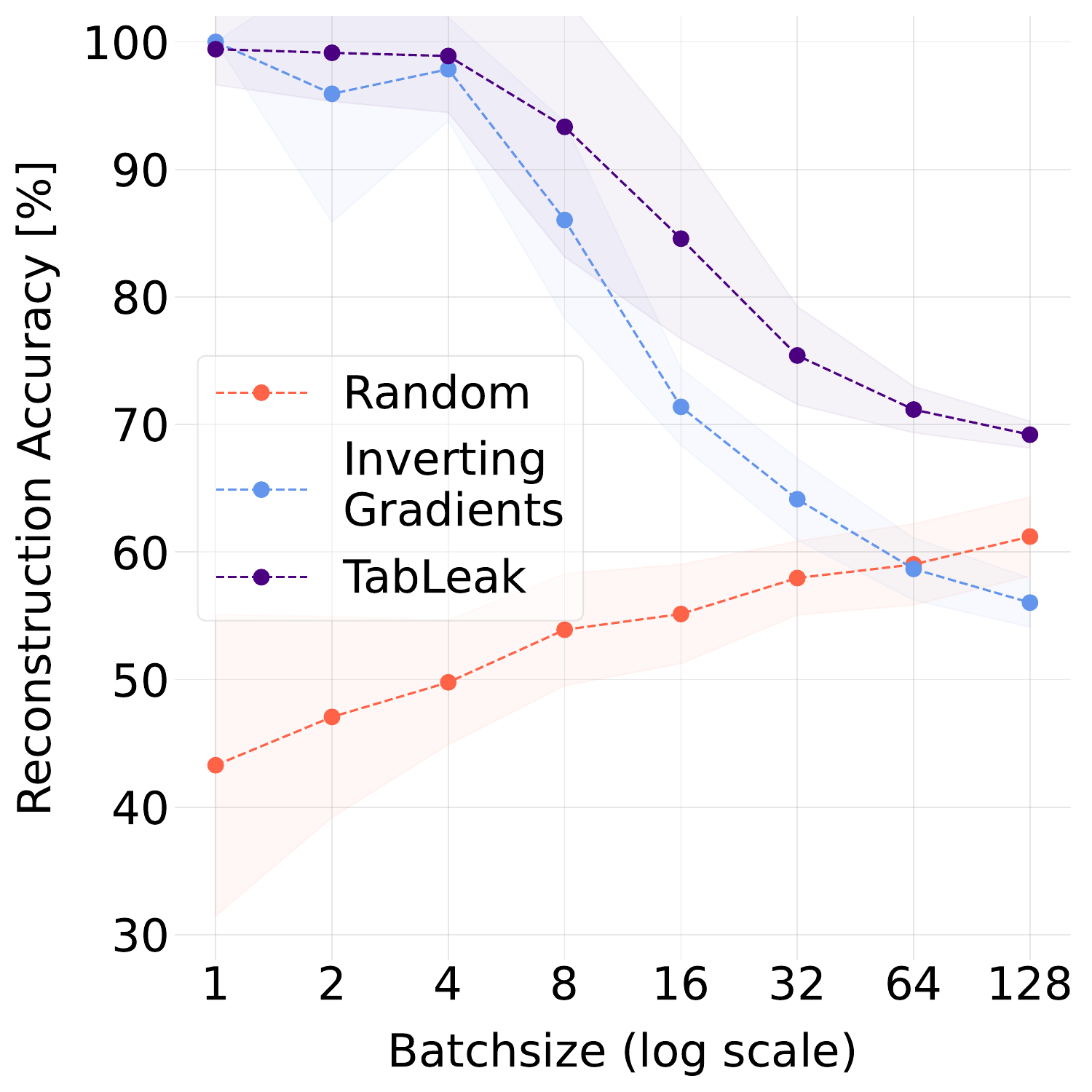}
		\subcaption{Noise standard deviation $= 0.001$}
	\end{subfigure}%
	\begin{subfigure}{.33\textwidth}
		\centering
		\includegraphics[width=0.95\textwidth]{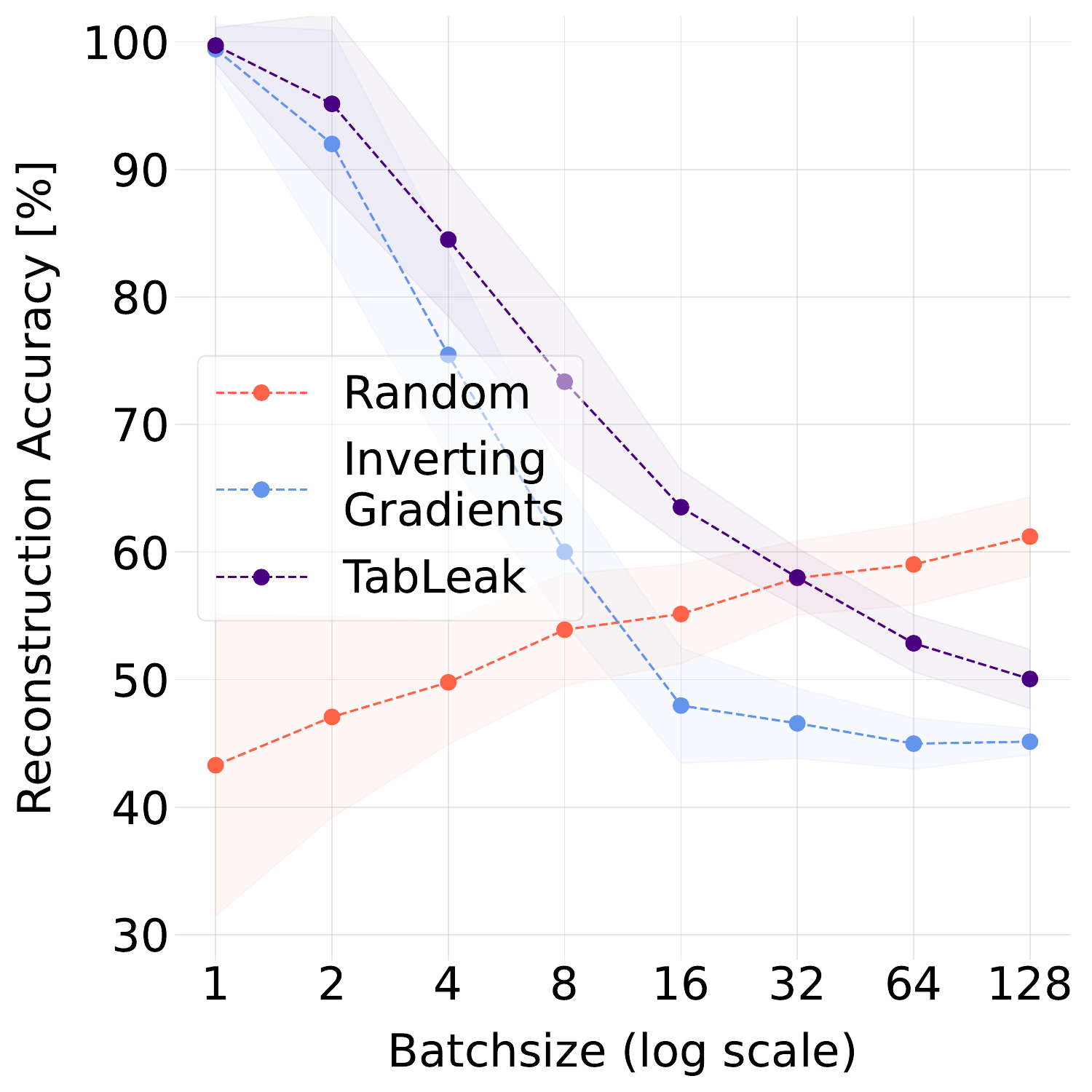}
		\subcaption{Noise standard deviation $= 0.01$}
	\end{subfigure}
	\begin{subfigure}{.33\textwidth}
		\centering
		\includegraphics[width=0.95\textwidth]{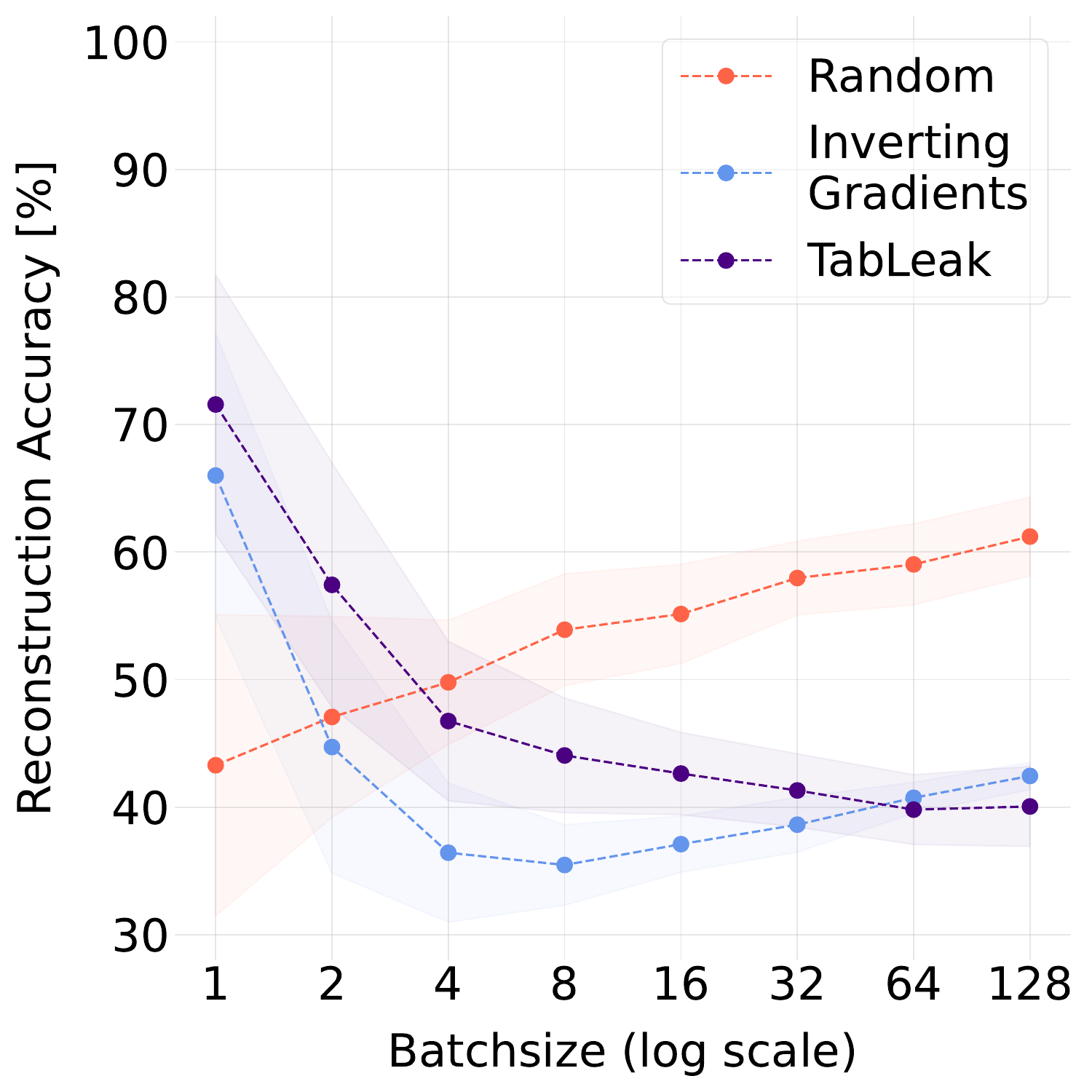}
		\subcaption{Noise standard deviation $= 0.1$}
	\end{subfigure}
	\caption{Mean and standard deviation accuracy [\%] curves over batch size at varying Gaussian noise level $\sigma$ added to the client gradients for differential privacy on the \textbf{Adult} dataset.}
	\label{fig:dp_adult}
\end{figure}

\begin{figure}
	\centering
	\begin{subfigure}{.33\textwidth}
		\centering
		\includegraphics[width=0.95\textwidth]{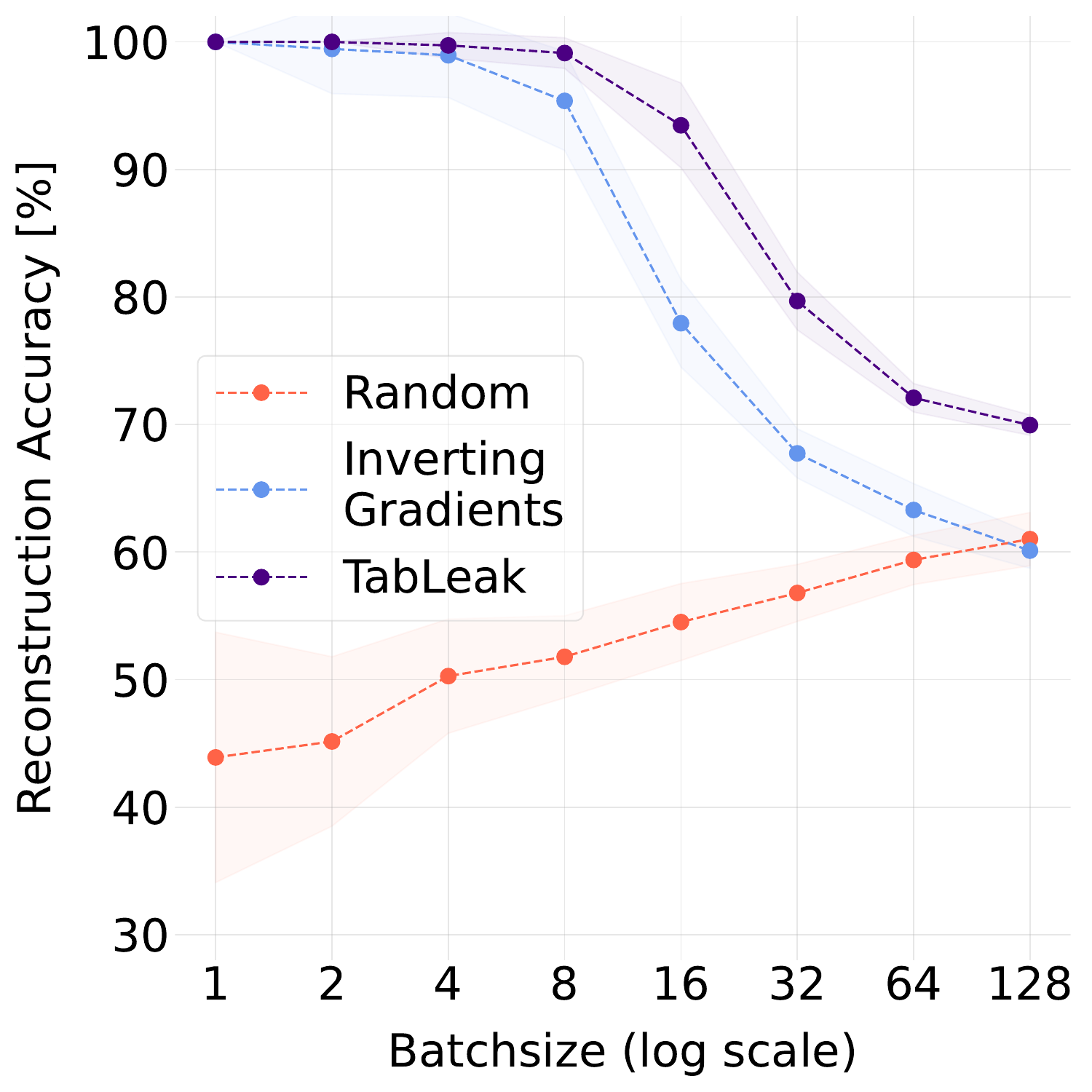}
		\subcaption{Noise standard deviation $= 0.001$}
	\end{subfigure}%
	\begin{subfigure}{.33\textwidth}
		\centering
		\includegraphics[width=0.95\textwidth]{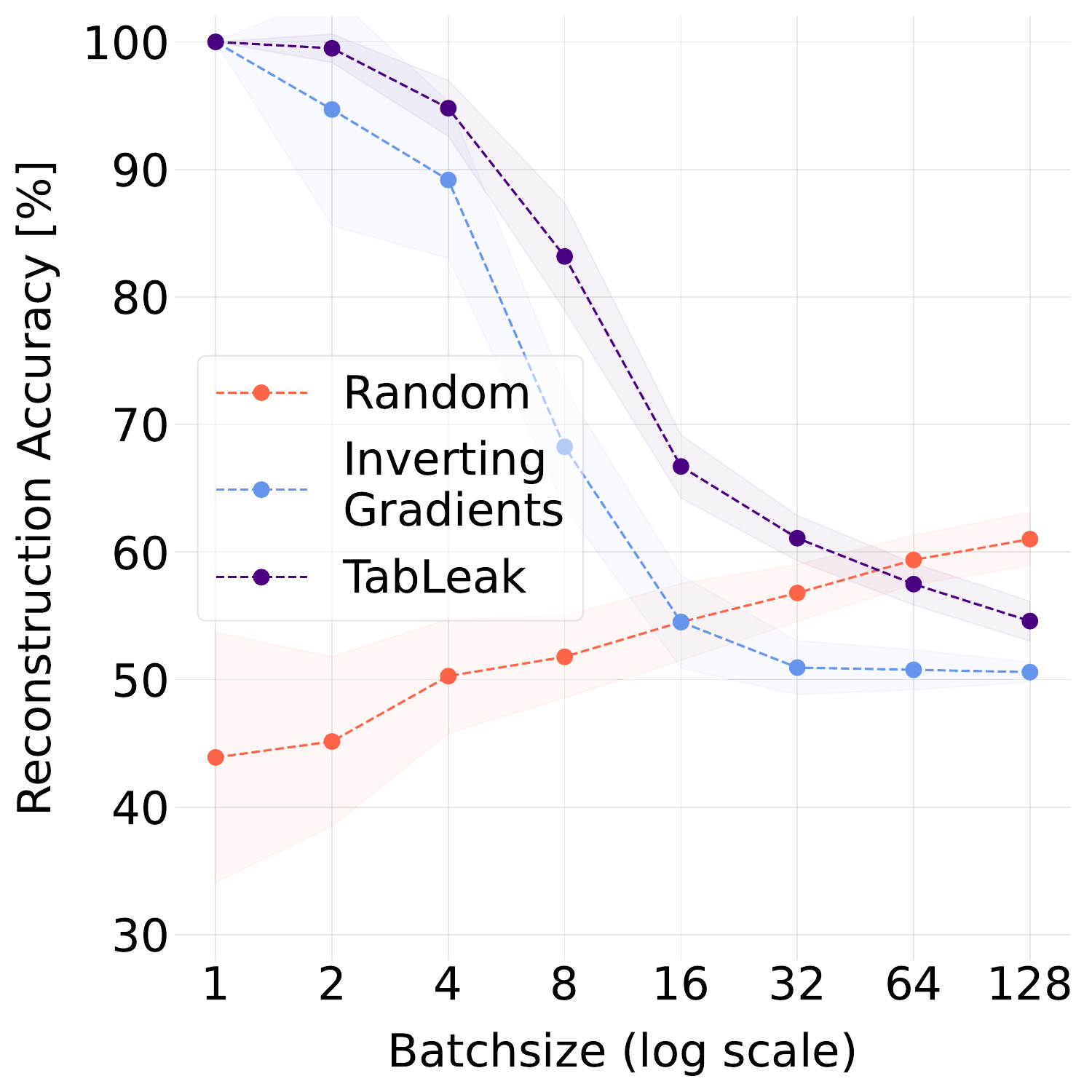}
		\subcaption{Noise standard deviation $= 0.01$}
	\end{subfigure}
	\begin{subfigure}{.33\textwidth}
		\centering
		\includegraphics[width=0.95\textwidth]{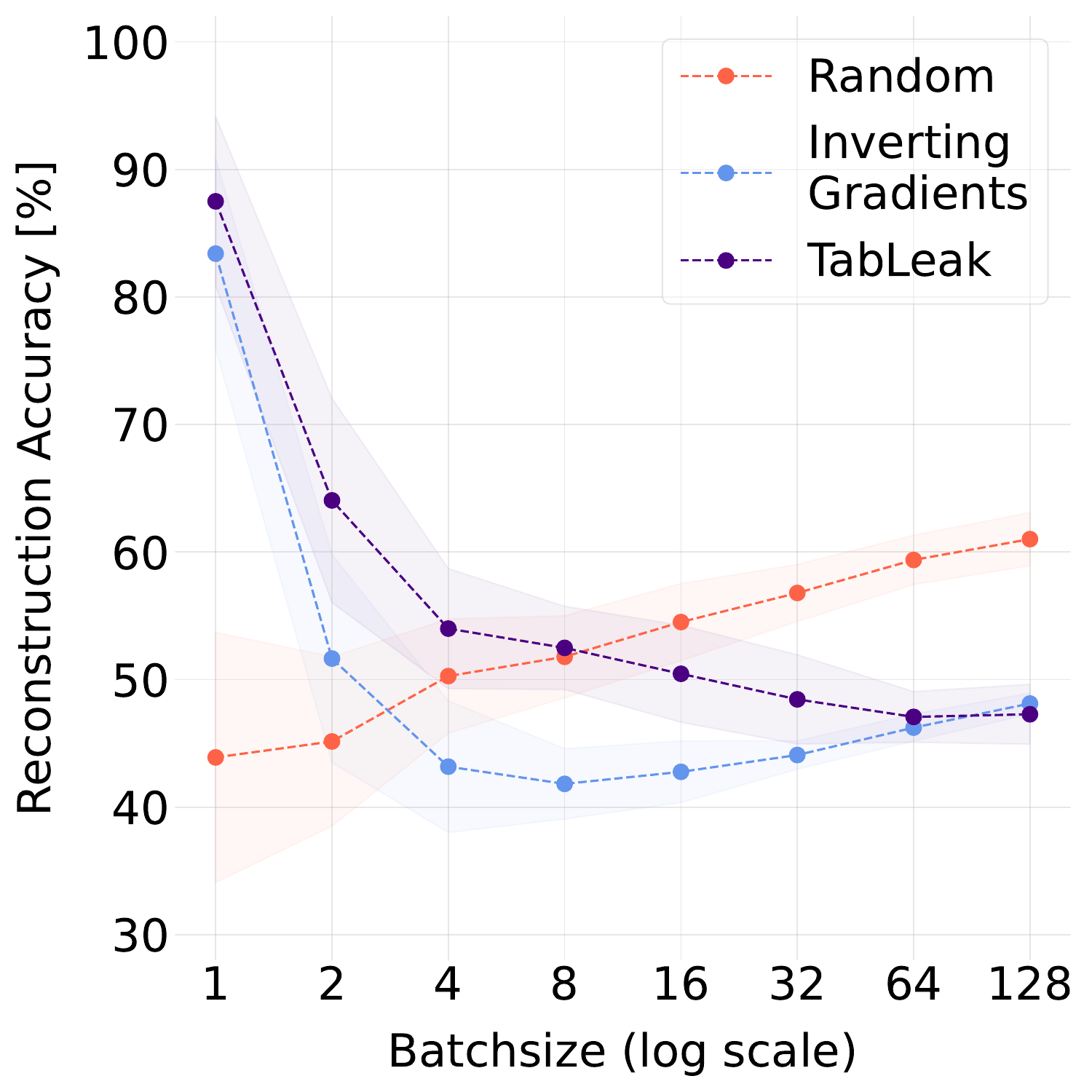}
		\subcaption{Noise standard deviation $= 0.1$}
	\end{subfigure}
	\caption{Mean and standard deviation accuracy [\%] curves over batch size at varying Gaussian noise level $\sigma$ added to the client gradients for differential privacy on the \textbf{German Credit} dataset.}
	\label{fig:dp_german}
\end{figure}

\begin{figure}
	\centering
	\begin{subfigure}{.33\textwidth}
		\centering
		\includegraphics[width=0.95\textwidth]{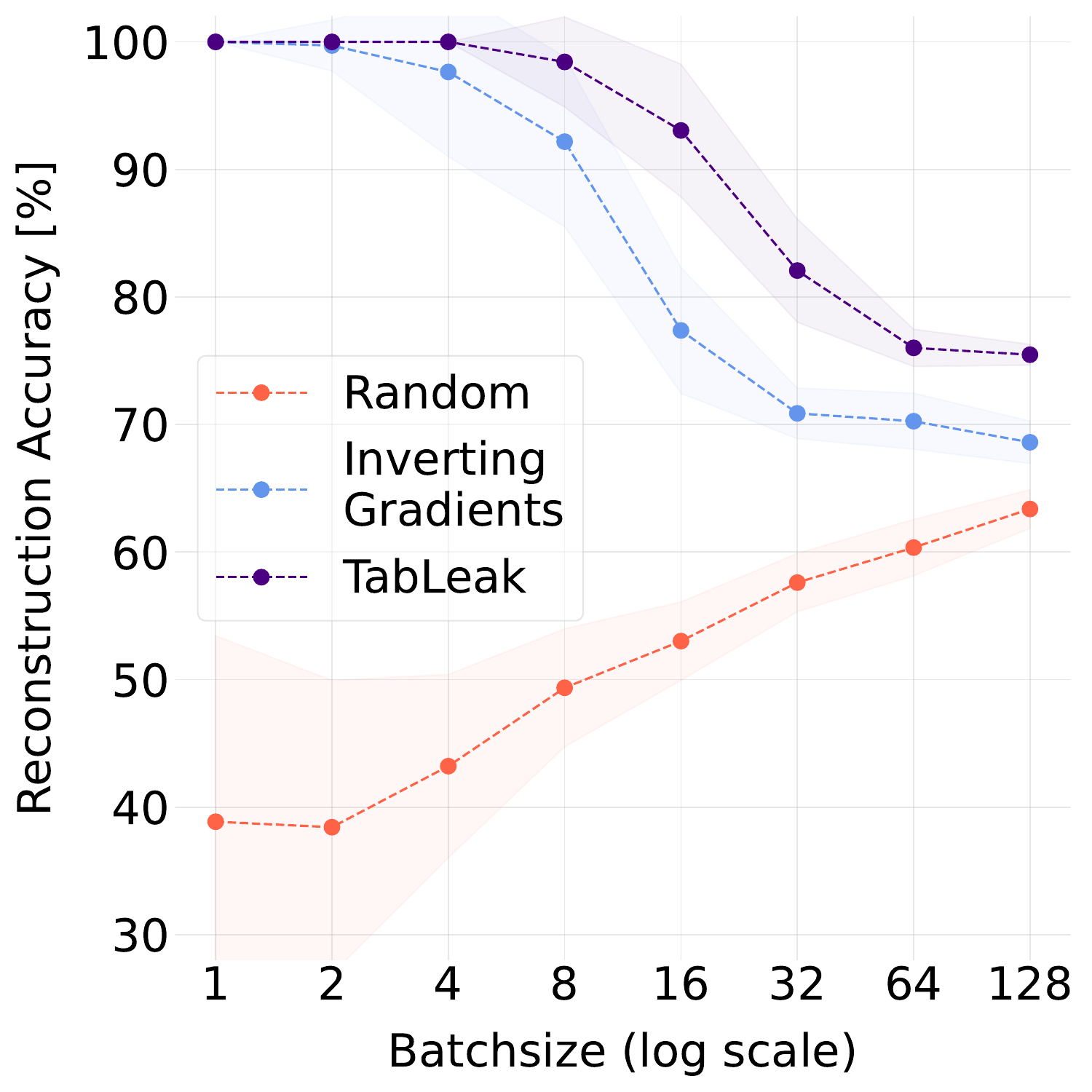}
		\subcaption{Noise standard deviation $= 0.001$}
	\end{subfigure}%
	\begin{subfigure}{.33\textwidth}
		\centering
		\includegraphics[width=0.95\textwidth]{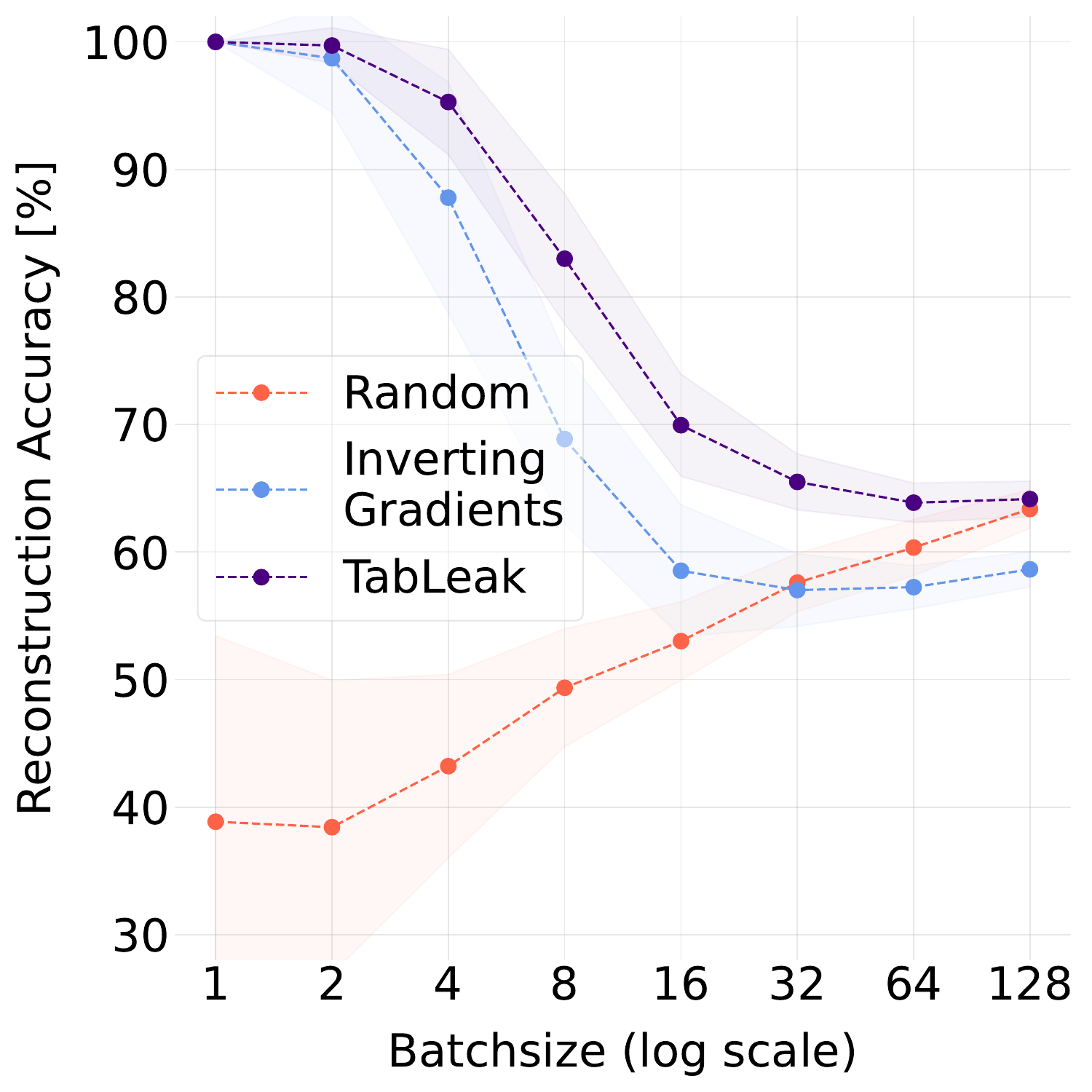}
		\subcaption{Noise standard deviation $= 0.01$}
	\end{subfigure}
	\begin{subfigure}{.33\textwidth}
		\centering
		\includegraphics[width=0.95\textwidth]{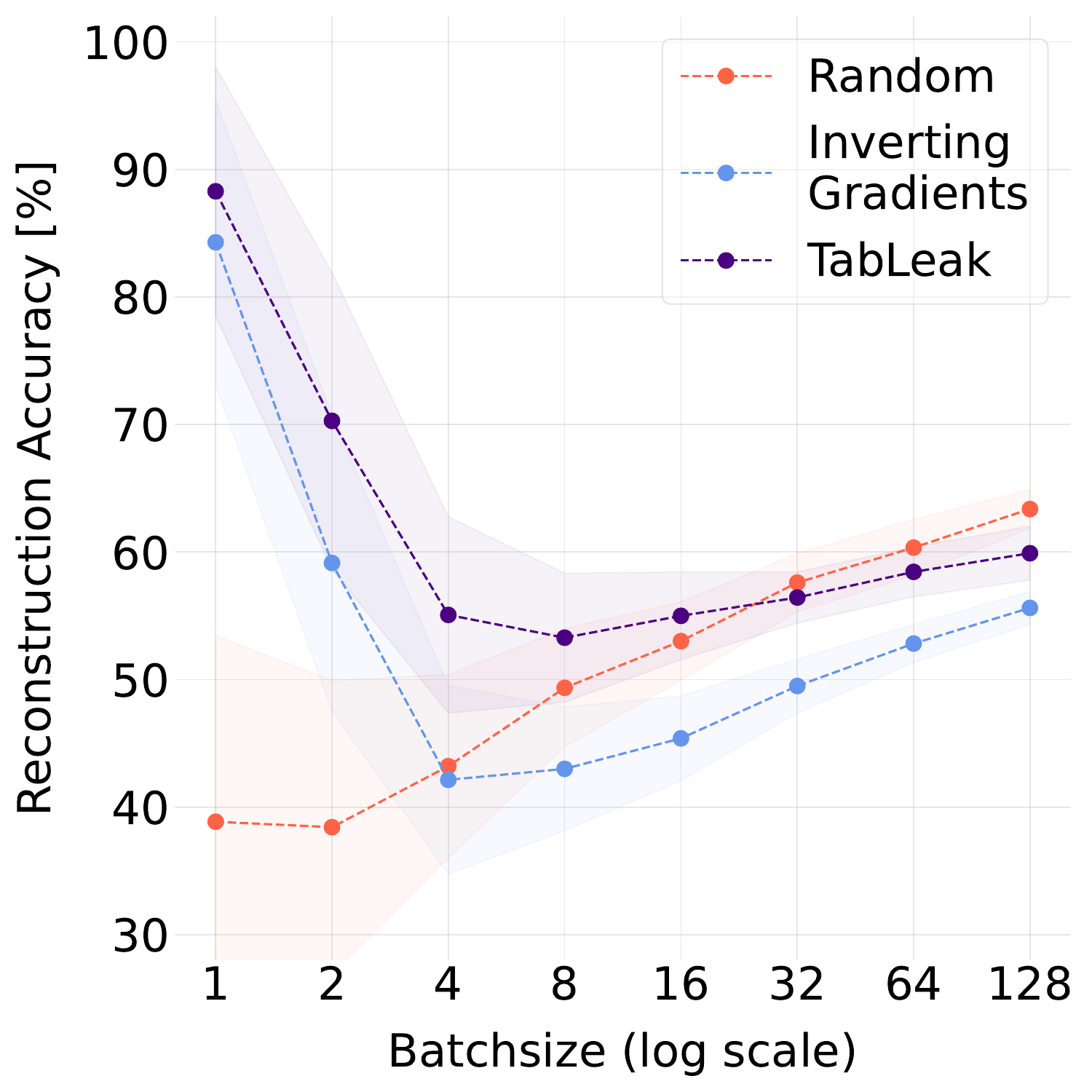}
		\subcaption{Noise standard deviation $= 0.1$}
	\end{subfigure}
	\caption{Mean and standard deviation accuracy [\%] curves over batch size at varying Gaussian noise level $\sigma$ added to the client gradients for differential privacy on the \textbf{Lawschool Admissions} dataset.}
	\label{fig:dp_lawschool}
\end{figure}

\begin{figure}
	\centering
	\begin{subfigure}{.33\textwidth}
		\centering
		\includegraphics[width=0.95\textwidth]{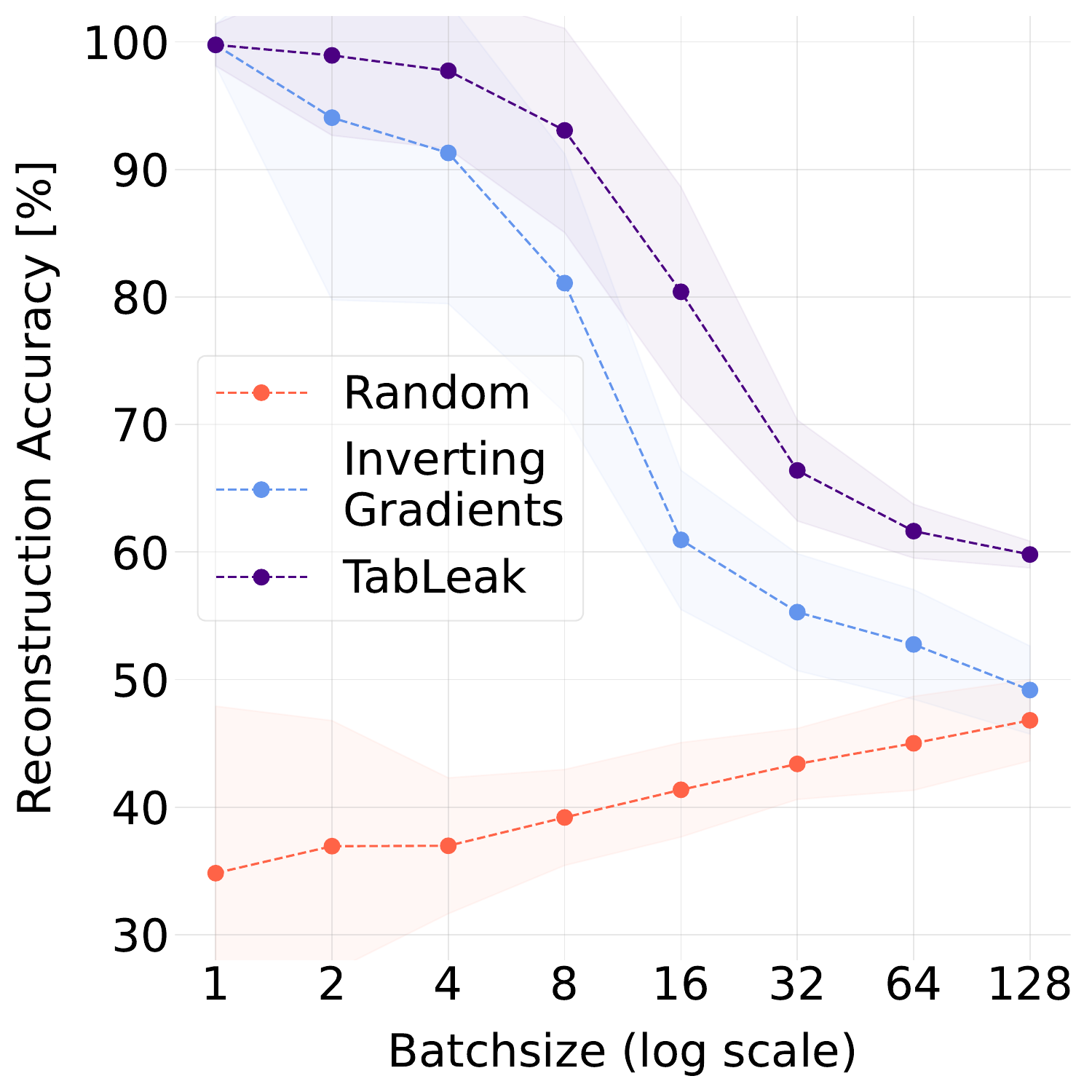}
		\subcaption{Noise standard deviation $= 0.001$}
	\end{subfigure}%
	\begin{subfigure}{.33\textwidth}
		\centering
		\includegraphics[width=0.95\textwidth]{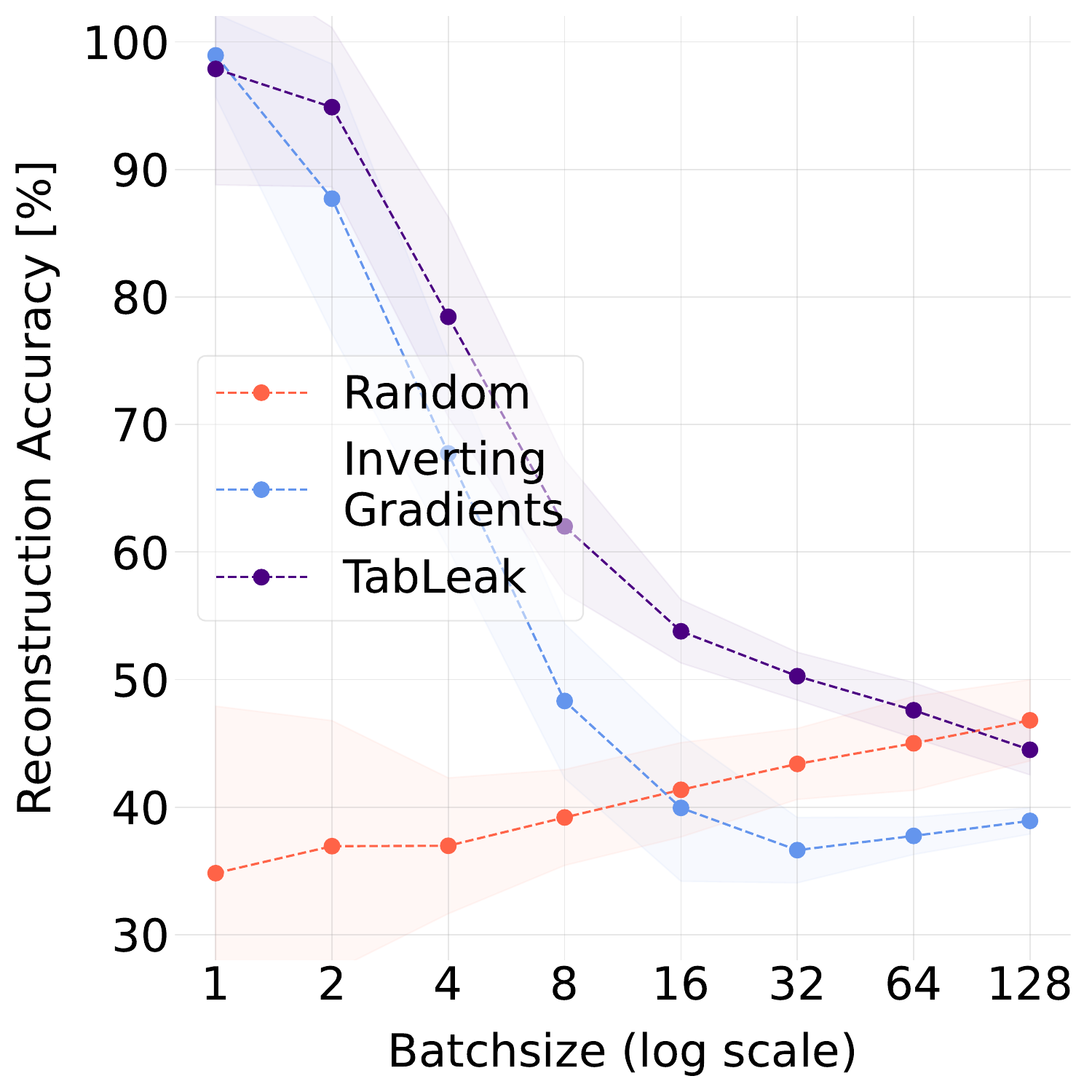}
		\subcaption{Noise standard deviation $= 0.01$}
	\end{subfigure}
	\begin{subfigure}{.33\textwidth}
		\centering
		\includegraphics[width=0.95\textwidth]{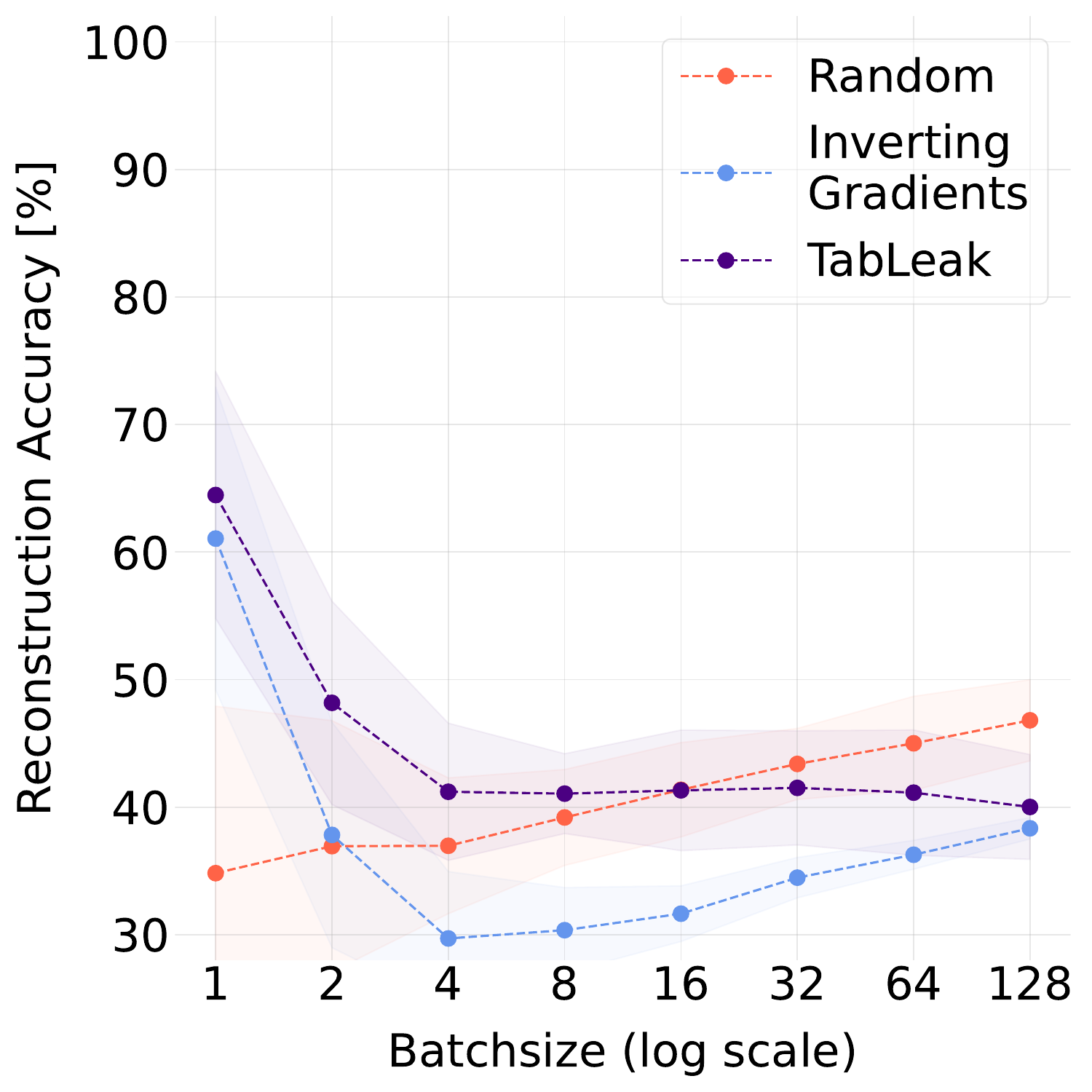}
		\subcaption{Noise standard deviation $= 0.1$}
	\end{subfigure}
	\caption{Mean and standard deviation accuracy [\%] curves over batch size at varying Gaussian noise level $\sigma$ added to the client gradients for differential privacy on the \textbf{Health Heritage} dataset.}
	\label{fig:dp_health}
\end{figure}

Differential privacy (DP) has recently gained popularity, as a way to prevent privacy violations in FL~\citep{Abadi2016,Zhu2019}. Unlike, empirical defenses which are often broken by specifically crafted adversaries~\citep{Balunovic2021}, DP provides guarantees on the amount of data leaked by a FL model, in terms of the magnitude of random noise the clients add to their gradients prior to sharing them with the server~\citep{Abadi2016,Zhu2019}. Naturally, DP methods balance privacy concerns with the accuracy of the produced model, since bigger noise results in worse models that are more private. In this subsection, we evaluate \OurMethod{}, and Inverting Gradients~\citep{Geiping2020} against DP-inspired defended gradient updates, where zero-mean Gaussian noise is added with standard deviations $0.001$, $0.01$, and $0.1$ to the client gradients. Note that this defense does not lead to DP guarantees, as the gradients are not clipped prior to adding noise. Nevertheless, this method is in line with prior works on gradient leakage~\cite{Zhu2019,Dimitrov2022,Dimitrov22}. We present our results on the Adult, German Credit, Lawschool Admissions, and Health Heritage datasets in \cref{fig:dp_adult}, \cref{fig:dp_german}, \cref{fig:dp_lawschool}, and \cref{fig:dp_health}, respectively.
Although both attack methods are affected by the defense, our method consistently produces better reconstructions than the baseline method. However, for high noise level (standard deviation $= 0.1$) and larger batch size both attacks break, advocating for the use of DP-inpsired defenses in tabular FL to prevent the high vulnerability exposed by this work.
\clearpage

\clearpage

\clearpage
\subsection{Varying Network Size}
\label{appendix:varying_network_size}
\begin{figure}
	\centering
	\begin{subfigure}{.5\textwidth}
		\centering
		\includegraphics[width=0.9\textwidth]{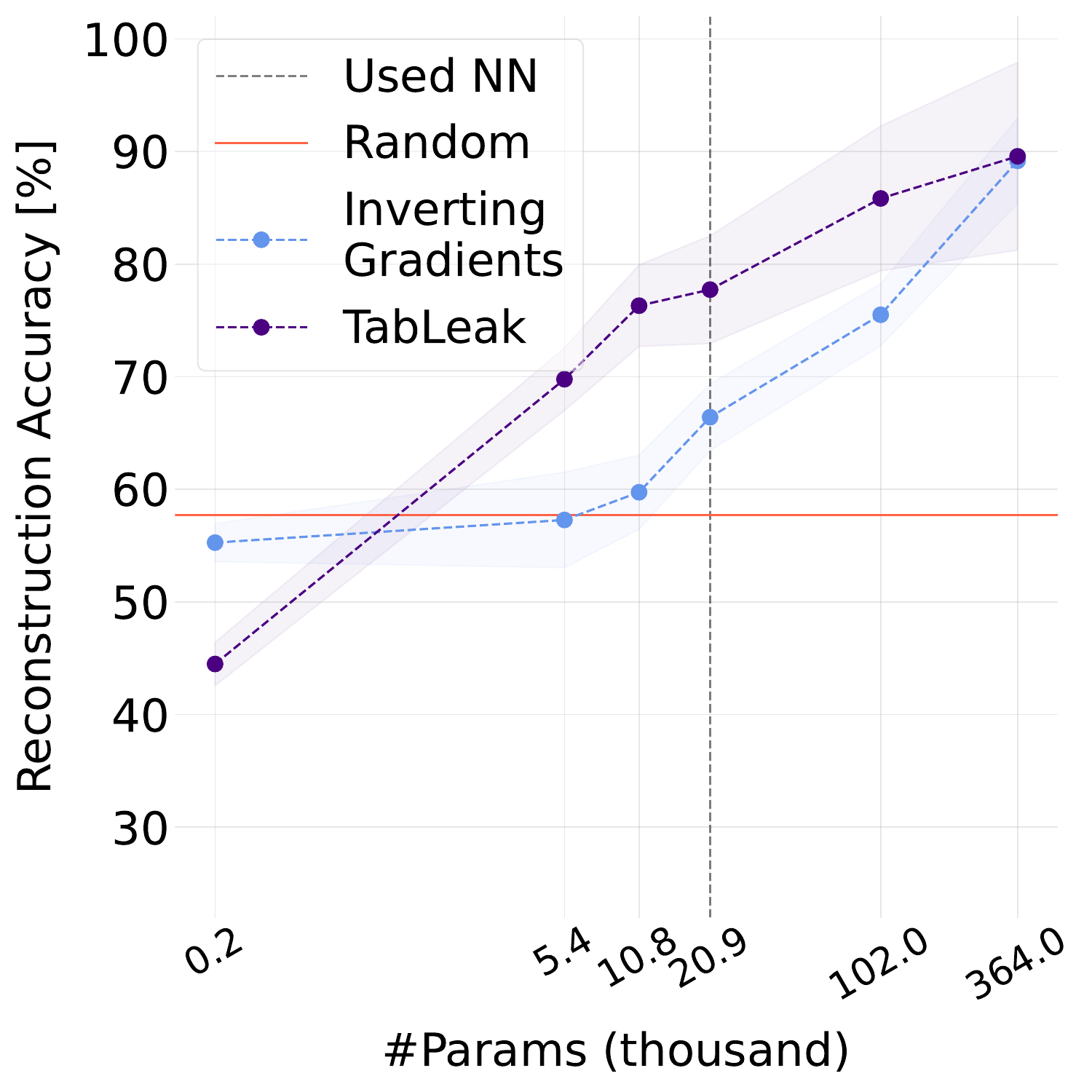}
		\subcaption{Adult: $d=105$}
	\end{subfigure}%
	\begin{subfigure}{.5\textwidth}
		\centering
		\includegraphics[width=0.9\textwidth]{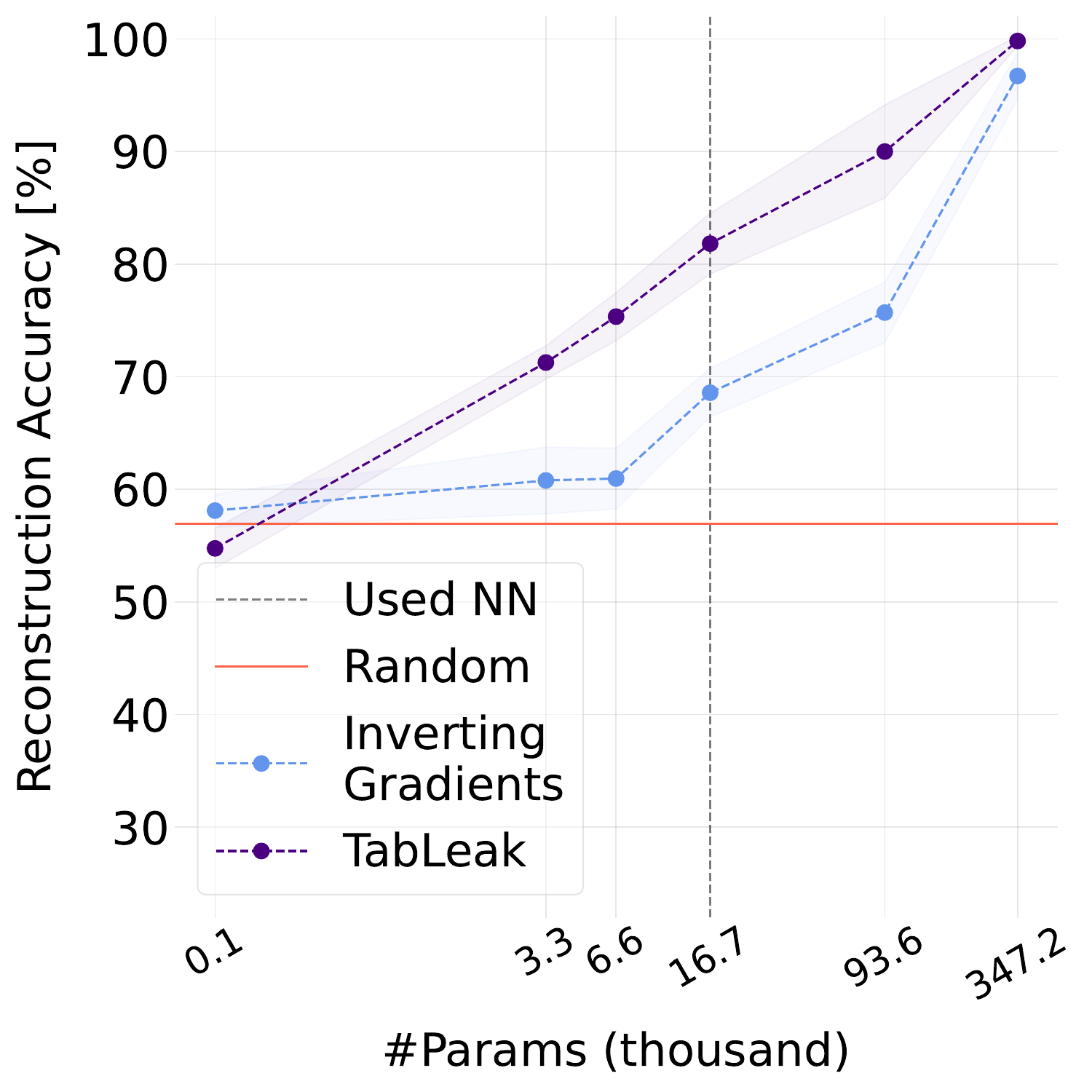}
		\subcaption{German Credit: $d=63$}
	\end{subfigure}
	\begin{subfigure}{.5\textwidth}
		\centering
		\includegraphics[width=0.9\textwidth]{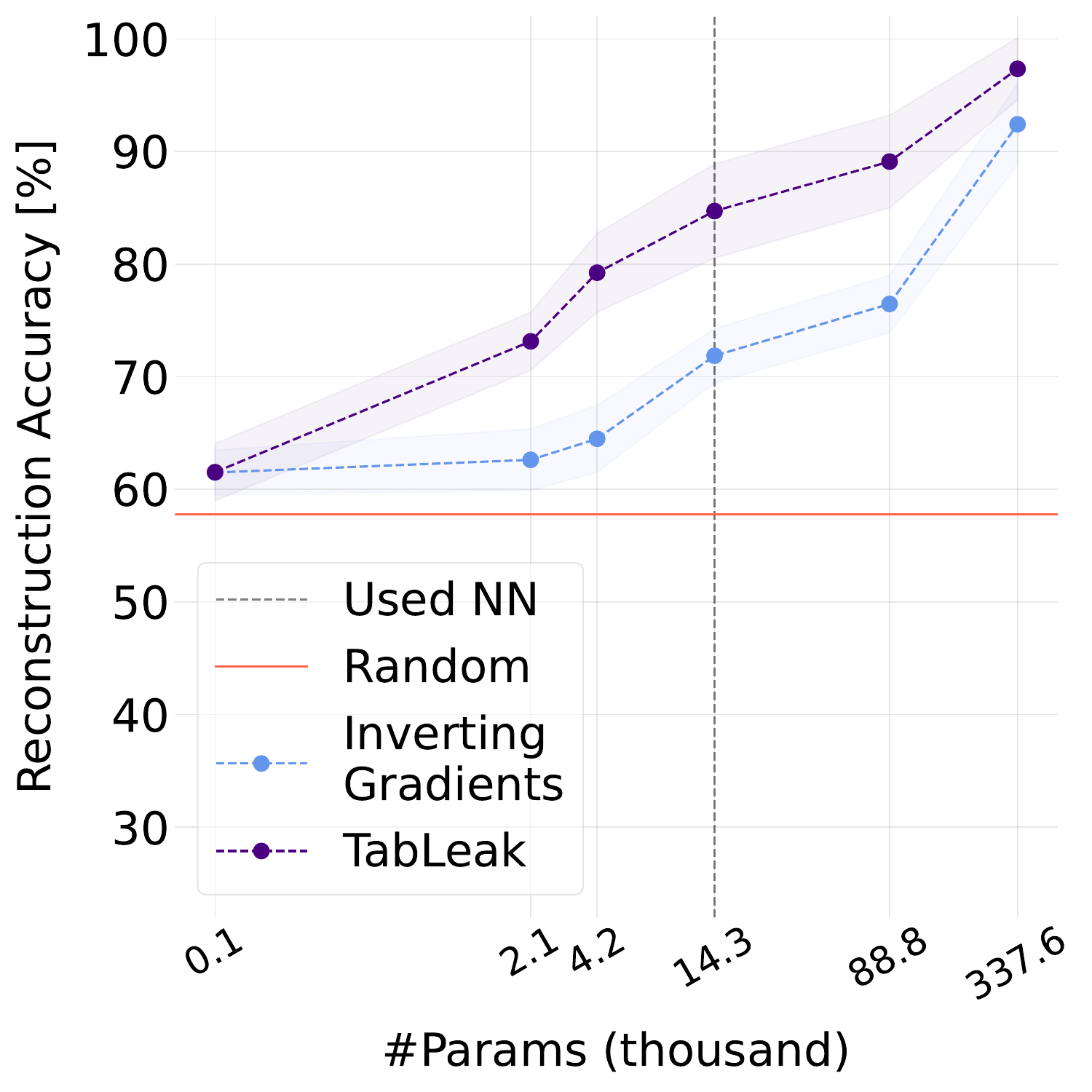}
		\subcaption{Lawschool Admissions: $d=39$}
	\end{subfigure}%
	\begin{subfigure}{.5\textwidth}
		\centering
		\includegraphics[width=0.9\textwidth]{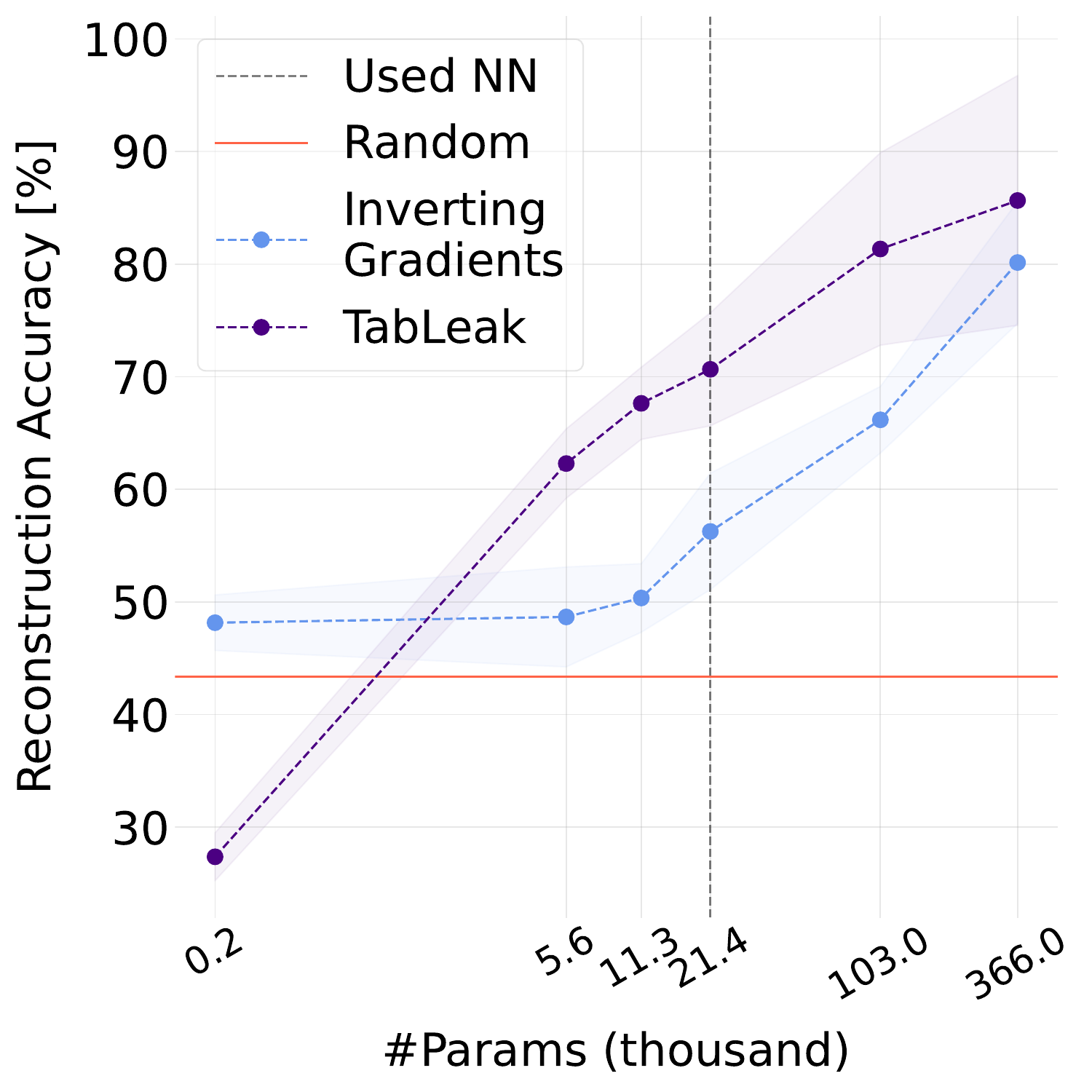}
		\subcaption{Health Heritage: $d=110$}
	\end{subfigure}
	\caption{Mean attack accuracy curves with standard deviation for batch size $32$ over varying network size (measured in number of parameters, \#Params, log scale) on all four datasets with $d$ number of features after encoding. We mark the network we used for our other experiments with a dashed vertical line. From left to right we have the following models: Linear, NN 1, NN 2, NN 3, NN 4, and NN 5.}
	\label{fig:network_size_all}
\end{figure}

To understand the effect the choice of the network has on the obtained reconstruction results, we defined $4$ additional fully connected networks, two smaller, and two bigger ones to evaluate \OurMethod{} on. As a simple linear model is often a good baseline for tabular data, we add it also to the range of attacked models. Concretely, we examined the following six models for our attack:
\begin{itemize}
	\item Linear: a linear classification network: $f_{W}(c(x)) = \sigma (Wc(x) + b)$,
	\item NN 1: a single hidden layer neural network with $50$ neurons,
	\item NN 2: a single hidden layer neural network with $100$ neurons,
	\item NN 3: a neural network with two hidden layers of $100$ neurons each (network used in main body),
	\item NN 4: a neural network with three hidden layers of $200$ neurons each,
	\item NN 5: a three hidden layer neural network with $400$ neurons in each layer.
\end{itemize}
We attack the above networks, aiming to reconstruct a batch of size $32$. We plot the accuracy of \OurMethod{} and of Inverting Gradients~\citep{Geiping2020} as a function of the number of parameters in the network in \cref{fig:network_size_all} for all four datasets. We can observe that with increasing number of parameters in the network, the reconstruction accuracy significantly increases on all datasets, and rather surprisingly, allowing for near perfect reconstruction of a batch as large as $32$ in some cases. Observe that on both ends of the presented parameter scale the differences between the methods degrade, \ie they either both converge to near-perfect reconstruction (large networks) or to random guessing (small networks). Therefore, the choice of our network for conducting the experiments was instructive in examining the differences between the methods.

\begin{figure}
	\centering
	\begin{subfigure}{.5\textwidth}
		\centering
		\includegraphics[width=0.9\textwidth]{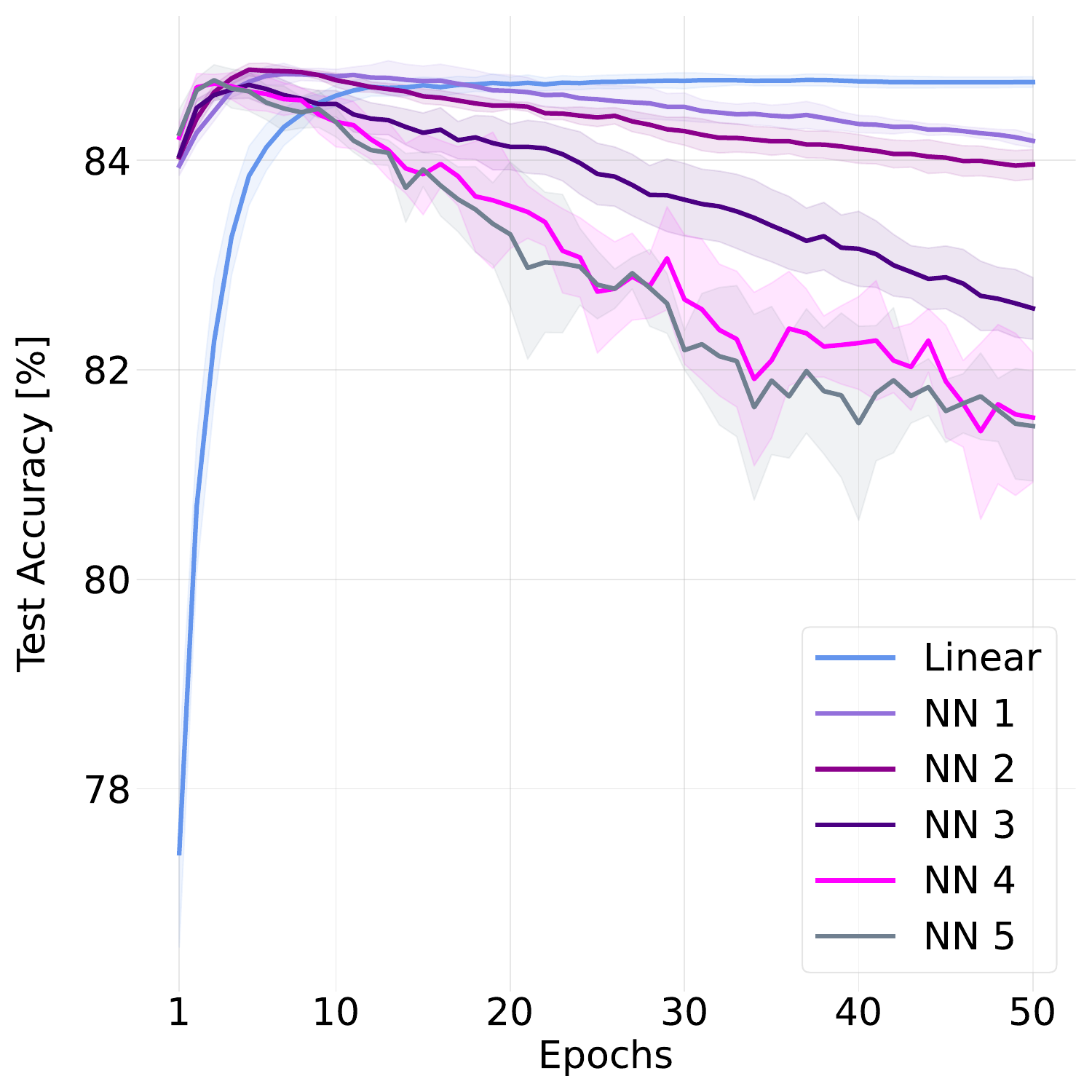}
		\subcaption{Adult}
	\end{subfigure}%
	\begin{subfigure}{.5\textwidth}
		\centering
		\includegraphics[width=0.9\textwidth]{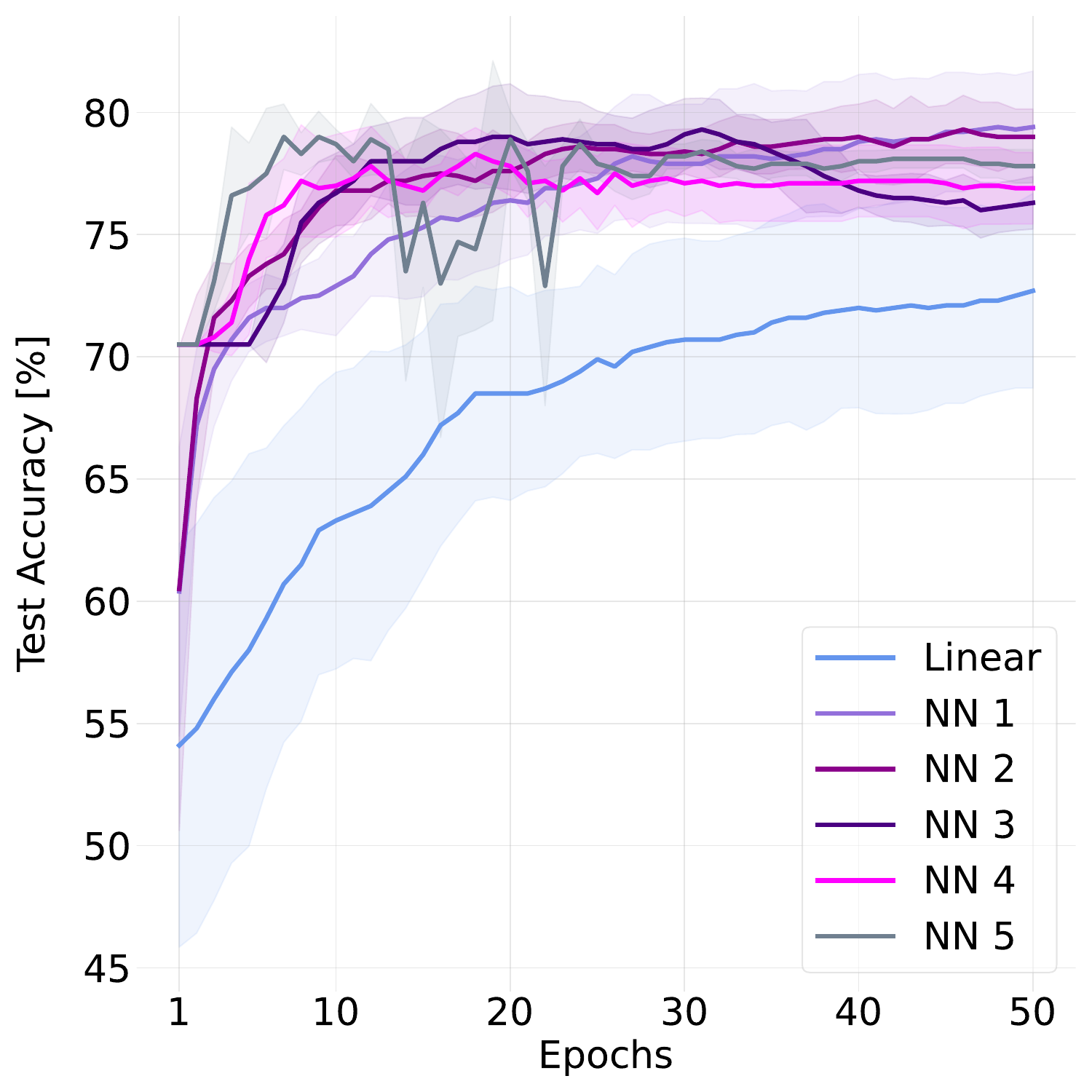}
		\subcaption{German Credit}
	\end{subfigure}
	\begin{subfigure}{.5\textwidth}
		\centering
		\includegraphics[width=0.9\textwidth]{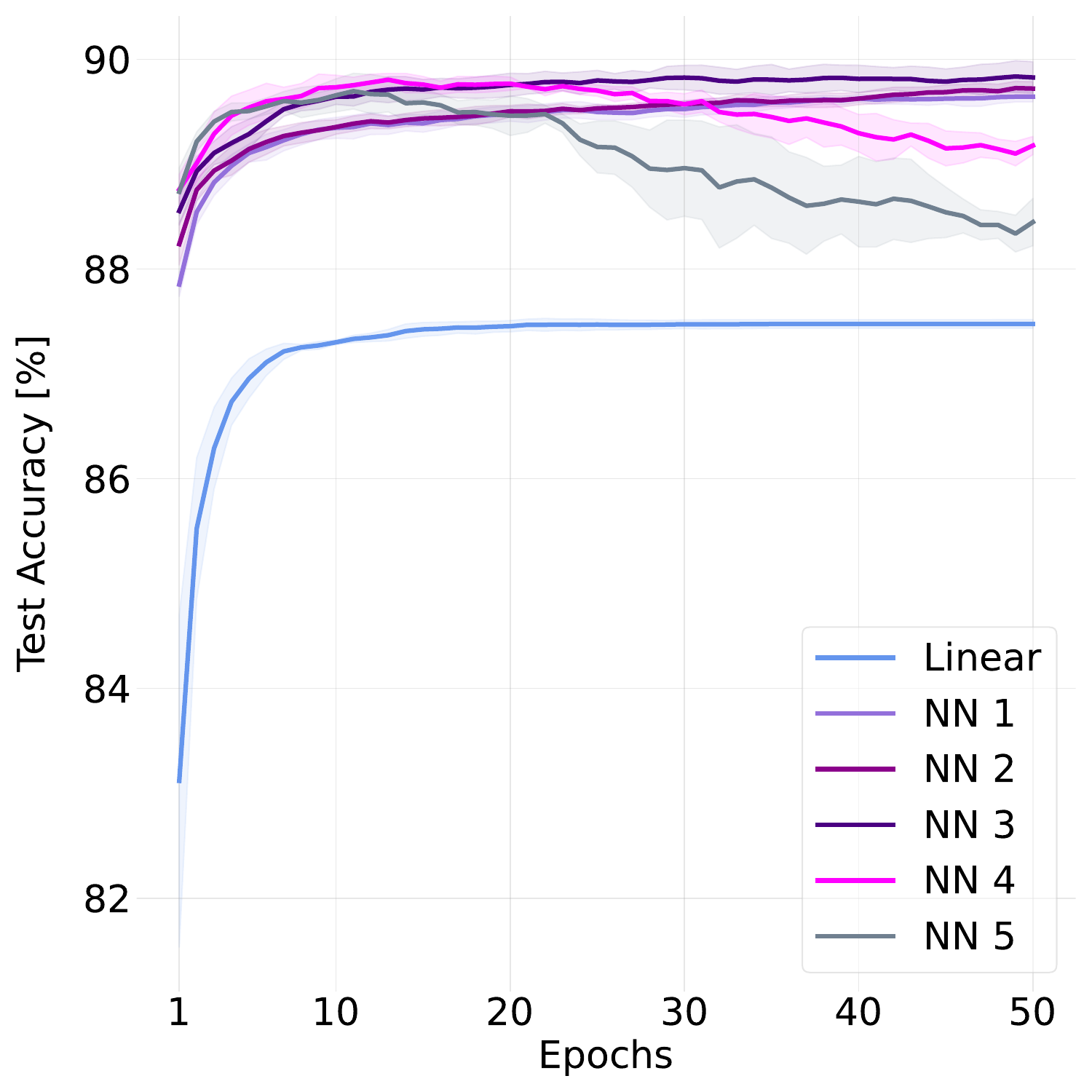}
		\subcaption{Lawschool Admissions}
	\end{subfigure}%
	\begin{subfigure}{.5\textwidth}
		\centering
		\includegraphics[width=0.9\textwidth]{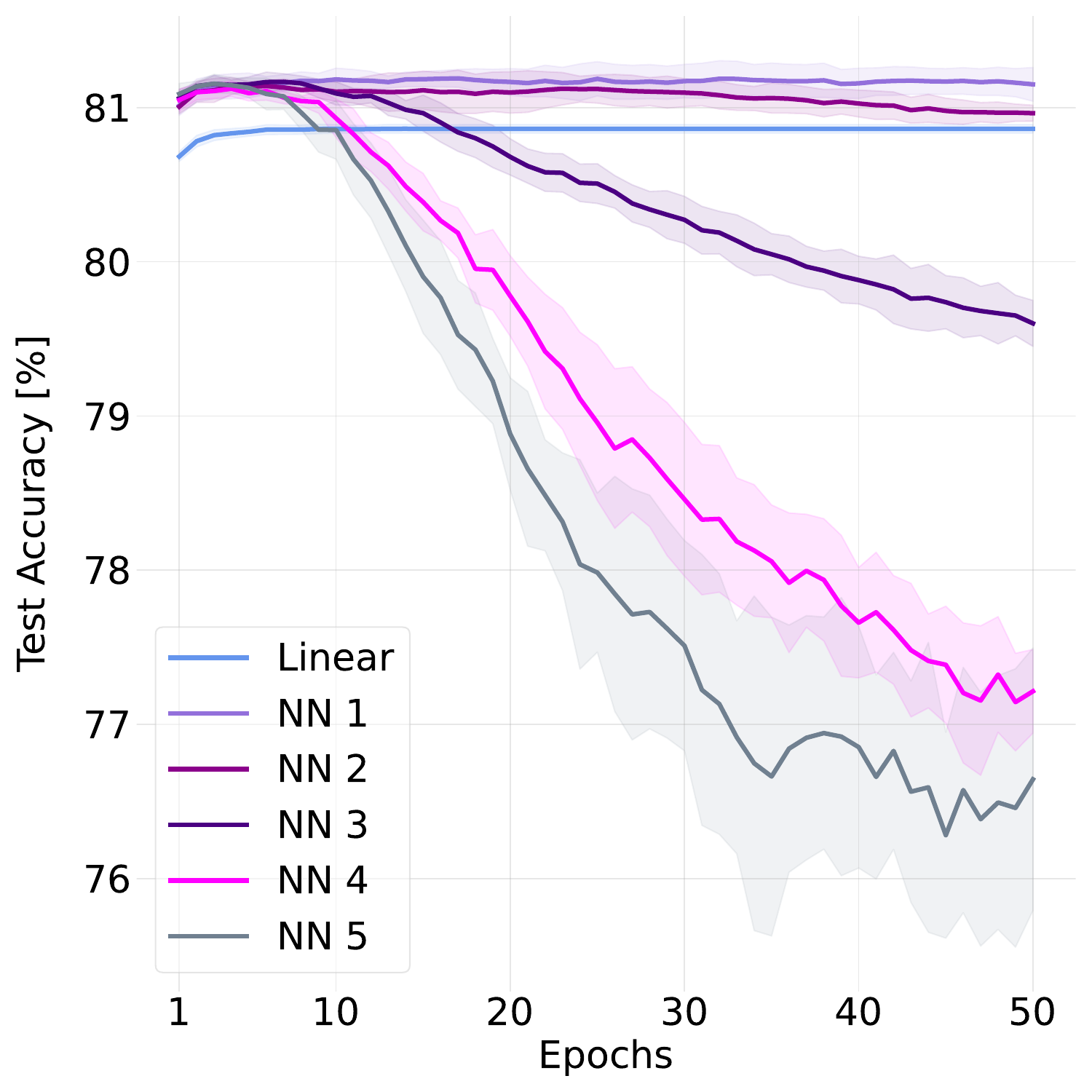}
		\subcaption{Health Heritage}
	\end{subfigure}
	\caption{Mean test and standard deviation of the test accuracy over epochs during five independent runs of training for each examined model on all four datasets. For our experiments elsewhere we used the network corresponding to Layout 3, marked in dark violet here.}
	\label{fig:network_size_training_curves}
\end{figure}

Additionally, to better understand the relevance of the models examined here, we train them on each of the datasets for 50 epochs and observe their behavior through monitoring their performance on a secluded test set of each dataset. We do this for 5 different initializations of each model, and report the mean and the standard deviation of the test accuracy at each training epoch for each model. Note that we do not train the models using any FL protocol, merely, this experiment serves to give a better understanding between the relation of the given dataset and the model used, putting also the attack success data in better perspective. For training, we use the Adam~\citet{Kingma2014} optimizer and batch size 256 for each of the datasets, except for the German Credit dataset, where we train with batch size 64 due to its small size. We provide all test accuracy curves over training in \cref{fig:network_size_training_curves}. From the accuracy curves we can observe that most large models that are easy to attack tend to overfit quickly to the data, indicating a heavily overparameterized regime. Additionally, in \cref{table:peak_accuracy_network_size} we provide the peak mean test accuracies per dataset and model, effectively corresponding to a 'perfect' early-stopping. The linear model could appear to be an overall good choice, as it is very hard to attack and shows good stability during training, however, it does not achieve competitive performance on most datasets. In \cref{table:peak_accuracy_network_size} the non-linear models always outperform the linear model, and achieve comparable performance across themselves in this ideal setting, where overfitting can be prevented by monitoring on the test data\footnote{In practice a proxy metric would be necessary to achieve early-stopping, such as monitoring the performance on a separate validation set split from the training data}. Conclusively, simpler non-linear models shall be pursued for FL on tabular data, as they are less prone to overfitting and provide better protection from data leakage attacks.

\begin{table}
	\caption{Mean and standard deviation of the peak test accuracy of each of the examined 6 models on the four discussed datasets over training.}
	\label{table:peak_accuracy_network_size}
	\centering
	\resizebox{0.9\columnwidth}{!}{
	\begin{tabular}{lcccccc}
		\toprule
		  & Linear & Layout 1 & Layout 2 & Layout 3 & Layout 4 & Layout 5\\
		\midrule
		Adult & $84.7 \pm 0.1$ & $84.9 \pm 0.1$ & $\mathbf{85.0 \pm 0.1}$ & $84.8 \pm 0.1$ & $84.8 \pm 0.1$ & $84.7 \pm 0.1$  \\
		German & $73.0 \pm 1.4$ & $80.0 \pm 1.1$ & $79.5 \pm 0.6$ & $\mathbf{80.9 \pm 0.7}$ & $78.9 \pm 1.0$ & $79.4 \pm 1.8$ \\
		Lawschool & $87.4 \pm 0.0$ & $89.6 \pm 0.1$ & $89.8 \pm 0.0$ & $\mathbf{90.0 \pm 0.1}$ & $89.8 \pm 0.1$ & $89.8 \pm 0.1$ \\
		Health Heritage & $80.9 \pm 0.0$ & $\mathbf{81.2 \pm 0.1}$ & $\mathbf{81.2 \pm 0.0}$ & $\mathbf{81.2 \pm 0.1}$ & $\mathbf{81.2 \pm 0.1}$ & $81.1 \pm 0.1$ \\
		\bottomrule
	\end{tabular}
	}
\end{table}

\subsection{Varying Network Architecture}
\label{appendix:varying_network_architecture}
To investigate the impact of the network architecture on the attack success, we test \OurMethod{} and two baseline methods, Inverting Gradients~\citep{Geiping2020} and GradInversion~\cite{Yin2021} (introduced in \cref{sec:results} in the main body) on various network architectures. The examined architectures are:
\begin{itemize}
	\item Linear: a linear classification network: $f_{W}(c(x)) = \sigma (Wc(x) + b)$;
	\item FC NN: a neural network with two hidden layers of $100$ neurons each (network used in the main body);
	\item FC NN large: a three hidden layer neural network with $400$ neurons in each layer;
	\item CNN (BN): a convolutional neural network with a single initial convolutional layer of kernel size $3$ and $16$ output channels, followed by a batch normalization layer (BN) and two fully connected hidden layers of $100$ neurons;
	\item ResNet (BN): a fully connected neural network with two residual blocks, each containing a batch normalization layer.
\end{itemize}

Our results on all four datasets are included in \cref{table:varying_architectures_appx_adult,table:varying_architectures_appx_german,table:varying_architectures_appx_lawschool,table:varying_architectures_appx_health}. Note that on the last two architectures we raised the number of iterations for all the attacks to $7\,000$. We can confirm the three observations we have already made in the main body of the paper: (i) \OurMethod{} is the strongest overall attack across various architectures, with BN layers not impacting its position either, and in line with \cref{appendix:varying_network_size}, we again observe that large networks are excessively vulnerable to all attacks (ii) and that the linear model is hard to break for any attack (iii). Therefore, we again argue for a conservative architecture choice, with as little parameters as possible that are still fit to solve the underlying task. 

\begin{table}[H]
	\caption{The mean and the standard deviation of the attack accuracy [\%] over different architectures inverting batches of size $32$ on the \textbf{Adult} dataset. The random baseline at this batch size is $58.0 \pm 2.9$.}
	\label{table:varying_architectures_appx_adult}
	\centering
	\resizebox{.95\columnwidth}{!}{
	\begin{tabular}{lccccc}
		\toprule
		Attack & Linear & FC NN & FC NN large & CNN (BN) & ResNet (BN)\\
		\midrule
		 Inverting Gradients~\citep{Geiping2020} & $55.3 \pm 1.7$ & $66.6 \pm 3.5$ & $89.2 \pm 3.8$ & $43.4 \pm 2.0$ & $61.7 \pm 4.0$ \\
		 GradInversion~\citep{Yin2021} & $\mathbf{61.3 \pm 2.7}$ & $67.7 \pm 2.5$  & $88.0 \pm 3.0$ & $72.8 \pm 2.7$ & $67.6 \pm 2.6$ \\
		 \OurMethod{} & $44.5 \pm 1.9$ & $\mathbf{79.3 \pm 4.5}$ & $\mathbf{89.6 \pm 8.3}$ & $\mathbf{83.7 \pm 2.7}$ & $\mathbf{71.4 \pm 9.2}$ \\
		\bottomrule
	\end{tabular}
	}
\end{table}

\begin{table}[H]
	\caption{The mean and the standard deviation of the attack accuracy [\%] over different architectures inverting batches of size $32$ on the \textbf{German Credit} dataset. The random baseline at this batch size is $56.8 \pm 2.2$.}
	\label{table:varying_architectures_appx_german}
	\centering
	\resizebox{.95\columnwidth}{!}{
	\begin{tabular}{lccccc}
		\toprule
		Attack & Linear & FC NN & FC NN large & CNN (BN) & ResNet (BN)\\
		\midrule
		 Inverting Gradients~\citep{Geiping2020} & $\mathbf{58.1 \pm 1.5}$ & $69.7 \pm 2.2$ & $96.7 \pm 2.1$ & $60.7 \pm 2.3$ & $66.4 \pm 3.0$ \\
		 GradInversion~\citep{Yin2021} & $57.8 \pm 1.4$ & $68.8 \pm 1.8$  & $97.0 \pm 2.1$ & $71.7 \pm 1.9$ & $70.9 \pm 2.3$ \\
		 \OurMethod{} & $54.8 \pm 1.8$ & $\mathbf{84.2 \pm 2.8}$ & $\mathbf{99.8 \pm 0.4}$ & $\mathbf{79.2 \pm 3.6}$ & $\mathbf{74.5 \pm 4.2}$ \\
		\bottomrule
	\end{tabular}
	}
\end{table}

\begin{table}[H]
	\caption{The mean and the standard deviation of the attack accuracy [\%] over different architectures inverting batches of size $32$ on the \textbf{Lawschool Admissions} dataset. The random baseline at this batch size is $57.6 \pm 2.3$.}
	\label{table:varying_architectures_appx_lawschool}
	\centering
	\resizebox{.95\columnwidth}{!}{
	\begin{tabular}{lccccc}
		\toprule
		Attack & Linear & FC NN & FC NN large & CNN (BN) & ResNet (BN)\\
		\midrule
		 Inverting Gradients~\citep{Geiping2020} & $\mathbf{61.5 \pm 2.0}$ & $71.0 \pm 2.8$ & $92.4 \pm 3.6$ & $61.1 \pm 2.1$ & $69.4 \pm 3.2$ \\
		 GradInversion~\citep{Yin2021} & $61.4 \pm 2.1$ & $71.8 \pm 2.4$  & $91.5 \pm 4.2$ & $74.7 \pm 3.0$ & $74.5 \pm 3.0$ \\
		 \OurMethod{} & $\mathbf{61.5 \pm 2.5}$ & $\mathbf{84.9 \pm 4.0}$ & $\mathbf{97.3 \pm 2.8}$ & $\mathbf{79.1 \pm 3.1}$ & $\mathbf{82.6 \pm 6.7}$ \\
		\bottomrule
	\end{tabular}
	}
\end{table}
\begin{table}[H]
	\caption{The mean and the standard deviation of the attack accuracy [\%] over different architectures inverting batches of size $32$ on the \textbf{Health Heritage} dataset. The random baseline at this batch size is $43.4 \pm 2.8$.}
	\label{table:varying_architectures_appx_health}
	\centering
	\resizebox{.95\columnwidth}{!}{
	\begin{tabular}{lccccc}
		\toprule
		Attack & Linear & FC NN & FC NN large & CNN (BN) & ResNet (BN)\\
		\midrule
		 Inverting Gradients~\citep{Geiping2020} & $48.2 \pm 2.5$ & $57.7 \pm 4.1$ & $80.1 \pm 5.4$ & $29.2 \pm 4.9$ & $54.1 \pm 6.1$ \\
		 GradInversion~\citep{Yin2021} & $\mathbf{48.6 \pm 3.1}$ & $58.2 \pm 2.6$  & $84.3 \pm 5.6$ & $64.9 \pm 3.1$ & $\mathbf{60.1 \pm 2.8}$ \\
		 \OurMethod{} & $27.4 \pm 2.1$ & $\mathbf{70.8 \pm 4.5}$ & $\mathbf{85.7 \pm 11.1}$ & $\mathbf{72.9 \pm 4.6}$ & $46.7 \pm 12.8$\\
		\bottomrule
	\end{tabular}
	}
\end{table}

\subsection{Continuous Feature Reconstruction Measured by RMSE}
\label{appendix:continuous_feature_recosntruction_measured_by_rmse}
\begin{figure}
	\centering
	\begin{subfigure}{.42\textwidth}
		\centering
		\includegraphics[width=0.9\textwidth]{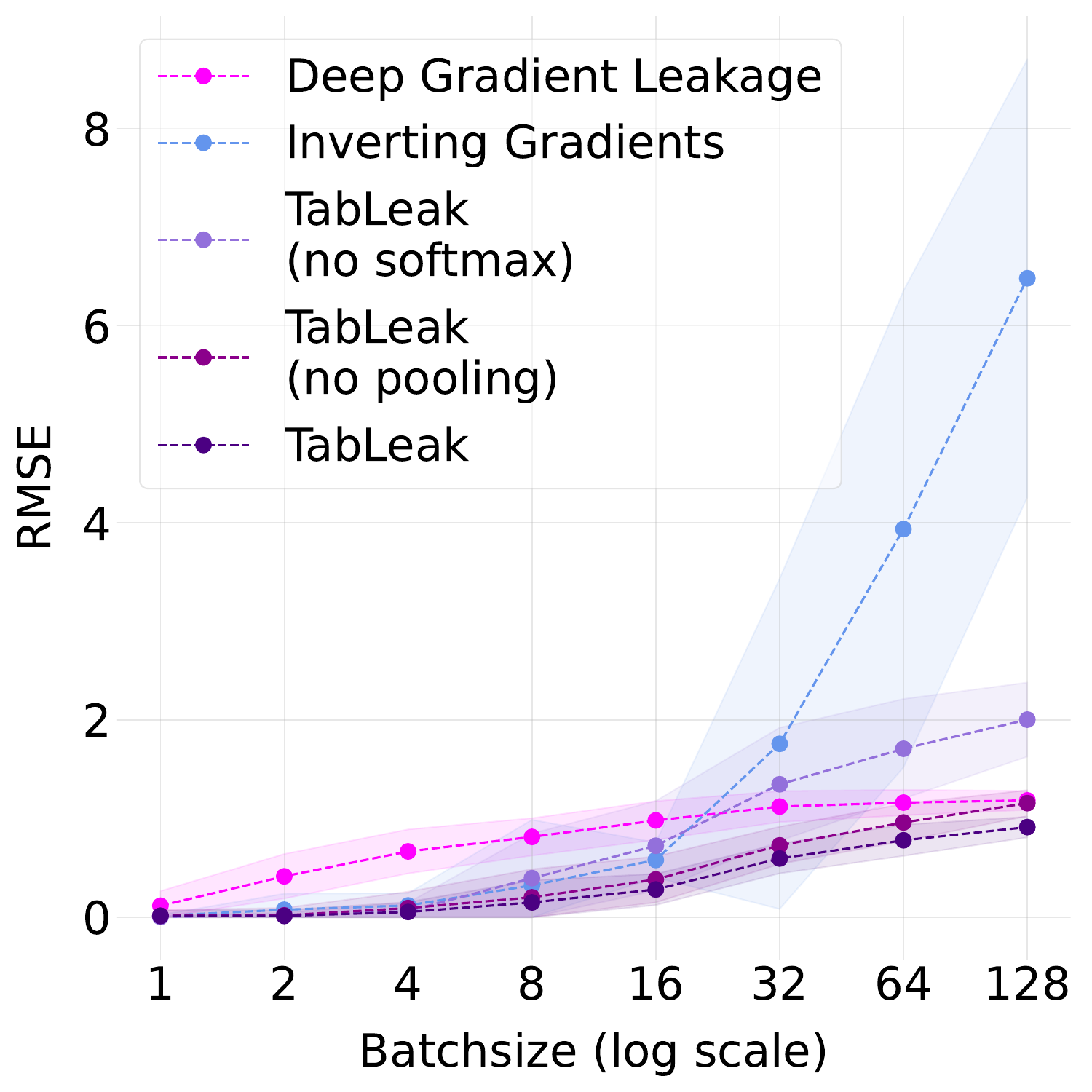}
		\subcaption{Adult}
	\end{subfigure}%
	\begin{subfigure}{.42\textwidth}
		\centering
		\includegraphics[width=0.9\textwidth]{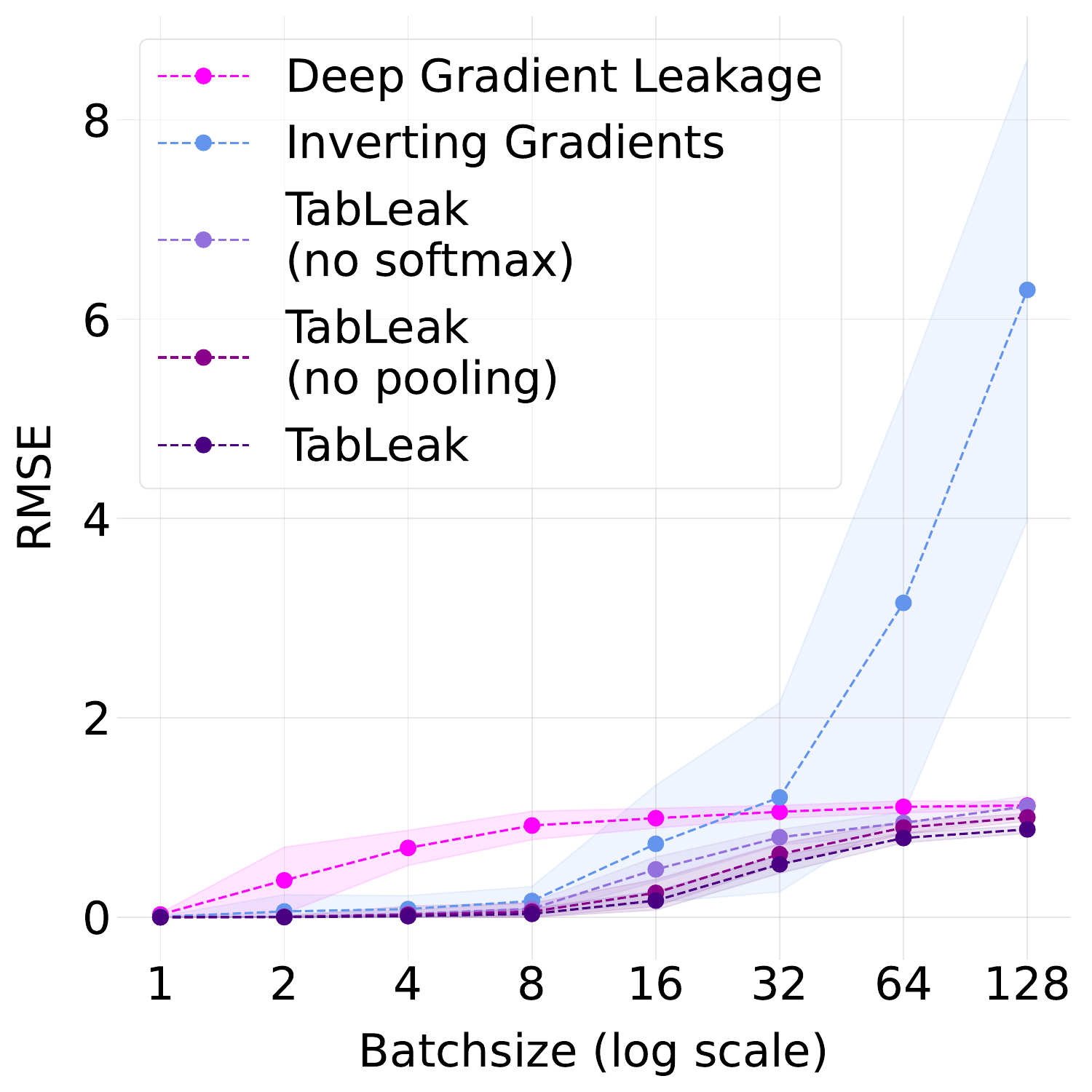}
		\subcaption{German Credit}
	\end{subfigure}
	\begin{subfigure}{.42\textwidth}
		\centering
		\includegraphics[width=0.9\textwidth]{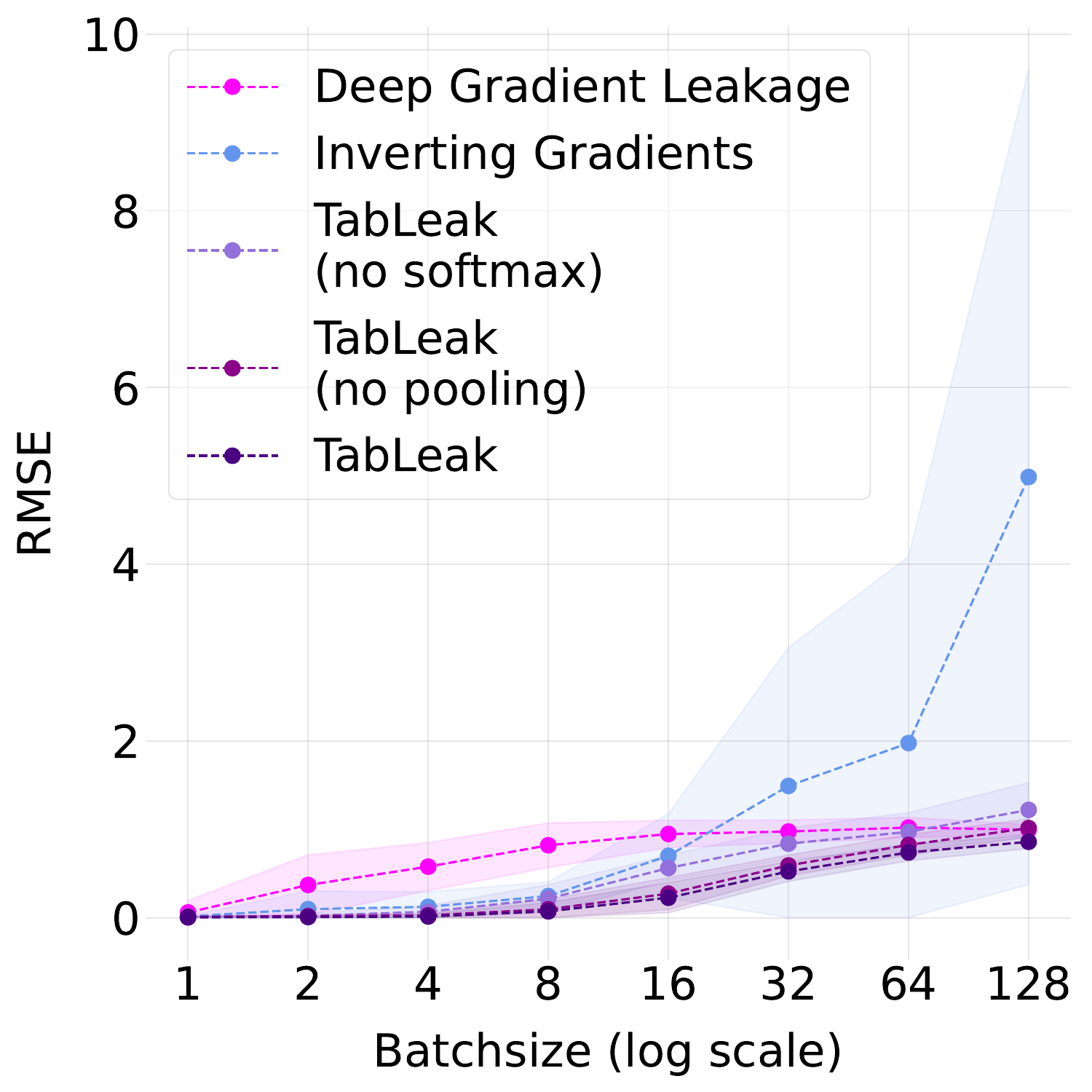}
		\subcaption{Lawschool Admissions}
	\end{subfigure}%
	\begin{subfigure}{.42\textwidth}
		\centering
		\includegraphics[width=0.9\textwidth]{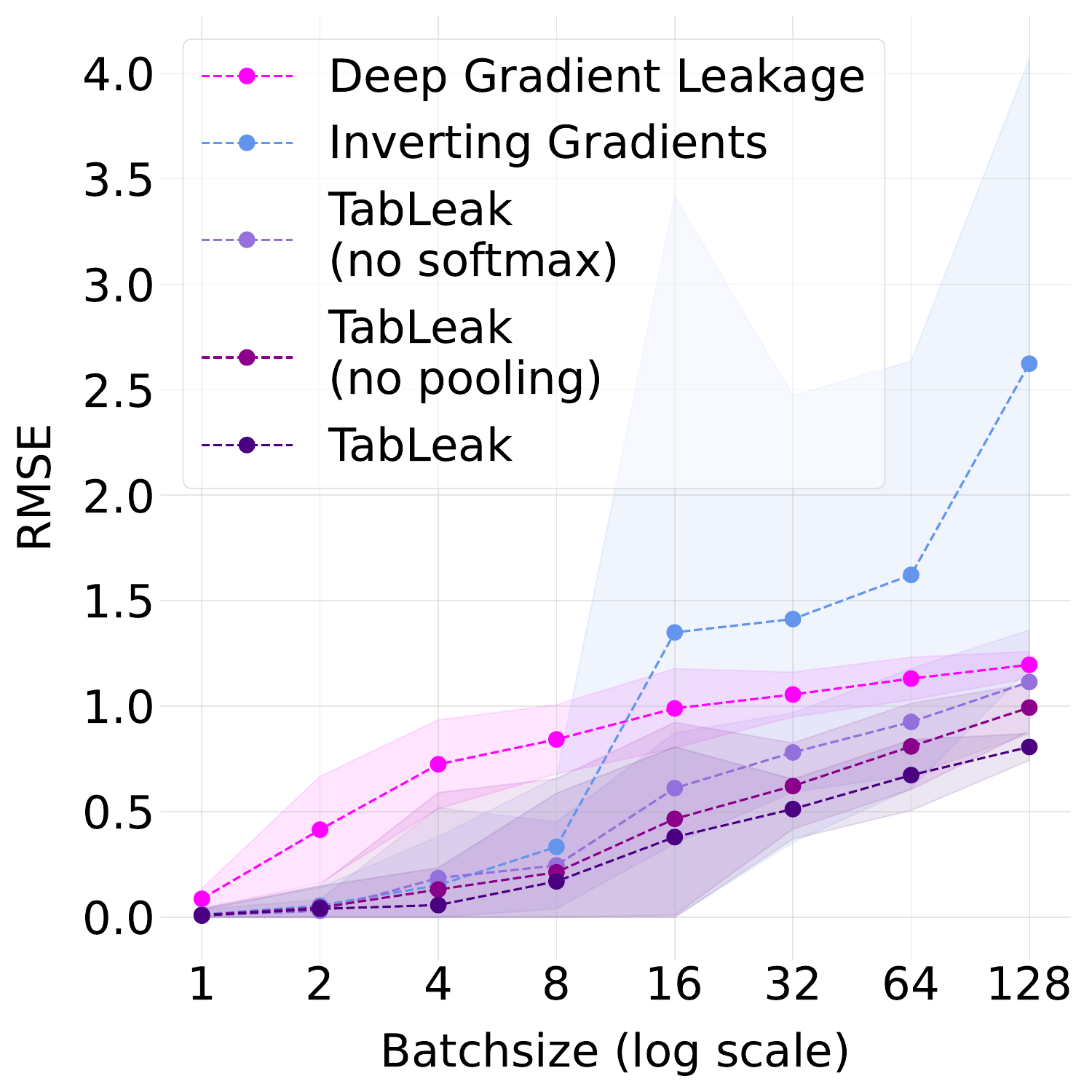}
		\subcaption{Health Heritage}
	\end{subfigure}
	\caption{The mean and standard deviation of the Root Mean Squared Error (RMSE) of the reconstructions of the continuous features on all four datasets over batch sizes.}
	\label{fig:rmse_cont}
\end{figure}

In order to examine the potential influence of our choice of reconstruction metric on the obtained results, we further measured the reconstruction quality of continuous features on the widely used Root Mean Squared Error (RMSE) metric as well. Concretely, we calculate the RMSE between the $L$ continuous features of our reconstruction $\hat{x}^C$ and the ground truth $x$ in a batch of size $n$ as:
\begin{equation}
	\text{RMSE}(x^C,\hat{x}^C)= \frac{1}{L} \sum_{i=1}^L \sqrt{\frac{1}{n} \sum_{j=1}^n ( x_{ij}^C - \hat{x}_{ij}^C)^2 }.
\end{equation}
As our results in \cref{fig:rmse_cont} demonstrate, \OurMethod{} achieves significantly lower RMSE than Inverting Gradients~\citep{Geiping2020} on large batch sizes, for all four datasets examined. This indicates that the strong results obtained by \OurMethod{} in the rest of the paper are not a consequence of our evaluation metric.

\subsection{Attacking High-Dimensional Datasets}
\label{appendix:attacking_high_dimensional_datasets}
To understand how gradient inversion attacks scale with the number of features and the encoded dimension of the dataset, we attack a synthetic dataset with $125$ discrete and $125$ continuous features (generated with the procedure explained in \cref{appendix:variance_study}), resulting in $1231$ dimensions when encoded, \ie around $12\times$ the dimension of Adult. We attack this setup both with \OurMethod{} and the baseline attack of \citet{Geiping2020}, reporting our results in \cref{table:high_dim_datasets}, for batch sizes $16$, $32$, and $64$. We can observe that while it is significantly harder to obtain high accuracy from such a high-dimensional dataset, \OurMethod{} still strongly outperforms both the baseline and random guessing, achieving at least $1.68\times$ higher accuracy.

\begin{table}[H]
	\caption{The mean and standard deviation of the attack accuracy [\%] on a synthetic dataset with $125$ discrete and $125$ continuous features ($1231$ dimensions one-hot encoded).}
	\label{table:high_dim_datasets}
	\centering
	\resizebox{0.5\columnwidth}{!}{
	\begin{tabular}{lccc}
		\toprule
		Batch & \OurMethod{} & Inverting Gradients & Random\\
		Size & & \citet{Geiping2020} &\\
		\midrule
		$16$  & $\mathbf{70.0 \pm 4.1}$ &  $39.5 \pm 3.2$ & $19.6 \pm 0.6$\\
		$32$  & $\mathbf{55.0 \pm 2.3}$ &  $30.6 \pm 2.1$ & $20.1 \pm 0.5$\\
		$64$  & $\mathbf{40.0 \pm 1.1}$  & $23.8 \pm 1.2$ & $20.7 \pm 0.4$\\
		\bottomrule
	\end{tabular}
	}
\end{table}

\clearpage
\subsection{Attacking During Training}
\label{appendix:attacking_during_training}
We evaluate how \OurMethod{} and the baseline attack of \citet{Geiping2020} perform when attacking a network not only at initialization, but after some epochs of federated training have already been conducted. We expect to confirm the findings of prior works~\citep{Geiping2020,Dimitrov22}, \ie that training degrades the performance of the attacks. Indeed, looking at our results collected in \cref{table:training_appx}, we can observe that on all four datasets training negatively impacts the attack success. Moreover, we confirm the partial observation made already on Adult in the main body of this paper, namely that \OurMethod{} preserves a high performance further into the training than the baseline attack, reinforcing the significance of the improvement \OurMethod{} brings over prior methods.

\begin{table}[H]
	\centering
	\begin{subtable}{.49\textwidth}
		\centering
		\begin{tabular}{lcc}
			\toprule
			Training & \OurMethod{} & Inverting Gradients \\
			Epochs & & \citet{Geiping2020} \\
			\midrule
			$1$  & $\mathbf{79.1 \pm 4.2}$ &  $67.8 \pm 2.1$  \\
			$5$  & $\mathbf{76.4 \pm 5.7}$ &   $64.5 \pm 3.8$  \\
			$10$  & $\mathbf{74.5 \pm 5.7}$  & $60.9 \pm 3.7$ \\
			$15$  & $\mathbf{64.5 \pm 7.1}$  & $57.8 \pm 4.0$ \\
			\bottomrule
		\end{tabular}
		\subcaption{Adult}
	\end{subtable}%
	\begin{subtable}{.49\textwidth}
		\centering
		\begin{tabular}{lcc}
			\toprule
			Training & \OurMethod{} & Inverting Gradients \\
			Epochs & & \citet{Geiping2020} \\
			\midrule
			$1$  & $\mathbf{94.2 \pm 4.4}$ &  $78.9 \pm 4.0$  \\
			$5$  & $\mathbf{92.6 \pm 5.0}$ &   $77.5 \pm 5.0$  \\
			$10$  & $\mathbf{92.2 \pm 3.7}$  & $76.0 \pm 4.6$ \\
			$15$  & $\mathbf{89.7 \pm 4.0}$  & $72.4 \pm 4.1$ \\
			\bottomrule
		\end{tabular}
		\subcaption{German Credit}
	\end{subtable}
	\par\bigskip
	\begin{subtable}{.49\textwidth}
		\centering
		\begin{tabular}{lcc}
			\toprule
			Training & \OurMethod{} & Inverting Gradients \\
			Epochs & & \citet{Geiping2020} \\
			\midrule
			$1$  & $\mathbf{85.7 \pm 4.2}$ &  $71.4 \pm 2.2$  \\
			$5$  & $\mathbf{78.0 \pm 4.6}$ &   $70.1 \pm 3.4$  \\
			$10$  & $\mathbf{74.9 \pm 2.7}$  & $68.3 \pm 3.5$ \\
			$15$  & $\mathbf{74.9 \pm 3.1}$  & $66.7 \pm 3.9$ \\
			\bottomrule
		\end{tabular}
		\subcaption{Lawschool Admissions}
	\end{subtable}
	\begin{subtable}{.49\textwidth}
		\centering
		\begin{tabular}{lcc}
			\toprule
			Training & \OurMethod{} & Inverting Gradients \\
			Epochs & & \citet{Geiping2020} \\
			\midrule
			$1$  & $\mathbf{69.5 \pm 4.5}$ &  $58.0 \pm 4.2$  \\
			$5$  & $\mathbf{69.3 \pm 5.3}$ &   $54.8 \pm 4.5$  \\
			$10$  & $\mathbf{64.8 \pm 6.0}$  & $52.1 \pm 3.1$ \\
			$15$  & $\mathbf{62.3 \pm 6.9}$  & $50.7 \pm 3.3$ \\
			\bottomrule
		\end{tabular}
		\subcaption{Health Heritage}
	\end{subtable}
	\caption{The mean and standard deviation of the attack accuracy [\%] attacking a network at the $1^{\text{st}}$, $5^{\text{th}}$, $10^{\text{th}}$, and $15^{\text{th}}$ epoch of FedSGD training on batch size $32$.}
	\label{table:training_appx}
\end{table}

\clearpage
\subsection{Impact of Attack Iterations}
\label{appendix:impact_of_attack_iterations}
We conduct an ablation study over the attack iterations to understand its influence on the attack success. In all the presented experiments we chose to run the attacks for $1\,500$ iterations before reporting our results; our goal here is to understand how this choice influenced our results. 
\cref{table:attack_iters_appx} shows that although the baseline attack is faster to converge, it tops out at a significantly lower accuracy level than \OurMethod{} on all datasets, while \OurMethod{} manages to improve on the accuracy presented in the paper on most cases, by allowing for more iterations. This further underlines the significant performance difference between \OurMethod{} and prior gradient inversion attacks.

\begin{table}[H]
	\centering
	\begin{subtable}{.49\textwidth}
		\centering
		\begin{tabular}{lcc}
			\toprule
			Attack & \OurMethod{} & Inverting Gradients \\
			Iterations & & \citet{Geiping2020} \\
			\midrule
			$10$  & $46.5 \pm 1.44$ &  $60.5 \pm 1.37$  \\
			$500$  & $65.8 \pm 3.86$ &   $66.8 \pm 2.35$  \\
			$1\,500$  & $78.8 \pm 4.25$  & $66.4 \pm 2.57$ \\
			$10\,000$  & $\mathbf{83.7 \pm 2.91}$  & $67.6 \pm 2.81$ \\
			\bottomrule
		\end{tabular}
		\subcaption{Adult}
	\end{subtable}%
	\begin{subtable}{.49\textwidth}
		\centering
		\begin{tabular}{lcc}
			\toprule
			Attack & \OurMethod{} & Inverting Gradients \\
			Iterations & & \citet{Geiping2020} \\
			\midrule
			$10$  & $55.4 \pm 1.62$ &  $62.0 \pm 1.14$  \\
			$500$  & $79.2 \pm 2.51$ &  $68.4 \pm 2.28$  \\
			$1\,500$  & $82.1 \pm 2.56$  & $68.7 \pm 2.75$ \\
			$10\,000$  & $\mathbf{84.7 \pm 2.92}$  & $68.8 \pm 3.18$ \\
			\bottomrule
		\end{tabular}
		\subcaption{German Credit}
	\end{subtable}
	\par\bigskip
	\begin{subtable}{.49\textwidth}
		\centering
		\begin{tabular}{lcc}
			\toprule
			Attack & \OurMethod{} & Inverting Gradients \\
			Iterations & & \citet{Geiping2020} \\
			\midrule
			$10$  & $61.0 \pm 1.80$ &  $65.7 \pm 1.81$  \\
			$500$  & $78.1 \pm 3.33$ &   $71.8 \pm 2.42$  \\
			$1\,500$  & $\mathbf{86.9 \pm 3.49}$  & $71.7 \pm 1.99$ \\
			$10\,000$  & $86.4 \pm 4.29$  & $71.8 \pm 2.53$ \\
			\bottomrule
		\end{tabular}
		\subcaption{Lawschool Admissions}
	\end{subtable}
	\begin{subtable}{.49\textwidth}
		\centering
		\begin{tabular}{lcc}
			\toprule
			Attack & \OurMethod{} & Inverting Gradients \\
			Iterations & & \citet{Geiping2020} \\
			\midrule
			$10$  & $29.8 \pm 1.31$ &  $51.9 \pm 1.79$  \\
			$500$  & $59.6 \pm 3.57$ &   $57.5 \pm 3.76$  \\
			$1\,500$  & $\mathbf{71.3 \pm 5.25}$  & $57.6 \pm 4.76$ \\
			$10\,000$  & $\mathbf{71.3 \pm 3.65}$  & $54.0 \pm 6.12$ \\
			\bottomrule
		\end{tabular}
		\subcaption{Health Heritage}
	\end{subtable}
	\caption{The mean and standard deviation of the attack accuracy [\%] on batch size $32$ over attack iterations.}
	\label{table:attack_iters_appx}
\end{table}

\clearpage
\section{All Main Results}
\label{appendix:all_main_results}
In this subsection, we include all the results presented in the main part of this paper for the Adult dataset alongside with the corresponding additional results on the German Credit, Lawschool Admissions, and the Health Heritage datasets.

\subsection{Full FedSGD Results on all Datasets}
\label{appendix:full_fedsgd_results_on_all_datasets}
In \cref{table:adult_results_all}, \cref{table:german_results_all}, \cref{table:lawschool_results_all}, and \cref{table:health_results_all} we provide the full attack results of our method compared to Inverting Gradients~\citep{Geiping2020} and the random baseline on the Adult, German Credit, Lawschool Admissions, and Health Heritage datasets, respectively. Looking at the results for all datasets, we can confirm the observations made in \cref{sec:results}, \ie (i) the lower batch sizes are vulnerable to any non-trivial attack, (ii) not knowing the ground truth labels does not significantly disadvantage the attacker for larger batch sizes, and (iii) \OurMethod{} provides a strong improvement over the baselines for practically relevant batch sizes over all datasets examined.

\begin{table}
	\caption{The mean inversion accuracy [\%] and standard deviation of different methods over varying batch sizes with given true labels (top) and with reconstructed labels (bottom) on the \textbf{Adult} dataset.}
	\label{table:adult_results_all}
	\centering
	\resizebox{.99\columnwidth}{!}{\begin{tabular}{llcccccc}
    \toprule
    Label & Batch & \OurMethod{} & \OurMethod{} & \OurMethod{} & Inverting Gradients & Deep Gradient Leakage & Random\\
    & Size & & (no pooling) & (no softmax) & \citet{Geiping2020} & \citet{Zhu2019} &\\
    \midrule
    \multirow{8}{*}{True $y$}
    & $1$  & $99.4 \pm 2.8$ &  $99.1 \pm 4.4$ &  $\mathbf{100.0 \pm 0.0}$ & $\mathbf{100.0 \pm 0.0}$ & $97.0 \pm 9.3$ & $43.3 \pm 11.8$\\
    & $2$  & $99.3 \pm 5.0$ &  $99.2 \pm 5.5$ &  $\mathbf{99.6 \pm 1.3}$ & $97.6 \pm 6.9$ & $77.5 \pm 12.8$ & $47.1 \pm 7.9$\\
    & $4$  & $98.1 \pm 4.7$ &  $96.6 \pm 7.8$ &  $\mathbf{98.7 \pm 3.4}$ & $96.4 \pm 7.2$ & $65.3 \pm 8.2$ & $49.8 \pm 4.9$\\
    & $8$  & $\mathbf{95.2 \pm 8.8}$ &  $92.5 \pm 11.8$ &  $91.3 \pm 7.1$ & $91.1 \pm 7.3$ & $61.2 \pm 4.7$ & $53.9 \pm 4.4$\\
    & $16$  & $\mathbf{89.9 \pm 7.3}$ &  $85.3 \pm 9.7$ &  $79.0 \pm 4.0$ & $75.0 \pm 5.2$ & $60.2 \pm 3.3$ & $55.1 \pm 3.9$\\
    & $32$  & $\mathbf{79.3 \pm 4.5}$ &  $74.3 \pm 4.5$ &  $70.8 \pm 3.3$ & $66.6 \pm 3.5$ & $60.8 \pm 1.9$ & $58.0 \pm 2.9$\\
    & $64$  & $\mathbf{73.4 \pm 3.0}$  & $68.9 \pm 3.1$ & $67.3 \pm 3.2$ & $62.5 \pm 3.1$ & $61.3 \pm 1.4$ & $59.0 \pm 3.2$\\
    & $128$  & $\mathbf{71.4 \pm 1.2}$ & $67.4 \pm 1.4$ &  $65.2 \pm 2.1$ &   $59.5 \pm 2.1$ & $62.9 \pm 1.0$ & $61.2 \pm 3.1$\\
    \midrule
    \multirow{8}{*}{Rec. $\hat{y}$}
    & $1$  & $99.4 \pm 2.8$ &  $99.3 \pm 3.6$ &  $\mathbf{100.0 \pm 0.0}$ & $\mathbf{100.0 \pm 0.0}$ & $98.9 \pm 2.6$ & $43.3 \pm 11.8$\\
    & $2$  & $98.1 \pm 9.6$ &  $98.1 \pm 9.6$ &  $\mathbf{98.7 \pm 7.1}$ & $95.9 \pm 11.5$ & $77.9 \pm 14.1$ & $47.1 \pm 7.9$\\
    & $4$  & $89.6 \pm 13.5$ &  $87.8 \pm 15.3$ &  $\mathbf{89.8 \pm 13.0}$ & $87.9 \pm 13.7$ & $58.1 \pm 12.4$ & $49.8 \pm 4.9$\\
    & $8$  & $\mathbf{86.7 \pm 12.2}$ &  $83.8 \pm 13.6$ &  $82.7 \pm 10.5$ & $83.3 \pm 9.7$ & $56.1 \pm 5.4$ & $53.9 \pm 4.4$\\
    & $16$  & $\mathbf{83.0 \pm 7.7}$ &  $78.6 \pm 8.1$ &  $76.4 \pm 5.4$ & $73.0 \pm 3.5$ & $57.2 \pm 3.4$ & $55.1 \pm 3.9$\\
    & $32$  & $\mathbf{76.9 \pm 4.8}$  & $72.4 \pm 4.8$ & $68.9 \pm 4.2$ & $66.3 \pm 3.4$ & $58.4 \pm 2.5$ & $58.0 \pm 2.9$\\
    & $64$  & $\mathbf{72.8 \pm 3.3}$ & $68.5 \pm 3.5$ & $66.8 \pm 2.9$ & $63.1 \pm 3.2$ & $60.1 \pm 1.7$ & $59.0 \pm 3.2$\\
    & $128$ & $\mathbf{71.4 \pm 1.3}$ & $67.5 \pm 1.5$ & $65.0 \pm 2.2$ & $59.5 \pm 2.1$ & $62.3 \pm 1.0$ & $61.2 \pm 3.1$\\
    \bottomrule
\end{tabular}}
\end{table}

\begin{table}
	\caption{The mean inversion accuracy [\%] and standard deviation of different methods over varying batch sizes with given true labels (top) and with reconstructed labels (bottom) on the \textbf{German Credit} dataset.}
	\label{table:german_results_all}
	\centering
	\resizebox{.99\columnwidth}{!}{\begin{tabular}{llcccccc}
    \toprule
    Label & Batch & \OurMethod{} & \OurMethod{} & \OurMethod{} & Inverting Gradients & Deep Gradient Leakage & Random\\
		& Size & & (no pooling) & (no softmax) & \citet{Geiping2020} & \citet{Zhu2019} &\\
    \midrule
    \multirow{8}{*}{True $y$}
    & $1$    & $\mathbf{100.0 \pm 0.0}$ &  $\mathbf{100.0 \pm 0.0}$ &  $\mathbf{100.0 \pm 0.0}$ & $\mathbf{100.0 \pm 0.0}$ & $100.0 \pm 0.0$ & $43.9 \pm 9.8$\\
    & $2$    & $\mathbf{100.0 \pm 0.0}$ &  $\mathbf{100.0 \pm 0.0}$ &  $99.9 \pm 0.4$ & $98.0 \pm 7.1$ & $84.2 \pm 14.9$ & $45.1 \pm 6.6$\\
    & $4$    & $\mathbf{99.9 \pm 0.4}$ &  $99.5 \pm 2.8$ &  $99.6 \pm 1.2$ & $97.8 \pm 6.0$ & $71.0 \pm 6.9$ & $50.3 \pm 4.5$\\
    & $8$    & $\mathbf{99.6 \pm 1.2}$ &  $99.1 \pm 2.4$ &  $98.4 \pm 2.2$ & $96.1 \pm 5.2$ & $64.1 \pm 2.7$ & $51.8 \pm 3.2$\\
    & $16$   & $\mathbf{96.3 \pm 3.4}$ &  $93.7 \pm 4.5$ &  $85.1 \pm 3.6$ & $79.3 \pm 4.4$ & $63.1 \pm 2.1$ & $54.5 \pm 3.0$\\
    & $32$   & $\mathbf{84.2 \pm 2.8}$ &  $80.1 \pm 3.2$ &  $72.8 \pm 1.9$ & $69.7 \pm 2.2$ & $63.4 \pm 1.4$ & $56.8 \pm 2.2$\\
    & $64$   & $\mathbf{74.4 \pm 1.4}$  & $71.8 \pm 1.6$ & $69.7 \pm 1.3$ & $66.6 \pm 1.8$ & $64.0 \pm 1.0$ & $59.4 \pm 1.9$\\
    & $128$  & $\mathbf{72.3 \pm 0.9}$ & $69.8 \pm 0.7$ &  $68.4 \pm 1.5$ & $64.5 \pm 1.5$ & $65.3 \pm 0.7$ & $61.0 \pm 2.1$\\
    \midrule
    \multirow{8}{*}{Rec. $\hat{y}$} %
    & $1$  & $\mathbf{100.0 \pm 0.0}$ &  $\mathbf{100.0 \pm 0.0}$ &  $\mathbf{100.0 \pm 0.0}$ & $\mathbf{100.0 \pm 0.0}$ & $99.1 \pm 6.3$ & $43.9 \pm 9.8$\\
    & $2$  & $\mathbf{100.0 \pm 0.0}$ &  $\mathbf{100.0 \pm 0.0}$ &  $99.9 \pm 0.4$ & $98.8 \pm 5.2$ & $86.2 \pm 14.3$ & $45.1 \pm 6.6$\\
    & $4$  & $\mathbf{99.5 \pm 3.2}$ &  $98.7 \pm 4.7$ &  $99.3 \pm 2.9$ & $97.4 \pm 6.4$ & $73.0 \pm 7.4$ & $50.3 \pm 4.5$\\
    & $8$  & $\mathbf{97.2 \pm 6.3}$ &  $96.1 \pm 7.6$ &  $96.2 \pm 6.4$ & $94.8 \pm 6.5$ & $63.5 \pm 4.8$ & $51.8 \pm 3.2$\\
    & $16$  & $\mathbf{92.0 \pm 6.5}$ &  $90.0 \pm 6.6$ &  $83.5 \pm 4.0$ & $77.9 \pm 4.6$ & $61.4 \pm 2.9$ & $54.5 \pm 3.0$\\
    & $32$  & $\mathbf{81.9 \pm 3.4}$  & $78.4 \pm 3.4$ & $71.8 \pm 1.9$ & $69.1 \pm 2.1$ & $62.1 \pm 1.4$ & $56.8 \pm 2.2$\\
    & $64$  & $\mathbf{73.8 \pm 1.5}$ & $71.4 \pm 1.3$ & $69.5 \pm 1.2$ & $66.5 \pm 1.7$ & $63.5 \pm 1.0$ & $59.4 \pm 1.9$\\
    & $128$ & $\mathbf{72.3 \pm 0.9}$ & $69.8 \pm 0.7$ & $68.2 \pm 1.6$ & $64.4 \pm 1.6$ & $65.0 \pm 0.6$ & $61.0 \pm 2.1$\\
    \bottomrule
\end{tabular}}
\end{table}

\begin{table}
	\caption{The mean inversion accuracy [\%] and standard deviation of different methods over varying batch sizes with given true labels (top) and with reconstructed labels (bottom) on the \textbf{Lawschool Admissions} dataset.}
	\label{table:lawschool_results_all}
	\centering
	\resizebox{.99\columnwidth}{!}{\begin{tabular}{llcccccc}
    \toprule
    Label & Batch & \OurMethod{} & \OurMethod{} & \OurMethod{} & Inverting Gradients & Deep Gradient Leakage & Random\\
		& Size & & (no pooling) & (no softmax) & \citet{Geiping2020} & \citet{Zhu2019} &\\
    \midrule
    \multirow{8}{*}{True $y$}
    & $1$    & $\mathbf{100.0 \pm 0.0}$ &  $\mathbf{100.0 \pm 0.0}$ &  $\mathbf{100.0 \pm 0.0}$ & $\mathbf{100.0 \pm 0.0}$ & $97.7 \pm 9.6$ & $38.9 \pm 14.6$\\
    & $2$    & $\mathbf{100.0 \pm 0.0}$ &  $\mathbf{100.0 \pm 0.0}$ &  $99.9 \pm 1.0$ & $96.3 \pm 10.4$ & $84.6 \pm 16.7$ & $38.4 \pm 11.5$\\
    & $4$    & $\mathbf{100.0 \pm 0.0}$ &  $\mathbf{100.0 \pm 0.0}$ &  $99.6 \pm 2.1$ & $97.6 \pm 6.9$ & $76.6 \pm 12.5$ & $43.2 \pm 7.2$\\
    & $8$    & $\mathbf{98.9 \pm 3.8}$ &  $98.5 \pm 4.5$ &  $95.6 \pm 5.1$ & $94.5 \pm 5.8$ & $68.5 \pm 5.4$ & $49.4 \pm 4.6$\\
    & $16$   & $\mathbf{95.0 \pm 5.8}$ &  $93.2 \pm 6.4$ &  $81.3 \pm 4.4$ & $77.3 \pm 5.5$ & $65.8 \pm 3.2$ & $53.0 \pm 3.1$\\
    & $32$   & $\mathbf{84.9 \pm 4.0}$ &  $82.0 \pm 3.7$ &  $73.1 \pm 2.5$ & $71.0 \pm 2.8$ & $67.9 \pm 2.3$ & $57.6 \pm 2.3$\\
    & $64$   & $\mathbf{78.1 \pm 2.17}$  & $76.6 \pm 2.2$ & $73.0 \pm 2.1$ & $71.7 \pm 2.2$ & $70.4 \pm 1.4$ & $60.4 \pm 2.2$\\
    & $128$  & $\mathbf{77.2 \pm 1.1}$ & $75.9 \pm 1.2$ &  $73.5 \pm 2.8$ & $71.8 \pm 2.7$ & $73.4 \pm 0.9$ & $63.4 \pm 1.5$\\
    \midrule
    \multirow{8}{*}{Rec. $\hat{y}$}
    & $1$  & $\mathbf{100.0 \pm 0.0}$ &  $\mathbf{100.0 \pm 0.0}$ &  $\mathbf{100.0 \pm 0.0}$ & $\mathbf{100.0 \pm 0.0}$ & $100.0 \pm 0.0$ & $38.9 \pm 14.6$\\
    & $2$  & $\mathbf{99.1 \pm 6.0}$ &  $\mathbf{99.1 \pm 6.0}$ &  $98.7 \pm 8.0$ & $95.9 \pm 12.0$ & $84.3 \pm 17.2$ & $38.4 \pm 11.5$\\
    & $4$  & $\mathbf{99.5 \pm 3.5}$ &  $99.1 \pm 4.3$ &  $98.7 \pm 5.9$ & $96.8 \pm 8.5$ & $79.9 \pm 12.6$ & $43.2 \pm 7.2$\\
    & $8$  & $\mathbf{95.9 \pm 7.8}$ &  $\mathbf{95.9 \pm 8.0}$ &  $93.4 \pm 7.8$ & $91.9 \pm 7.9$ & $66.9 \pm 7.2$ & $49.4 \pm 4.6$\\
    & $16$  & $\mathbf{91.2 \pm 7.4}$ &  $88.5 \pm 8.7$ &  $80.4 \pm 5.0$ & $77.4 \pm 5.4$ & $64.9 \pm 3.9$ & $53.0 \pm 3.1$\\
    & $32$  & $\mathbf{83.1 \pm 4.2}$  & $81.2 \pm 4.5$ & $72.9 \pm 2.5$ & $71.0 \pm 2.0$ & $66.1 \pm 2.0$ & $57.6 \pm 2.3$\\
    & $64$  & $\mathbf{77.3 \pm 2.2}$ & $75.9 \pm 2.0$ & $72.5 \pm 2.1$ & $71.5 \pm 2.4$ & $69.3 \pm 1.2$ & $60.4 \pm 2.2$\\
    & $128$ & $\mathbf{77.0 \pm 1.1}$ & $75.8 \pm 1.2$ & $73.8 \pm 2.5$ & $71.8 \pm 2.8$ & $72.8 \pm 1.0$ & $63.4 \pm 1.5$\\
    \bottomrule
\end{tabular}}
\end{table}

\begin{table}
	\caption{The mean inversion accuracy [\%] and standard deviation of different methods over varying batch sizes with given true labels (top) and with reconstructed labels (bottom) on the \textbf{Health Heritage} dataset.}
	\label{table:health_results_all}
	\centering
	\resizebox{.99\columnwidth}{!}{\begin{tabular}{llcccccc}
    \toprule
    Label & Batch & \OurMethod{} & \OurMethod{} & \OurMethod{} & Inverting Gradients & Deep Gradient Leakage & Random\\
		& Size & & (no pooling) & (no softmax) & \citet{Geiping2020} & \citet{Zhu2019} &\\
    \midrule
    \multirow{8}{*}{True $y$}
    & $1$    & $\mathbf{99.8 \pm 1.6}$ &  $\mathbf{99.8 \pm 1.6}$ &  $\mathbf{99.8 \pm 1.6}$ & $\mathbf{99.8 \pm 1.6}$ & $97.3 \pm 6.4$ & $34.8 \pm 13.1$\\
    & $2$    & $97.6 \pm 7.9$ &  $97.1 \pm 9.3$ &  $\mathbf{98.9 \pm 3.3}$ & $97.9 \pm 5.6$ & $70.7 \pm 16.9$ & $36.9 \pm 9.8$\\
    & $4$    & $\mathbf{97.9 \pm 7.7}$ &  $96.4 \pm 10.8$ &  $95.4 \pm 7.8$ & $95.6 \pm 8.1$ & $52.0 \pm 7.2$ & $37.0 \pm 5.3$\\
    & $8$    & $\mathbf{95.5 \pm 9.2}$ &  $93.1 \pm 11.5$ &  $88.6 \pm 10.8$ & $86.2 \pm 9.0$ & $50.1 \pm 3.9$ & $39.2 \pm 3.8$\\
    & $16$   & $\mathbf{85.4 \pm 9.9}$ &  $79.8 \pm 10.5$ &  $68.3 \pm 5.3$ & $63.6 \pm 5.5$ & $50.8 \pm 2.2$ & $41.4 \pm 3.7$\\
    & $32$   & $\mathbf{70.8 \pm 4.5}$ &  $65.5 \pm 4.2$ &  $61.8 \pm 4.0$ & $57.7 \pm 4.1$ & $51.6 \pm 1.7$ & $43.4 \pm 2.8$\\
    & $64$   & $\mathbf{65.5 \pm 2.8}$  & $61.3 \pm 2.7$ & $61.2 \pm 4.4$ & $57.4 \pm 4.7$ & $54.0 \pm 1.5$ & $45.0 \pm 3.7$\\
    & $128$  & $\mathbf{63.5 \pm 1.7}$ & $59.3 \pm 1.6$ &  $58.6 \pm 4.4$ & $55.6 \pm 4.8$ & $55.4 \pm 0.8$ & $46.8 \pm 3.2$\\
    \midrule
    \multirow{8}{*}{Rec. $\hat{y}$}
    & $1$  & $\mathbf{99.8 \pm 1.6}$ &  $\mathbf{99.8 \pm 1.6}$ &  $\mathbf{99.8 \pm 1.6}$ & $99.6 \pm 2.5$ & $96.1 \pm 7.7$ & $34.8 \pm 13.1$\\
    & $2$  & $95.1 \pm 14.2$ &  $95.3 \pm 13.7$ &  $\mathbf{95.4 \pm 14.2}$ & $92.5 \pm 16.9$ & $68.2 \pm 19.6$ & $36.9 \pm 9.8$\\
    & $4$  & $\mathbf{86.3 \pm 21.1}$ &  $85.0 \pm 22.2$ &  $83.4 \pm 19.6$ & $83.5 \pm 20.7$ & $48.2 \pm 11.7$ & $37.0 \pm 5.3$\\
    & $8$  & $\mathbf{81.5 \pm 16.0}$ &  $77.6 \pm 16.9$ &  $76.8 \pm 13.4$ & $74.5 \pm 13.8$ & $48.0 \pm 6.3$ & $39.2 \pm 3.8$\\
    & $16$  & $\mathbf{75.3 \pm 13.1}$ &  $71.2 \pm 12.9$ &  $65.2 \pm 7.6$ & $60.9 \pm 6.3$ & $49.6 \pm 4.5$ & $41.4 \pm 3.7$\\
    & $32$  & $\mathbf{65.5 \pm 5.6}$  & $62.0 \pm 5.2$ & $60.3 \pm 4.1$ & $56.9 \pm 4.0$ & $51.0 \pm 3.2$ & $43.4 \pm 2.8$\\
    & $64$  & $\mathbf{63.9 \pm 2.9}$ & $60.9 \pm 2.5$ & $61.2 \pm 4.4$ & $57.7 \pm 4.7$ & $53.7 \pm 1.3$ & $45.0 \pm 3.7$\\
    & $128$ & $\mathbf{63.8 \pm 1.8}$ & $60.3 \pm 1.9$ & $58.8 \pm 4.7$ & $55.7 \pm 5.0$ & $55.4 \pm 0.9$ & $46.8 \pm 3.2$\\
    \bottomrule
\end{tabular}}
\end{table}

\subsection{Categorical vs. Continuous Features on all Datasets}
\label{appendix:categorical_vs_continuous_features_on_all_datasets}
\begin{figure}
	\centering
	\begin{subfigure}{.5\textwidth}
		\centering
		\includegraphics[width=0.9\textwidth]{figures/ADULT_128_catvscont_0_46.pdf}
		\subcaption{Adult}
	\end{subfigure}%
	\begin{subfigure}{.5\textwidth}
		\centering
		\includegraphics[width=0.9\textwidth]{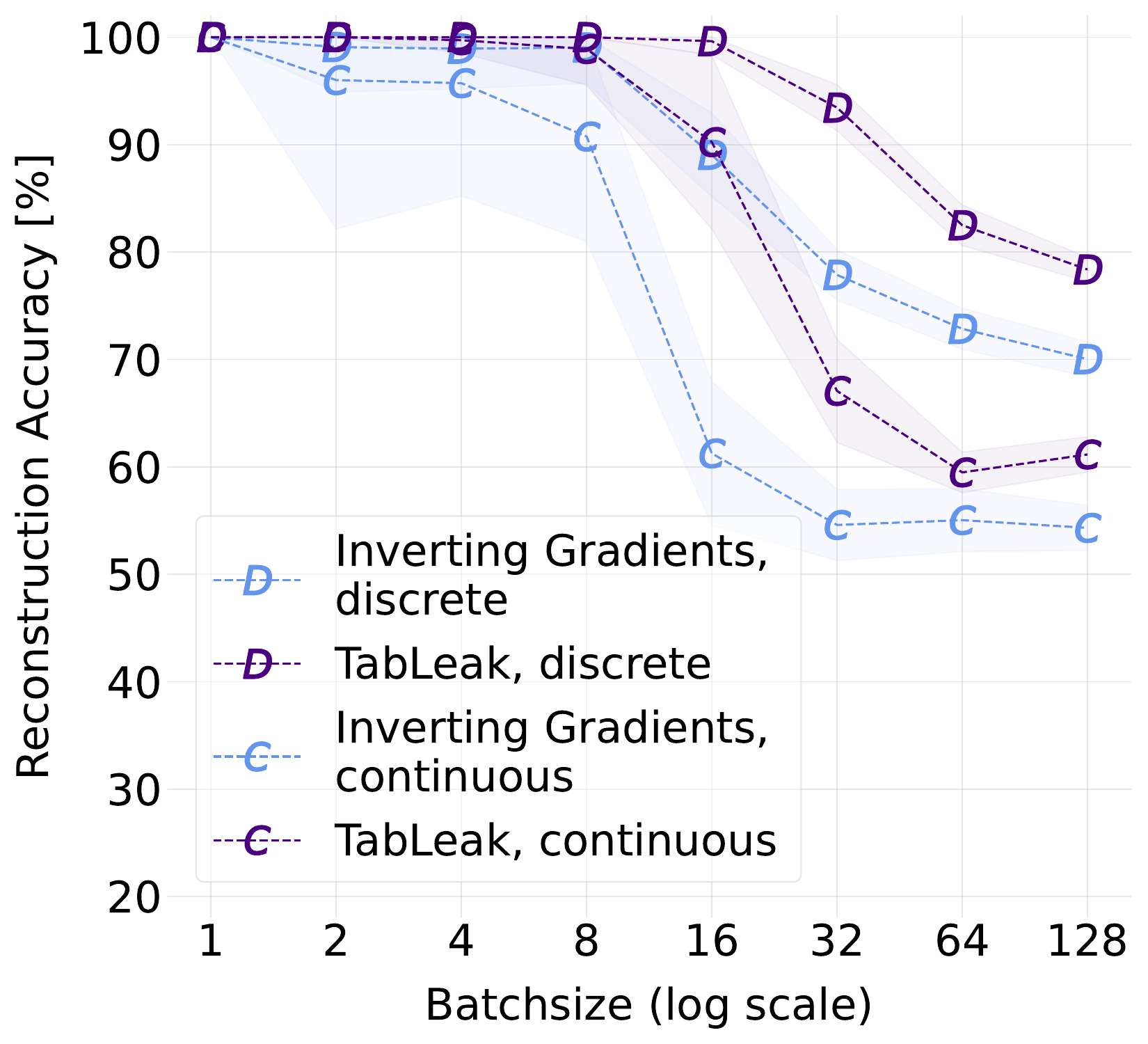}
		\subcaption{German Credit}
	\end{subfigure}
	\begin{subfigure}{.5\textwidth}
		\centering
		\includegraphics[width=0.9\textwidth]{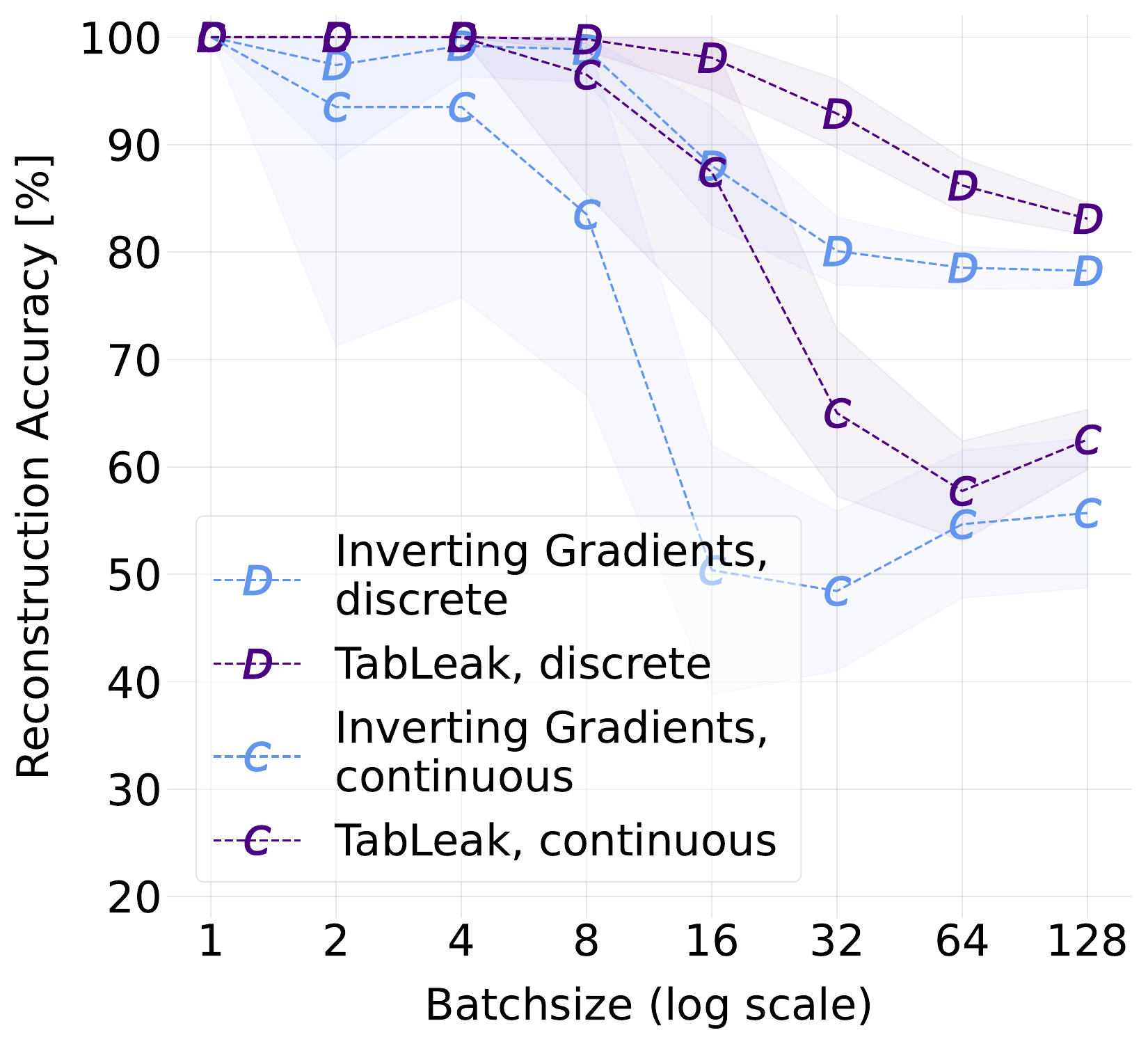}
		\subcaption{Lawschool Admissions}
	\end{subfigure}%
	\begin{subfigure}{.5\textwidth}
		\centering
		\includegraphics[width=0.9\textwidth]{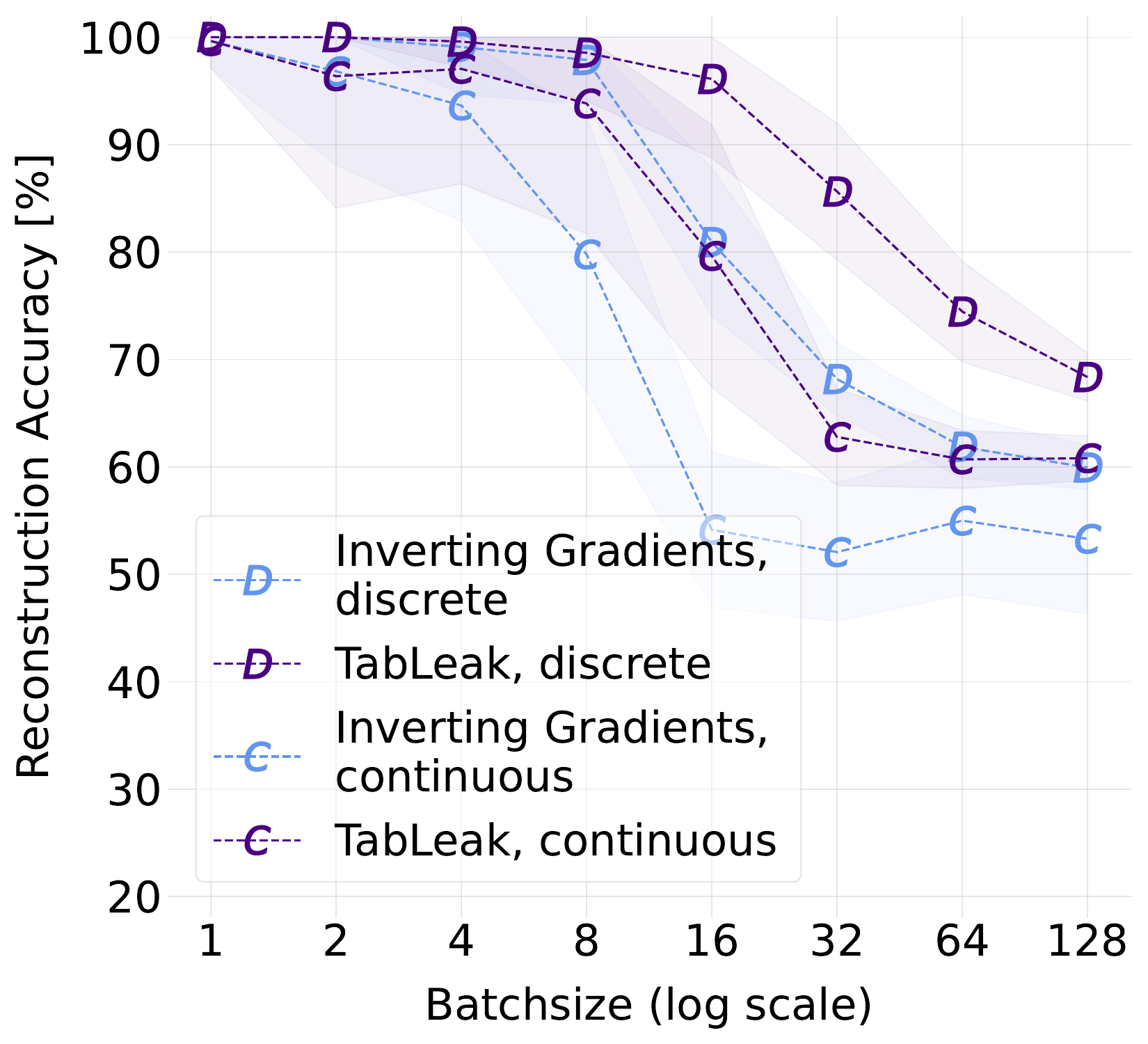}
		\subcaption{Health Heritage}
	\end{subfigure}
\caption{Mean reconstruction accuracy curves with corresponding standard deviations over varying batch size, separately for the discrete and the continuous features on all four datasets.}
\label{fig:cat_vs_cont_all}
\end{figure}

In \cref{fig:cat_vs_cont_all}, we compare the reconstruction accuracy of the continuous and the discrete features on all four datasets. We confirm our observations, shown in \cref{fig:attack_errors_adult_cat_cont} in the main text, that a strong dichotomy between continuous and discrete feature reconstruction accuracy exists on all 4 datasets.

\subsection{Federated Averaging Results on all Datasets}
\label{appendix:federated_averaging_results_on_all_datasets}
In \cref{table:rec_fedavg_adult}, \cref{table:rec_fedavg_german}, \cref{table:rec_fedavg_lawschool}, and \cref{table:rec_fedavg_health} we present our results on attacking the clients in FedAvg training on the Adult, German Credit, Lawschool Submissions, and Health Heritage datasets, respectively. We described the details of the experiment in \cref{appendix:further_experimental_details} above. Confirming our conclusions drawn in the main part of this manuscript, we observe that \OurMethod{} achieves non-trivial reconstruction accuracy over all settings and even for large numbers of updates, while the baseline attack often fails to outperform random guessing, when the number of local updates is increased. 

\begin{table}
	\caption{Mean and standard deviation of the inversion accuracy [\%] with local dataset size of $32$ in FedAvg training on the \textbf{Adult} dataset. The accuracy of the random baseline for $32$ datapoints is $58.0\pm 2.9$.}
	\label{table:rec_fedavg_adult}
	\centering
	\resizebox{0.75\columnwidth}{!}{
	\begin{tabular}{lcccccc}
		\toprule
		 & \multicolumn{3}{c}{\OurMethod{}} & \multicolumn{3}{c}{Inverting Gradients~\citep{Geiping2020}}\\
		\cmidrule[1pt](l{5pt}r{5pt}){2-4}\cmidrule[1pt](l{5pt}r{5pt}){5-7}
		 n. batches & 1 epoch & 5 epochs & 10 epochs & 1 epoch & 5 epochs & 10 epochs \\
		\midrule
		
		 1 & $\mathbf{80.7 \pm 3.8}$ & $\mathbf{75.8 \pm 3.3}$ & $\mathbf{72.8 \pm 3.2}$ & $65.2 \pm 2.7$ & $56.1 \pm 4.1$ & $53.2 \pm 4.2$ \\
		 2 & $\mathbf{79.2 \pm 4.2}$ & $\mathbf{75.6 \pm 2.7}$  & $\mathbf{73.1 \pm 5.0}$ & $64.8 \pm 3.3$ & $56.4 \pm 4.8$ & $56.2 \pm 4.8$ \\
		 4 & $\mathbf{79.7 \pm 3.6}$ & $\mathbf{76.2 \pm 3.0}$ & $\mathbf{73.7 \pm 3.6}$ & $64.8 \pm 3.4$ & $58.7 \pm 4.6$ & $56.6 \pm 5.0$ \\
		\bottomrule
	\end{tabular}
	}
\end{table}

\begin{table}
	\caption{Mean and standard deviation of the inversion accuracy [\%] with local dataset size of $32$ in FedAvg training on the \textbf{German Credit} dataset. The accuracy of the random baseline for $32$ datapoints is $56.9 \pm 2.1$.}
	\label{table:rec_fedavg_german}
	\centering
	\resizebox{0.75\columnwidth}{!}{
	\begin{tabular}{lcccccc}
		\toprule
		 & \multicolumn{3}{c}{\OurMethod{}} & \multicolumn{3}{c}{Inverting Gradients~\citep{Geiping2020}}\\
		\cmidrule[1pt](l{5pt}r{5pt}){2-4}\cmidrule[1pt](l{5pt}r{5pt}){5-7}
		n. batches & 1 epoch & 5 epochs & 10 epochs & 1 epoch & 5 epochs & 10 epochs \\
		\midrule
		
		 1 & $\mathbf{96.0 \pm 3.4}$ & $\mathbf{87.3 \pm 8.2}$ & $\mathbf{85.9 \pm 6.2}$ & $78.2 \pm 4.6$ & $65.4 \pm 6.2$ & $62.5 \pm 6.1$ \\
		 2 & $\mathbf{96.2 \pm 3.0}$ & $\mathbf{87.2 \pm 5.4}$  & $\mathbf{85.4 \pm 9.0}$ & $78.3 \pm 5.8$ & $68.8 \pm 6.6$ & $63.4 \pm 4.8$ \\
		 4 & $\mathbf{96.1 \pm 3.6}$ & $\mathbf{85.3 \pm 8.0}$ & $\mathbf{83.8 \pm 8.1}$ & $79.2 \pm 4.9$ & $67.4 \pm 4.8$ & $62.6 \pm 6.5$ \\
		\bottomrule
	\end{tabular}
	}
\end{table}

\begin{table}
	\caption{Mean and standard deviation of the inversion accuracy [\%] with local dataset size of $32$ in FedAvg training on the \textbf{Lawschool Admissions} dataset. The accuracy of the random baseline for $32$ datapoints is $57.8 \pm 2.3$.}
	\label{table:rec_fedavg_lawschool}
	\centering
	\resizebox{0.75\columnwidth}{!}{
	\begin{tabular}{lcccccc}
		\toprule
		 & \multicolumn{3}{c}{\OurMethod{}} & \multicolumn{3}{c}{Inverting Gradients~\citep{Geiping2020}}\\
		\cmidrule[1pt](l{5pt}r{5pt}){2-4}\cmidrule[1pt](l{5pt}r{5pt}){5-7}
		n. batches & 1 epoch & 5 epochs & 10 epochs & 1 epoch & 5 epochs & 10 epochs \\
		\midrule
		
		 1 & $\mathbf{85.4 \pm 4.2}$ & $\mathbf{82.9 \pm 3.1}$ & $\mathbf{81.7 \pm 4.0}$ & $72.2 \pm 2.6$ & $68.1 \pm 3.1$ & $65.2 \pm 2.8$ \\
		 2 & $\mathbf{86.2 \pm 4.3}$ & $\mathbf{82.8 \pm 3.0}$  & $\mathbf{81.4 \pm 3.1}$ & $72.5 \pm 1.9$ & $68.3 \pm 4.4$ & $66.2 \pm 2.8$ \\
		 4 & $\mathbf{85.7 \pm 4.4}$ & $\mathbf{81.5 \pm 3.8}$ & $\mathbf{80.3 \pm 4.5}$ & $72.5 \pm 2.4$ & $69.4 \pm 3.9$ & $67.9 \pm 3.8$ \\
		\bottomrule
	\end{tabular}
	}
\end{table}

\begin{table}
	\caption{Mean and standard deviation of the inversion accuracy [\%] with local dataset size of $32$ in FedAvg training on the \textbf{Health Heritage} dataset. The accuracy of the random baseline for $32$ datapoints is $43.4 \pm 3.5$.}
	\label{table:rec_fedavg_health}
	\centering
	\resizebox{0.75\columnwidth}{!}{
	\begin{tabular}{lcccccc}
		\toprule
		 & \multicolumn{3}{c}{\OurMethod{}} & \multicolumn{3}{c}{Inverting Gradients~\citep{Geiping2020}}\\
		\cmidrule[1pt](l{5pt}r{5pt}){2-4}\cmidrule[1pt](l{5pt}r{5pt}){5-7}
		n. batches & 1 epoch & 5 epochs & 10 epochs & 1 epoch & 5 epochs & 10 epochs \\
		\midrule
		
		 1 & $\mathbf{68.5 \pm 5.0}$ & $\mathbf{62.2 \pm 3.5}$ & $\mathbf{57.4 \pm 3.0}$ & $53.8 \pm 5.5$ & $41.4 \pm 3.6$ & $41.1 \pm 3.4$ \\
		 2 & $\mathbf{68.1 \pm 4.9}$ & $\mathbf{62.4 \pm 4.1}$  & $\mathbf{57.0 \pm 2.8}$ & $52.4 \pm 5.7$ & $43.4 \pm 4.28$ & $44.4 \pm 4.3$ \\
		 4 & $\mathbf{67.3 \pm 5.8}$ & $\mathbf{62.0 \pm 3.5}$ & $\mathbf{57.0 \pm 3.0}$ & $52.5 \pm 6.6$ & $43.4 \pm 5.7$ & $44.8 \pm 4.4$ \\
		\bottomrule
	\end{tabular}
	}
\end{table}

\clearpage
\subsection{Full Results on Entropy on all Datasets}
\label{appendix:full_results_on_entropy_on_all_datasets}
In \cref{table:global_entropy_adult_full}, \cref{table:global_entropy_german_full}, \cref{table:global_entropy_lawschool_full}, and \cref{table:global_entropy_health_full} we provide the mean and standard deviation of the reconstruction accuracy and the entropy of the continuous and the categorical features over increasing batch size for attacking with \OurMethod{} on the four datasets. Additionally, at each batch size we calculate and report the Kendall's $\tau$ rank correlation coefficient~\citep{Kendall1938} between the mean entropy of the features and the mean accuracy of the features over different batches. Note that if all features are correctly reconstructed, we can not calculate a rank correlation, in these cases we replace the missing value by NaN. We can observe on all datasets a trend of increasing entropy over decreasing reconstruction accuracy as the batch size is increased; and as such providing a signal to the attacker about their overall reconstruction success.

To generalize our results presented in \cref{sec:results} beyond Adult, we present the corresponding tables on the top and bottom quarters of the data based on the entropy ranking in \cref{table:local_entropy_adult_full}, \cref{table:local_entropy_german_full}, \cref{table:local_entropy_lawschool_full}, and \cref{table:local_entropy_health_full} for all four datasets, respectively. We can observe that the entropy is still effective in separating the poorly reconstructed features from the well reconstructed ones. The scheme is especially strong on the categorical features, which is concerning, because, as discussed in \cref{sec:results}, they are already much more vulnerable to leakage attacks.

\begin{table}
	\caption{The mean accuracy [\%] and entropies with the corresponding standard deviations over batch sizes of the categorical and the continuous features on the \textbf{Adult} dataset, together with the rank correlation between mean batch accuracy and mean batch entropy at the given batch size.}
	\label{table:global_entropy_adult_full}
	\centering
	\resizebox{0.8\columnwidth}{!}{
	\begin{tabular}{lcccccc}
		\toprule
			& \multicolumn{3}{c}{Discrete} & \multicolumn{3}{c}{Continuous}\\
		\cmidrule[1pt](l{5pt}r{5pt}){2-4}\cmidrule[1pt](l{5pt}r{5pt}){5-7}
			& Accuracy & Entropy & Kendall's $\tau$ & Accuracy & Entropy & Kendall's $\tau$\\
		\midrule
			1 & $100.0 \pm 0.0$ & $0.02 \pm 0.04$ & NaN & $98.7 \pm 6.5$ & $-4.00 \pm 0.72$ & $-0.28$\\
			2 & $99.8 \pm 1.7$ & $0.02 \pm 0.05$ & $-0.20$ & $98.7 \pm 9.3$ & $-3.75 \pm 0.93$ & $-0.20$\\
			4 & $99.6 \pm 1.7$ & $0.08 \pm 0.11$ & $-0.38$ & $96.2 \pm 9.0$ & $-2.61 \pm 1.34$ & $-0.56$\\
			8 & $98.3 \pm 6.1$ & $0.15 \pm 0.14$ & $-0.52$ & $91.0 \pm 14.3$ & $-1.69 \pm 1.14$ & $-0.66$\\
			16 & $97.2 \pm 4.3$ & $0.25 \pm 0.11$ & $-0.64$ & $80.0 \pm 12.9$ & $-0.63 \pm 0.62$ & $-0.65$\\
			32 & $91.5 \pm 4.1$ & $0.39 \pm 0.06$ & $-0.55$ & $63.1 \pm 6.7$ & $0.17 \pm 0.31$ & $-0.53$\\
			64 & $83.7 \pm 3.7$ & $0.47 \pm 0.04$ & $-0.65$ & $59.6 \pm 2.5$ & $0.57 \pm 0.22$ & $-0.34$\\
			128 & $79.2 \pm 1.6$ & $0.51 \pm 0.03$ & $-0.42$ & $61.3 \pm 1.6$ & $0.80 \pm 0.14$ & $-0.34$\\
		\bottomrule
	\end{tabular}
	}
\end{table}

\begin{table}
	\caption{The mean accuracy [\%] and entropies with the corresponding standard deviations over batch sizes of the categorical and the continuous features on the \textbf{German Credit} dataset, together with the rank correlation between mean batch accuracy and mean batch entropy at the given batch size.}
	\label{table:global_entropy_german_full}
	\centering
	\resizebox{0.8\columnwidth}{!}{
	\begin{tabular}{lcccccc}
		\toprule
			& \multicolumn{3}{c}{Discrete} & \multicolumn{3}{c}{Continuous}\\
		\cmidrule[1pt](l{5pt}r{5pt}){2-4}\cmidrule[1pt](l{5pt}r{5pt}){5-7}
			& Accuracy & Entropy & Kendall's $\tau$ & Accuracy & Entropy & Kendall's $\tau$\\
		\midrule
			1 & $100.0 \pm 0.0$ & $0.00 \pm 0.01$ & NaN & $100.0 \pm 0.0$ & $-4.81 \pm 0.62$ & NaN\\
			2 & $100.0 \pm 0.0$ & $0.02 \pm 0.03$ & NaN & $100.0 \pm 0.0$ & $-4.12 \pm 1.36$ & NaN\\
			4 & $100.0 \pm 0.0$ & $0.06 \pm 0.05$ & NaN & $99.7 \pm 1.2$ & $-2.75 \pm 1.33$ & $-0.19$\\
			8 & $100.0 \pm 0.0$ & $0.11 \pm 0.07$ & NaN & $98.9 \pm 3.4$ & $-1.92 \pm 1.01$ & $-0.23$\\
			16 & $99.6 \pm 1.3$ & $0.24 \pm 0.08$ & $-0.35$ & $90.1 \pm 8.0$ & $-0.74 \pm 0.33$ & $-0.38$\\
			32 & $93.4 \pm 2.2$ & $0.42 \pm 0.04$ & $-0.52$ & $67.3 \pm 4.8$ & $0.28 \pm 0.17$ & $-0.28$\\
			64 & $82.5 \pm 1.8$ & $0.55 \pm 0.02$ & $-0.64$ & $59.3 \pm 2.1$ & $0.80 \pm 0.07$ & $-0.29$\\
			128 & $78.5 \pm 1.1$ & $0.58 \pm 0.02$ & $-0.27$ & $61.2 \pm 1.3$ & $1.01 \pm 0.04$ & $-0.25$\\
		\bottomrule
	\end{tabular}
	}
\end{table}

\begin{table}
	\caption{The mean accuracy [\%] and entropies with the corresponding standard deviations over batch sizes of the categorical and the continuous features on the \textbf{Lawschool Admissions} dataset, together with the rank correlation between mean batch accuracy and mean batch entropy at the given batch size.}
	\label{table:global_entropy_lawschool_full}
	\centering
	\resizebox{0.8\columnwidth}{!}{
	\begin{tabular}{lcccccc}
		\toprule
			& \multicolumn{3}{c}{Discrete} & \multicolumn{3}{c}{Continuous}\\
		\cmidrule[1pt](l{5pt}r{5pt}){2-4}\cmidrule[1pt](l{5pt}r{5pt}){5-7}
			& Accuracy & Entropy & Kendall's $\tau$ & Accuracy & Entropy & Kendall's $\tau$\\
		\midrule
		1 & $100.0 \pm 0.0$ & $0.01 \pm 0.02$ & NaN & $100.0 \pm 0.0$ & $-2.94 \pm 0.32$ & NaN\\
		2 & $100.0 \pm 0.0$ & $0.02 \pm 0.05$ & NaN & $100.0 \pm 0.0$ & $-2.52 \pm 0.67$ & NaN\\
		4 & $100.0 \pm 0.0$ & $0.03 \pm 0.04$ & NaN & $100.0 \pm 0.0$ & $-2.25 \pm 0.53$ & NaN\\
		8 & $99.8 \pm 1.1$ & $0.10 \pm 0.10$ & $-0.28$ & $96.5 \pm 11.1$ & $-1.66 \pm 0.47$ & $-0.25$\\
		16 & $98.4 \pm 2.8$ & $0.23 \pm 0.11$ & $-0.37$ & $87.6 \pm 14.0$ & $-0.62 \pm 0.42$ & $-0.50$\\
		32 & $93.4 \pm 2.9$ & $0.42 \pm 0.08$ & $-0.50$ & $65.3 \pm 8.5$ & $0.21 \pm 0.20$ & $-0.45$\\
		64 & $86.9 \pm 2.5$ & $0.55 \pm 0.05$ & $-0.51$ & $58.4 \pm 4.6$ & $0.80 \pm 0.11$ & $-0.20$\\
		128 & $83.5 \pm 1.6$ & $0.60 \pm 0.03$ & $-0.36$ & $62.7 \pm 3.1$ & $1.06 \pm 0.10$ & $-0.18$\\
		\bottomrule
	\end{tabular}
	}
\end{table}

\begin{table}
	\caption{The mean accuracy [\%] and entropies with the corresponding standard deviations over batch sizes of the categorical and the continuous features on the \textbf{Health Heritage} dataset, together with the rank correlation between mean batch accuracy and mean batch entropy at the given batch size.}
	\label{table:global_entropy_health_full}
	\centering
	\resizebox{0.8\columnwidth}{!}{
	\begin{tabular}{lcccccc}
		\toprule
			& \multicolumn{3}{c}{Discrete} & \multicolumn{3}{c}{Continuous}\\
		\cmidrule[1pt](l{5pt}r{5pt}){2-4}\cmidrule[1pt](l{5pt}r{5pt}){5-7}
			& Accuracy & Entropy & Kendall's $\tau$ & Accuracy & Entropy & Kendall's $\tau$\\
		\midrule
		1 & $100.0 \pm 0.0$ & $0.02 \pm 0.04$ & NaN & $99.6 \pm 2.5$ & $-3.45 \pm 0.70$ & $-0.18$\\
		2 & $100.0 \pm 0.0$ & $0.05 \pm 0.09$ & NaN & $96.4 \pm 12.3$ & $-2.88 \pm 0.97$ & $-0.47$\\
		4 & $99.6 \pm 2.4$ & $0.08 \pm 0.10$ & $-0.28$ & $97.0 \pm 10.6$ & $-1.92 \pm 0.96$ & $-0.34$\\
		8 & $98.4 \pm 5.2$ & $0.13 \pm 0.11$ & $-0.43$ & $93.9 \pm 11.8$ & $-1.19 \pm 0.76$ & $-0.56$\\
		16 & $96.0 \pm 8.5$ & $0.26 \pm 0.10$ & $-0.61$ & $79.5 \pm 12.1$ & $-0.25 \pm 0.49$ & $-0.55$\\
		32 & $85.8 \pm 5.9$ & $0.42 \pm 0.06$ & $-0.62$ & $63.2 \pm 4.3$ & $0.47 \pm 0.24$ & $-0.48$\\
		64 & $73.9 \pm 4.5$ & $0.50 \pm 0.04$ & $-0.60$ & $60.6 \pm 2.9$ & $0.78 \pm 0.20$ & $-0.48$\\
		128 & $68.1 \pm 2.0$ & $0.55 \pm 0.02$ & $-0.26$ & $61.0 \pm 2.3$ & $1.03 \pm 0.11$ & $-0.42$\\
		\bottomrule
	\end{tabular}
	}
\end{table}

\begin{table}
	\caption{The mean and standard deviation of the accuracy [\%] of each feature type in the top $25\%$ and the bottom $25\%$ when ranked in the batch according to the entropy on the \textbf{Adult} dataset.}
	\centering
	\resizebox{.6\columnwidth}{!}{
	\begin{tabular}{lcccc}
		\toprule
		 Batch & \multicolumn{2}{c}{Categorical} & \multicolumn{2}{c}{Continuous}\\
		\cmidrule[1pt](l{5pt}r{5pt}){2-3}\cmidrule[1pt](l{5pt}r{5pt}){4-5}
		Size  & Top $25\%$ & Bottom $25\%$ & Top $25\%$ & Bottom $25\%$ \\
		\midrule
		 1 & $100.0 \pm 0.0$ & $100.0 \pm 0.0$ & $99.0 \pm 7.0$ & $98.0 \pm 9.8$\\
		 2 & $100.0 \pm 0.0$ & $99.3 \pm 4.7$ & $98.7 \pm 9.3$ & $98.7 \pm 9.3$\\
		 4 & $100.0 \pm 0.0$ & $98.0 \pm 7.2$ & $99.7 \pm 2.3$ & $92.7 \pm 18.6$\\
		 8 & $99.7 \pm 2.3$ & $96.0 \pm 10.3$ & $97.2 \pm 9.4$ & $84.2 \pm 21.6$\\
		 16 & $99.9 \pm 0.6$ & $89.8 \pm 13.0$ & $98.2 \pm 3.5$ & $60.8 \pm 19.9$\\
		 32 & $99.1 \pm 2.6$ & $75.5 \pm 8.0$ & $94.2 \pm 4.7$ & $43.6 \pm 8.2$\\
		 64 & $97.8 \pm 2.6$ & $66.1 \pm 5.3$ & $92.9 \pm 3.5$ & $41.3 \pm 5.8$\\
		 128 & $94.3 \pm 1.9$ & $62.8 \pm 3.8$ & $93.5 \pm 2.3$ & $42.2 \pm 3.7$\\
		\bottomrule
	\end{tabular}}
	\vspace{-0.9em}
	\label{table:local_entropy_adult_full}
\end{table}

\begin{table}
	\caption{The mean and standard deviation of the accuracy [\%] of each feature type in the top $25\%$ and the bottom $25\%$ when ranked in the batch according to the entropy \textbf{German Credit} dataset.}
	\centering
	\resizebox{.6\columnwidth}{!}{
	\begin{tabular}{lcccc}
		\toprule
		 Batch & \multicolumn{2}{c}{Categorical} & \multicolumn{2}{c}{Continuous}\\
		\cmidrule[1pt](l{5pt}r{5pt}){2-3}\cmidrule[1pt](l{5pt}r{5pt}){4-5}
		Size  & Top $25\%$ & Bottom $25\%$ & Top $25\%$ & Bottom $25\%$ \\
		\midrule
		 1 & $100.0 \pm 0.0$ & $100.0 \pm 0.0$ & $100.0 \pm 0.0$ & $100.0 \pm 0.0$\\
		 2 & $100.0 \pm 0.0$ & $100.0 \pm 0.0$ & $100.0 \pm 0.0$ & $100.0 \pm 0.0$\\
 		 4 & $100.0 \pm 0.0$ & $100.0 \pm 0.0$ & $100.0 \pm 0.0$ & $98.9 \pm 4.8$\\
		 8 & $100.0 \pm 0.0$ & $100.0 \pm 0.0$ & $99.7 \pm 2.0$ & $97.7 \pm 5.4$\\
		 16 & $100.0 \pm 0.0$ & $97.8 \pm 6.2$ & $98.2 \pm 3.5$ & $80.6 \pm 13.7$\\
		 32 & $100.0 \pm 0.0$ & $75.2 \pm 7.1$ & $87.3 \pm 5.9$ & $51.2 \pm 7.1$\\
		 64 & $98.9 \pm 1.3$ & $66.7 \pm 4.0$ & $77.9 \pm 5.6$ & $47.2 \pm 4.8$\\
		 128 & $96.9 \pm 1.4$ & $66.0 \pm 2.8$ & $78.0 \pm 3.0$ & $48.9 \pm 2.9$\\
		\bottomrule
	\end{tabular}}
	\vspace{-0.9em}
	\label{table:local_entropy_german_full}
\end{table}

\begin{table}
	\caption{The mean and standard deviation of the accuracy [\%] of each feature type in the top $25\%$ and the bottom $25\%$ when ranked in the batch according to the entropy \textbf{Lawschool Admissions} dataset.}
	\centering
	\resizebox{.6\columnwidth}{!}{
	\begin{tabular}{lcccc}
		\toprule
		 Batch & \multicolumn{2}{c}{Categorical} & \multicolumn{2}{c}{Continuous}\\
		\cmidrule[1pt](l{5pt}r{5pt}){2-3}\cmidrule[1pt](l{5pt}r{5pt}){4-5}
		Size  & Top $25\%$ & Bottom $25\%$ & Top $25\%$ & Bottom $25\%$ \\
		\midrule
		1 & $100.0 \pm 0.0$ & $100.0 \pm 0.0$ & $100.0 \pm 0.0$ & $100.0 \pm 0.0$ \\
		2 & $100.0 \pm 0.0$ & $100.0 \pm 0.0$ & $100.0 \pm 0.0$ & $100.0 \pm 0.0$ \\
		4 & $100.0 \pm 0.0$ & $100.0 \pm 0.0$ & $100.0 \pm 0.0$ & $100.0 \pm 0.0$ \\
		8 & $100.0 \pm 0.0$ & $98.5 \pm 7.8$ & $99.5 \pm 3.5$ & $91.5 \pm 23.2$ \\
		16 & $100.0 \pm 0.0$ & $89.2 \pm 17.3$ & $96.5 \pm 10.9$ & $75.2 \pm 24.6$ \\
		32 & $99.9 \pm 0.9$ & $77.5 \pm 9.3$ & $76.6 \pm 11.9$ & $53.4 \pm 13.4$ \\
		64 & $96.6 \pm 3.5$ & $76.0 \pm 9.4$ & $62.7 \pm 9.7$ & $53.0 \pm 9.3$ \\
		128 & $93.6 \pm 3.1$ & $70.5 \pm 7.6$ & $68.2 \pm 5.9$ & $56.6 \pm 6.0$ \\
		\bottomrule
	\end{tabular}}
	\vspace{-0.9em}
	\label{table:local_entropy_lawschool_full}
\end{table}

\begin{table}
	\caption{The mean and standard deviation of the accuracy [\%] of each feature type in the top $25\%$ and the bottom $25\%$ when ranked in the batch according to the entropy \textbf{Health Heritage} dataset.}
	\centering
	\resizebox{.6\columnwidth}{!}{
	\begin{tabular}{lcccc}
		\toprule
		 Batch & \multicolumn{2}{c}{Categorical} & \multicolumn{2}{c}{Continuous}\\
		\cmidrule[1pt](l{5pt}r{5pt}){2-3}\cmidrule[1pt](l{5pt}r{5pt}){4-5}
		Size  & Top $25\%$ & Bottom $25\%$ & Top $25\%$ & Bottom $25\%$ \\
		\midrule
		1 & $100.0 \pm 0.0$ & $100.0 \pm 0.0$ & $100.0 \pm 0.0$ & $98.7 \pm 9.3$\\
		2 & $100.0 \pm 0.0$ & $100.0 \pm 0.0$ & $97.7 \pm 10.5$ & $94.3 \pm 16.9$\\
		4 & $99.8 \pm 1.3$ & $99.3 \pm 4.0$ & $98.0 \pm 10.8$ & $95.8 \pm 13.6$\\
		8 & $99.8 \pm 1.3$ & $96.7 \pm 10.4$ & $97.7 \pm 5.8$ & $88.5 \pm 20.6$\\
		16 & $99.0 \pm 5.4$ & $92.7 \pm 12.4$ & $92.3 \pm 9.3$ & $64.8 \pm 16.8$\\
		32 & $95.3 \pm 3.8$ & $75.9 \pm 8.8$ & $77.2 \pm 6.7$ & $51.1 \pm 8.4$\\
		64 & $88.3 \pm 4.5$ & $59.6 \pm 6.0$ & $73.3 \pm 5.6$ & $52.3 \pm 6.1$\\
		128 & $82.4 \pm 2.3$ & $54.5 \pm 3.1$ & $71.2 \pm 4.3$ & $56.3 \pm 4.0$\\
		\bottomrule
	\end{tabular}}
	\vspace{-0.9em}
	\label{table:local_entropy_health_full}
\end{table}

\pagebreak
\section{Studying Pooling}
\label{appendix:studying_pooling}
In this subsection, we present three further experiments on justifying and understanding our choices in pooling:
\begin{itemize}
	\item Experiments on synthetic datasets for understanding the motivation for pooling in \cref{appendix:variance_study}.
	\item Ablation study on understanding the impact of the number of samples $N$ in the collection before pooling on the performance of \OurMethod{} in \cref{appendix:the_impact_of_the_number_of_samples_n}.
	\item Comparison of using mean and median pooling on \OurMethod{} in \cref{appendix:choice_of_the_pooling_function}.
\end{itemize}
Note that in the experiments below we do not make use of the sigmoid restricting the continuous features.

\subsection{Variance Study}
\label{appendix:variance_study}
A unique challenge (challenge (i)) of tabular data leakage is that the mix of discrete and continuous features introduces further variance in the final reconstructions. As a solution to this challenge, we propose to produce $N$ independent reconstructions of the same batch, and ensemble them using the pooling scheme described in \cref{subsec:pooled_ensembling}. In this subsection, we provide empirical evidence for the subject of challenge (i) and the effectiveness of our proposed solution to it.

\paragraph{Experimental Setup} 
We create $6$ synthetic binary classification datasets, each with $10$ features, however of varying modality. Concretely; we have the following setups:
\begin{itemize}
	\item Synthetic dataset with 0 discrete and 10 continuous columns,
	\item Synthetic dataset with 2 discrete and 8 continuous columns,
	\item Synthetic dataset with 4 discrete and 6 continuous columns,
	\item Synthetic dataset with 6 discrete and 4 continuous columns,
	\item Synthetic dataset with 8 discrete and 2 continuous columns,
	\item Synthetic dataset with 10 discrete and 0 continuous columns.
\end{itemize}
The continuous features are Gaussians with means between 0 and 5, and standard deviations between 1 and 3. The discrete features have domain sizes between 2 and 6, and the probabilities are drawn randomly. On each of these datasets we sample 50 batches of size 32 and reconstruct them using TabLeak (no pooling) starting from 30 different initializations in the same experimental setup elaborated in \cref{sec:results} and in \cref{appendix:further_experimental_details}. We then proceed to calculate the standard deviation of the accuracy for each of the 50 batches over their 30 independent reconstructions, providing us 50 statistically independent data points for understanding the variance in the non-pooled reconstruction problem. Further, from the 30 independent reconstructions of each batch, we build 6 independent mini-ensembles of size 5 and conduct median pooling on them (essentially, TabLeak with $N=5$). We then measure the standard deviation of the error for each of the 50 batches over the 6 obtained pooled reconstructions, obtaining 50 independent data points for analyzing the variance of pooled reconstruction.

\begin{figure}
	\centering
	\begin{subfigure}{.33\textwidth}
		\centering
		\includegraphics[width=0.95\textwidth]{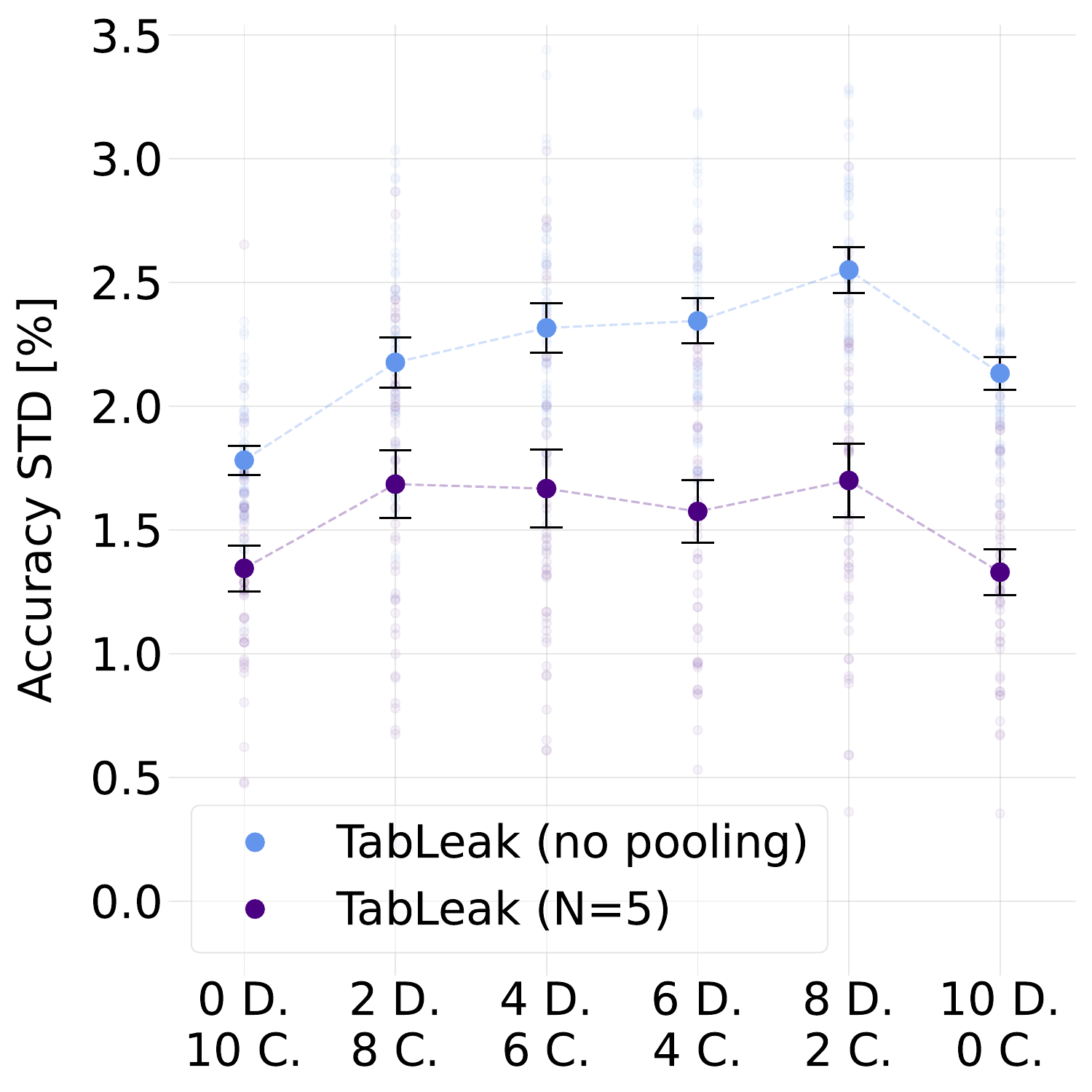}
		\subcaption{Combined Accuracy STD}
		\label{fig:variance_study_a}
	\end{subfigure}%
	\begin{subfigure}{.33\textwidth}
		\centering
		\includegraphics[width=0.95\textwidth]{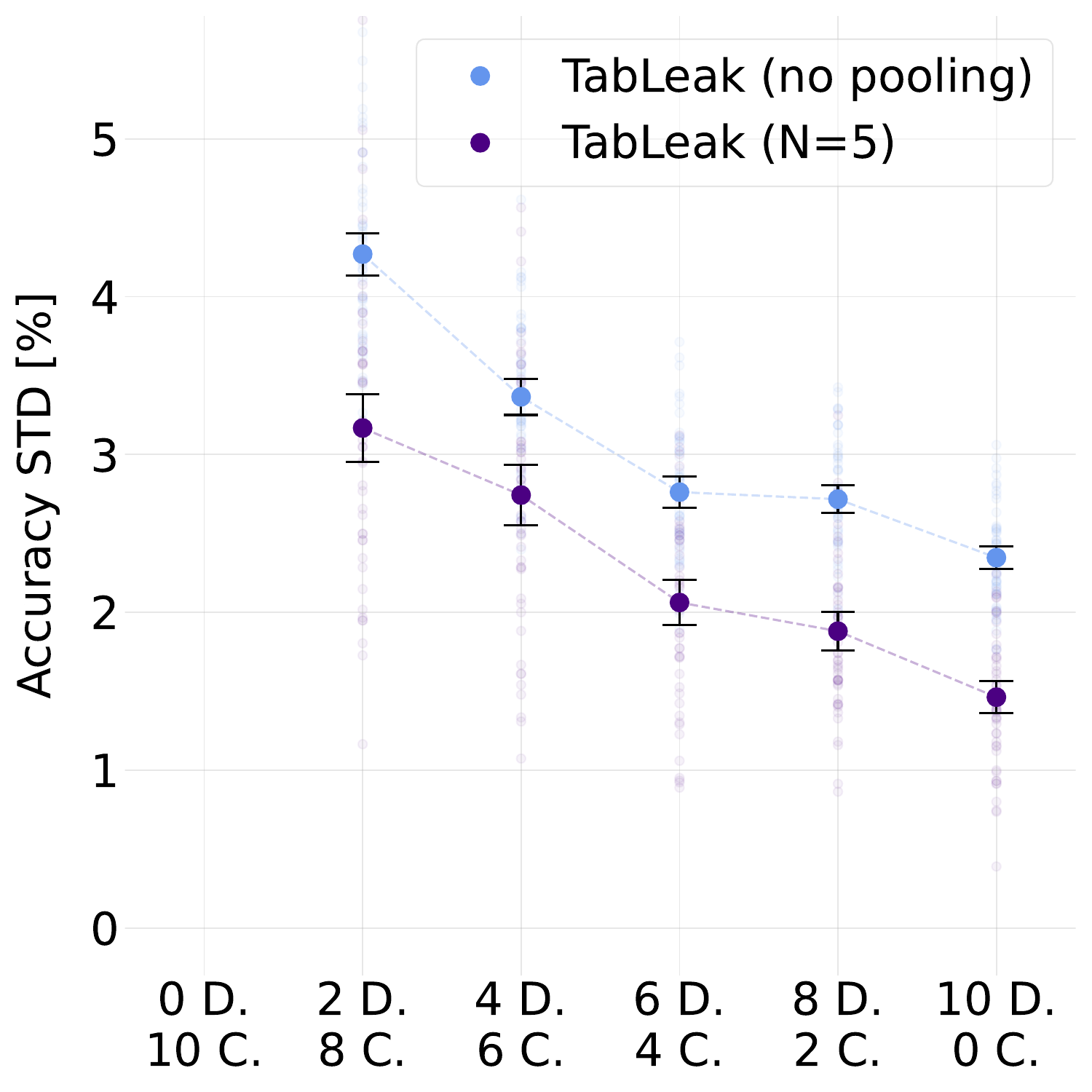}
		\subcaption{Discrete Accuracy STD}
		\label{fig:variance_study_b}
	\end{subfigure}
	\begin{subfigure}{.33\textwidth}
		\centering
		\includegraphics[width=0.95\textwidth]{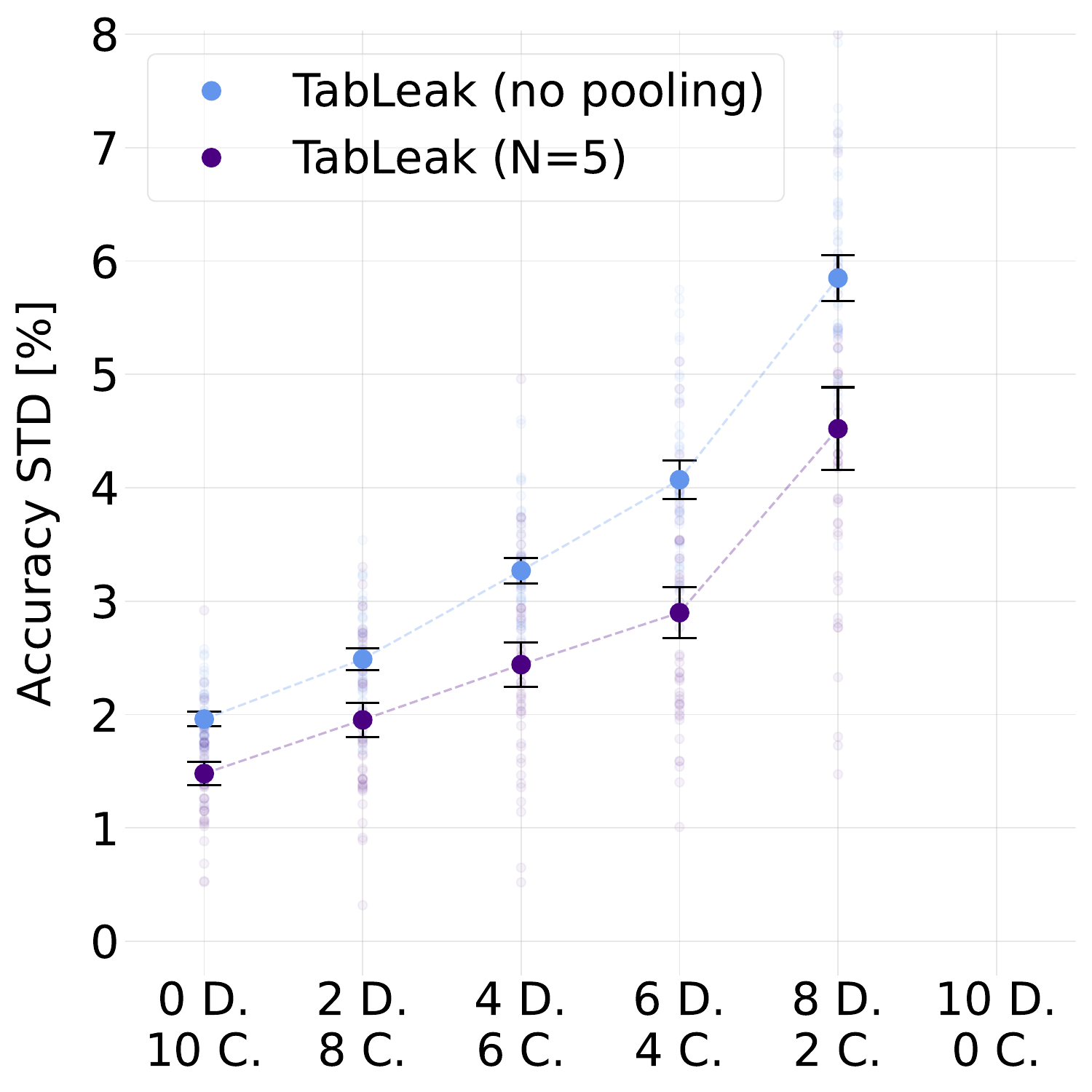}
		\subcaption{Continuous Accuracy STD}
		\label{fig:variance_study_c}
	\end{subfigure}
	\caption{Mean same-batch reconstruction accuracy standard deviation and 90\% confidence interval at batch size 32 estimated from 50 independent batches over synthetic datasets with varying number of discrete (D.) and continuous (C.) features.}
	\label{fig:variance_study}
\end{figure}

\paragraph{Results}
We present the results of the experiment in \cref{fig:variance_study}; additionally to measuring the same-batch reconstruction accuracy standard deviation for all features together, we also present the resulting measurements when only considering the discrete and the continuous features, respectively. The figures are organized such that the x-axis begins with the synthetic dataset consisting only of continuous features and progresses to the right by decreasing the number of continuous and increasing the number of discrete features at each step by 2. Roughly speaking, the very left column of the figures is similar to data leakage in the image domain, where all features are continuous, and the very right relates to data leakage in the text domain, containing only discrete features. Looking at \cref{fig:variance_study_a}, we observe that the mean same-batch STD is indeed higher for datasets consisting of mixed types, providing empirical evidence underlining the first challenge of tabular data leakage. Further, it can be clearly seen that pooling, even with a small ensemble of just 5 samples, decisively decreases the variance of the reconstruction problem, providing strong justification for using pooling in the tabular setting. Finally, from \cref{fig:variance_study_b} and \cref{fig:variance_study_c} we gain interesting insight in the underlying dynamics of the interplay between discrete and continuous features at reconstruction. Concretely, we observe that as the presence of a given modality is decreasing and its place is taken up by the other, the recovery of this modality becomes increasingly noisier. Much in line with the observations on the difference in the recovery success between discrete and continuous features, these results also argue for future work to pursue methods that decrease the disparity between the two different feature types in the mixed setting.

\subsection{The Impact of the Number of Samples N}
\label{appendix:the_impact_of_the_number_of_samples_n}
In \cref{table:effect_of_N} we present the results of an ablation study we conducted on \OurMethod{} at batch size 32 to understand the impact of the size of the ensemble $N$ on the performance of the attack. We observed that with increasing N the performance of the attack gets steadily better, albeit, producing diminishing returns, showing signs of saturation on some datasets after $N=25$. Note that this behavior is expected, and suggests using the largest $N$ that is not yet computationally  prohibitive. We chose $N=30$ for all our experiments with \OurMethod{} (unless explicitly stated otherwise); this allowed us to conduct large-scale experiments while still extracting good performance from \OurMethod{}.

\begin{table}
	\caption{Reconstruction accuracy [\%] and standard deviation of \OurMethod{} on batch size 32 over the size of the ensemble $N$ used for pooling.}
	\label{table:effect_of_N}
	\centering
	\resizebox{0.99\columnwidth}{!}{
	\begin{tabular}{lccccccc}
		\toprule
		 $N =$ & 1 & 5 & 10 & 15 & 20 & 25 & 30\\
		\midrule
		Adult & $71.8 \pm 4.6$ & $75.1 \pm 4.6$ & $76.5 \pm 4.6$ & $76.5 \pm 4.8$ & $77.1 \pm 4.7$ & $77.0 \pm 4.7$ & $\mathbf{77.4 \pm 4.8}$ \\
		German & $79.0 \pm 3.0$ & $81.6 \pm 2.9$ & $82.6 \pm 2.9$ & $82.9 \pm 2.9$ & $83.2 \pm 2.8$ & $83.4 \pm 2.8$ & $\mathbf{83.6 \pm 2.7}$ \\
		Lawschool & $82.3 \pm 4.0$ & $84.4 \pm 3.7$ & $84.7 \pm 4.1$ & $85.2 \pm 3.9$ & $85.1 \pm 3.9$ & $\mathbf{85.3 \pm 4.0}$ & $\mathbf{85.3 \pm 4.0}$ \\
		Health Heritage & $64.6 \pm 4.2$ & $67.5 \pm 4.3$ & $69.1 \pm 4.4$ & $69.1 \pm 4.5$ & $69.7 \pm 4.5$ & $69.5 \pm 4.3$ & $\mathbf{70.1 \pm 4.0}$ \\
		\bottomrule
	\end{tabular}
	}
\end{table}

\subsection{Choice of the Pooling Function}
\label{appendix:choice_of_the_pooling_function}
We compare \OurMethod{} using median pooling to \OurMethod{} with mean pooling in \cref{table:mean_vs_median} over the four datasets. As we can observe, in most cases both methods produce similar results, hence the effectiveness of \OurMethod{} is mostly independent of this choice. However, as median pooling demonstrates to provide a slight edge in some cases, we opt for using median pooling in our main experiments with \OurMethod{}.

\begin{table}[b]
	\centering
	\begin{subtable}{.49\textwidth}
		\centering
		\begin{tabular}{lcc}
			\toprule
			batch	& \OurMethod{} & \OurMethod{}\\
			size	& (median) & (mean) \\
			\midrule
				1 & $\mathbf{99.4 \pm 2.8}$ & $\mathbf{99.4 \pm 2.8}$  \\
				2 & $99.2 \pm 5.5$ & $\mathbf{99.3 \pm 5.0}$ \\
				4 & $\mathbf{98.0 \pm 4.5}$ & $97.7 \pm 5.3$ \\
				8 & $\mathbf{95.1 \pm 9.2}$ & $94.8 \pm 9.0$  \\
				16 & $\mathbf{89.4 \pm 7.6}$ & $88.9 \pm 7.7$ \\
				32 & $\mathbf{77.6 \pm 4.8}$ & $77.1 \pm 4.7$  \\
				64 & $71.2 \pm 2.8$ & $\mathbf{71.7 \pm 2.8}$ \\
				128 & $68.8 \pm 1.3$ & $\mathbf{69.4 \pm 1.4}$  \\
			\bottomrule
		\end{tabular}
		\subcaption{Adult}
	\end{subtable}%
	\begin{subtable}{.49\textwidth}
		\centering
		\begin{tabular}{lcc}
			\toprule
			batch & \OurMethod{} & \OurMethod{}\\
			size & (median) & (mean) \\
			\midrule
				1 & $\mathbf{100.0 \pm 0.0}$ & $\mathbf{100.0 \pm 0.0}$  \\
				2 & $\mathbf{100.0 \pm 0.0}$ & $\mathbf{100.0 \pm 0.0}$ \\
				4 & $\mathbf{99.9 \pm 0.4}$ & $\mathbf{99.9 \pm 0.4}$ \\
				8 & $\mathbf{99.7 \pm 1.1}$ & $99.6 \pm 1.1$  \\
				16 & $\mathbf{95.9 \pm 3.4}$ & $95.6 \pm 3.3$ \\
				32 & $\mathbf{83.6 \pm 2.9}$ & $83.1 \pm 3.0$  \\
				64 & $\mathbf{73.0 \pm 1.3}$ & $72.6 \pm 1.3$ \\
				128 & $\mathbf{71.3 \pm 0.8}$ & $70.8 \pm 0.9$  \\
			\bottomrule
		\end{tabular}
		\subcaption{German Credit}
	\end{subtable}
	\par\bigskip
	\begin{subtable}{.49\textwidth}
		\centering
		\begin{tabular}{lcc}
			\toprule
			batch & \OurMethod{} & \OurMethod{}\\
			size & (median) & (mean) \\
			\midrule
				1 & $\mathbf{100.0 \pm 0.0}$ & $\mathbf{100.0 \pm 0.0}$  \\
				2 & $\mathbf{100.0 \pm 0.0}$ & $\mathbf{100.0 \pm 0.0}$ \\
				4 & $\mathbf{100.0 \pm 0.0}$ & $\mathbf{100.0 \pm 0.0}$ \\
				8 & $98.7 \pm 3.8$ & $\mathbf{98.8 \pm 3.4}$  \\
				16 & $\mathbf{94.8 \pm 5.6}$ & $94.6 \pm 5.4$ \\
				32 & $\mathbf{84.8 \pm 3.9}$ & $84.7 \pm 3.9$  \\
				64 & $\mathbf{78.2 \pm 2.0}$ & $\mathbf{78.2 \pm 2.2}$ \\
				128 & $77.3 \pm 1.2$ & $\mathbf{77.5 \pm 1.2}$  \\
			\bottomrule
		\end{tabular}
		\subcaption{Lawschool Admissions}
	\end{subtable}
	\begin{subtable}{.49\textwidth}
		\centering
		\begin{tabular}{lcc}
			\toprule
			batch & \OurMethod{} & \OurMethod{}\\
			size & (median) & (mean) \\
			\midrule
				1 & $\mathbf{99.8 \pm 1.6}$ & $\mathbf{99.8 \pm 1.6}$  \\
				2 & $\mathbf{97.7 \pm 8.3}$ & $97.4 \pm 9.0$ \\
				4 & $\mathbf{98.2 \pm 6.5}$ & $98.0 \pm 6.7$ \\
				8 & $\mathbf{96.0 \pm 8.2}$ & $95.6 \pm 8.6$  \\
				16 & $\mathbf{86.1 \pm 8.8}$ & $84.9 \pm 9.3$ \\
				32 & $\mathbf{70.0 \pm 4.5}$ & $69.7 \pm 4.4$  \\
				64 & $64.7 \pm 2.8$ & $\mathbf{64.8 \pm 2.9}$ \\
				128 & $63.0 \pm 1.4$ & $\mathbf{63.6 \pm 1.5}$  \\
			\bottomrule
		\end{tabular}
		\subcaption{Health Heritage}
	\end{subtable}
	\caption{Mean and standard deviation of the reconstruction accuracy [\%] using \OurMethod{} with either median or mean pooling, assuming full knowledge of the true labels.}
	\label{table:mean_vs_median}
\end{table}

\end{document}